\RequirePackage{fix-cm}
\documentclass[twocolumn, natbib]{svjour3}          
\smartqed  
\usepackage{graphicx}
\usepackage{amsmath}
\usepackage{amssymb}
\usepackage{subfigure}
\usepackage{rotating}
\usepackage{array}
\usepackage{xspace}
\usepackage{booktabs}
\usepackage{multirow}
\usepackage{tabularx}
\usepackage{longtable}
\usepackage[usenames,dvipsnames]{color}
\usepackage{mathtools}

\def\etal{\emph{et al}. }
\def\eg{\emph{e.g}. }

\def\x2{\chi^2}

\makeatother

\makeatletter
\def\input@path{{images/}}
\makeatother

\newenvironment{myQuote}
    {\begin{center}\begin{minipage}[h]{.45\textwidth}\noindent\fontfamily{\ttdefault}\bfseries\fontencoding{T1}\selectfont}
    {\end{minipage}\end{center}}
    
\newcolumntype{M}{>{\centering\arraybackslash}m{\dimexpr.25\linewidth-2\tabcolsep}}

\graphicspath{{images/}}

\begin{document}

\title{Discovering beautiful attributes for aesthetic image analysis}


\author{Luca Marchesotti \and
		Naila Murray \and
        Florent Perronnin 
}


\institute{Luca Marchesotti \at
Beautifeye\\
Suite 203, Media Cube, IADT‎\\
Baile \'{A}tha Cliath, Ireland\\
Xerox Research Centre Europe\\
Meylan, France\\
\email{luca@beautifeye.co}\\
\and
			  Naila Murray and Florent Perronnin \at
              Xerox Research Centre Europe\\
Meylan, France\\
              \email{Firstname.Lastname@xrce.xerox.com}
}

\date{Received: date / Accepted: date}

\maketitle

\begin{abstract}
Aesthetic image analysis is the study and assessment of the aesthetic properties of images.
Current computational approaches to aesthetic image analysis either provide accurate {\em or} interpretable results.
To obtain both accuracy {\em and} interpretability by humans,
we advocate the use of {\em learned} and {\em nameable} visual attributes as mid-level features.
For this purpose, we propose to discover and learn the visual appearance of
attributes automatically, using
a recently introduced database, called AVA, which contains more than 250,000 images
together with their aesthetic scores and textual comments given by photography enthusiasts.
We provide a detailed analysis of these annotations as well as the context in which they were given.
We then describe how these three key components of AVA - images, scores, and comments - can be effectively 
leveraged to learn visual attributes.
Lastly, we show that these learned attributes can be successfully used in three applications: aesthetic quality
prediction, image tagging and retrieval.

\end{abstract}


\section{Introduction}
The volume of visual data we handle on a daily basis is growing exponentially,
and will continue to do so due to the availability of ubiquitous and cheap sensors, sharing platforms and new social trends. 
Artificial intelligence systems have proven useful for processing and interpreting this preponderance of data. 
In the last decade, the computer vision and image retrieval community was focused on developing tools for semantic analysis of multimedia content.
While this is still a very active research field, new questions are arising.
These questions are about visual properties {\em beyond} visual semantics,
such as image preference \citep{DJLW06}, affectiveness \citep{Machajdik2010}, and memorability \citep{isola2011understanding}, as well as object importance \citep{berg2012understanding}.
Answering subjective, human-centric questions such as ``would someone find this image aesthetically pleasing"
is very challenging, even for humans. 
However, it was experimentally shown that these visual cognition phenomena can be predicted using data-driven approaches \citep{Luo2008,DJLW08,Machajdik2010,dhar2011,Marches2011,murray12}. 
In this work we focus on image preference: that is, whether people will like an image and which visual elements makes it un/attractive.

Early work on image preference prediction \citep{DJLW06,Ke2006} proposed to mimic the best practices of professional photographers.
In a nutshell, the idea was 
(i) to select rules (\eg ``contains opposing colors'') from photographic resources such as \citep{KODAK82} and 
(ii) to design for each rule a visual feature to predict the image compliance (\eg a color histogram).
Many subsequent works have focused on adding new photographic rules and on improving the visual features
of existing rules \citep{Luo2008,dhar2011}.
As noted for instance in \citep{dhar2011} these rules can be understood as visual attributes 
\citep{ferrari2008learning,lampert2009learning,farhadi2009describing},
i.e. medium-level descriptions whose purpose is to bridge the gap 
between the high-level concepts to be recognized (beautiful vs. ugly in our case) 
and the low-level pixels.
However, there are at least two issues with such an approach to aesthetic prediction.
Firstly, the hand-selection of attributes from a photographic guide is not exhaustive
and does not give any indication of when, and to what extent, such rules are used.
Secondly, hand-designed visual features only imperfectly model the corresponding
rules. 

As an alternative to rules and hand-designed features, 
it was proposed in \citep{Marches2011} to rely on generic features such as  
the GIST~\citep{OT01}, the bag-of-visual-words (BOV)~\citep{CD04} or the Fisher vector (FV)~\citep{PSM10}.
While it was shown experimentally that such an approach can lead to improved results with respect
to hand-designed attribute techniques, a major shortcoming is that interpretability of the results is lost.
In other words, while it is possible to say that an image has a high or low aesthetic value, 
it is impossible to tell why.
We thus raise the following question: 
{\em can we preserve the advantages of generic features \textbf{and} obtain interpretable results?}
In this work, we will address this problem by {\em discovering and learning attributes automatically}.

As described by \citep{parikh2011relative}, \textit{``[a]ttributes represent a class-discriminative, but not class-specific property that both computers and humans can decide on''}.
Such a statement implies that attributes should be understandable by humans.
Because selecting attributes by hand-picking photographic rules is problematic,
we intend to automatically discover attributes using a data-driven approach.
A natural way to enforce interpretability of the automatically discovered attributes is to mine them from natural text corpora,
as done for instance in \citep{berg2010automatic}.
We adopt this approach, and mine attributes using aesthetics-related textual terms associated with images.
The discovery process is as follows: 
(i) textual image meta-data are used to form a vocabulary of aesthetic terms; 
(ii) the discriminability of each vocabulary term is assessed and the {\em most discriminative} terms are retained as textual attributes;
(iii) visual appearance models for these textual attributes are trained using generic image descriptors and the {\em most detectable} models are retained as visual attributes.

Such an approach however, has a key requirement: a database with a unique conjunction of aesthetics-related content, namely
(i) textual meta-data from which to mine for aesthetic terms; 
(ii) aesthetic preference scores to provide supervisory information when assessing the discriminability of attributes;
(iii) images on which to train visual attribute models for textual aesthetic attributes.
While several datasets exist which contain images and associated preference scores, 
to our knowledge only the recently-introduced AVA dataset \citep{murray12} contains the full set of required content.
AVA contains more than 250,000 images along with preference score distributions and textual comments given to images by photography enthusiasts. 
As such, we propose to leverage AVA as an essential resource for our approach.

\begin{figure*}[t!]
\centering
\includegraphics[width=\linewidth]{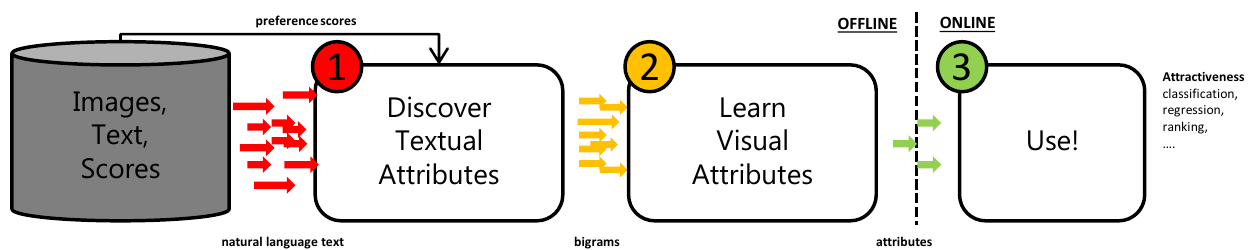}
\caption{We propose an end-to-end pipeline capable of learning visual attributes using information contained in the AVA dataset. 
Textual attributes are automatically discovered using textual comments and preference scores. Visual attribute models for these textual attributes are then learned using generic features extracted from images, as well as preference scores. See sections~\ref{sec:mining} and \ref{sec:learning} for further details.}
\label{fig:pipeline}
\end{figure*}

The main contributions of our proposed method are thus the following: 
\begin{enumerate}
\item An in-depth analysis of the AVA dataset, and in particular its textual comments and aesthetic preference scores.
\item A novel approach to aesthetic image analysis which combines the benefits of ``attribute-based'' and ``gen- eric'' techniques by 
(i) automatically discovering discriminative textual attributes using user comments and preference scores (step 1 in Fig.~\ref{fig:pipeline}); and 
(ii) supervised learning of detectable visual attributes using textual attributes and generic visual features (step 2 in Fig.~\ref{fig:pipeline}).
\item The application of the learned visual attributes to three different scenarios: 
aesthetic quality prediction, image classification and retrieval (step 3 in Fig.~\ref{fig:pipeline}).
\end{enumerate}

The remainder of this work is organized as follows: in section~\ref{sec:litreview} we review works related to aesthetic analysis and attributes. We then introduce
the AVA dataset and explain why it is an essential 
resource for aesthetic attribute learning (section \ref{sec:AVA}).
In section~\ref{sec:ava_analysis} we analyze aesthetic preference as expressed by real-valued scores while in section~\ref{sec:comments} we analyze aesthetic preference as expressed in textual comments. 
We then introduce the proposed approach to discover attributes that consists of 
(i) mining for discriminative textual attributes using the user comments and user scores (section \ref{sec:mining}) and 
(ii) learning visual attributes by modeling the visual appearance of textual attributes using generic visual features (section \ref{sec:learning}).
In section \ref{sec:apps}, we show practical applications of these attributes.

This paper extends our previous work \citep{murray12,marchesotti2013learning} with 1) a more detailed quantitative and qualitative analysis of the textual comments included in the AVA corpus, 2) an expanded quantitative evaluation of the textual features derived from these comments, 3) a quantitative assessment of the generalization performance of our learned visual attributes on a different image corpus and 4) an expanded image retrieval application to include joint attribute-semantic queries.

\section{Related work}\label{sec:litreview}

The study of aesthetics spans millennia, from the works of philosophers such Plato to those of researchers today in fields as diverse as neuroscience, psychology, and computer science \citep{sep-aesthetic-concept,leder2004model, chatterjee2011neuroaesthetics}. This highly inter-disciplinary interest in the topic is a natural outcome of the complex and multi-faceted nature of aesthetics, which is defined in the American Heritage\textsuperscript{\textregistered} Dictionary of the English Language \citep{Dictionary.com2012} as "the study of the mind and emotions in relation to the sense of beauty."

One major debate in the aesthetics research community surrounds the relative influence of subjective versus objective factors in aesthetic appreciation \citep{sep-aesthetic-concept}. This debate has been ongoing at least since Baumgarten argued that aesthetic appreciation was the result of objective reasoning \citep{hammermeister2002germana}, while David Hume and Edmund Burke\citep{sep-hume-aesthetics, sep-aesthetics-18th-british} took the opposing view that aesthetic appreciation was due to induced feelings.

For photography, which is the subject of our work, there are generally-accepted principles and techniques that are used by artists themselves to enhance the aesthetic quality of their artworks \citep{krages2005photography}. Examples include the ``rule of thirds" compositional rule and ``color harmony" guidelines \citep{krages2005photography,jacobson1948color}. Note that these and other guidelines may be applicable to other pictorial art-forms such as paintings, which nonetheless remain out of the scope of this work.

These principles and techniques may have arisen due to both objective and subjective/cultural factors. However, what is critical for data-driven image aesthetics analysis is that they are often {\em detectable} using machine learning techniques and training data. The computer vision community has used detectable principles and techniques in order to design systems that attempt to predict the {\em average response} of an observer when asked questions such as ``do you find this image aesthetically pleasing?", or ``how would you rate this image on a score of 1 to 10?".

As this work discovers attributes relevant for image aesthetics analsyis, we review the literature on aesthetics prediction and attributes.\\[2mm]
{\bf Computational image aesthetics prediction:}
As mentioned above, the computer vision community has in recent years developed data-driven approaches for analyzing pictorial artworks, particularly paintings and photographs.
Such approaches use standard machine learning techniques such as linear classifiers or regressors to predict aesthetic annotations.
Therefore the bulk of research effort has focused on designing appropriate visual features for representing image aesthetic characteristics.
In general, these features attempt to capture specific aesthetic principles and techniques related to composition and the use of color and light \citep{DJLW06,Ke2006,Luo2008,obrador2010role,dhar2011,Luo2011,joshi2011aesthetics,SanPedro2012,obrador2012towards}.


Datta's seminal work on aesthetic prediction extracted 56 visual features from an image and used these to train a statistical model to automatically classify an image as being of ``beautiful" or ``ugly" aesthetic quality \citep{DJLW06}. The features included relative color frequencies, mean pixel intensity, mean pixel saturation and mean pixel hue. Photographic rules of thumb such as the rule-of-thirds were also incorporated as well as other features related to texture, aspect ratio, and low depth-of-field.

There have been many other works in this line, such as that of Ke \etal \citep{Ke2006} who proposed features capturing the spatial distribution of edges, color, blur, and brightness. Luo \& Tang \citep{Luo2008} extracted semantic features describing lighting, color, and composition from the foreground image region after segmentation. Dhar \etal \citep{dhar2011} proposed the use of human-describable attributes related to composition, illumination and the image content.
In \citep{li2010towards}, face-specific aesthetic features such as individual face expressions, individual face poses, and between-face distances were captured and used to assess and improve portraiture.

As mentioned before, it is difficult to define an exhaustive list of aesthetics-relevant image descriptors. An alternative approach, proposed by \citep{Marches2011} is to use general-purpose image signatures to train aesthetics models. In this work the Bag-Of-Visual-words descriptor \citep{CD04,SZ03} and the Fisher Vector (FV, ~\citep{PD07,PSM10}), based on SIFT~\citep{Lowe2004} and color statistics features, were shown to achieve state-of-the-art aesthetic classification results. The authors posited that generic features are able to implicitly encode the aesthetic properties of an image. In addition, the spatial pyramid framework~\citep{LSP06} was able to roughly encode compositional information. Some recent works have complemented visual features with aethetic features mined from textual data \citep{SanPedro2012, geng2011}, by generating word frequency or TF-IDF vectors from comments given to images by individuals.

The promising results obtained by various aesthetics models have enabled the development of prototypes for not only assessing but also improving image aesthetics \citep{joshi2011aesthetics}. In particular, the web application ACQUINE \citep{DattaW10} allows one to upload images and receive a real-valued aesthetic score. Another such system, OSCAR \citep{Yao2012}, is a mobile application which provides on-line feedback to assist the user in improving an image's composition or colorfulness.\\[2mm]
{\bf Visual and textual attributes:}
There is a significant body of work on attribute learning in the computer vision and multimedia literature.
This is a cost-effective alternative to hand-listing attributes \citep{ferrari2008learning,lampert2009learning} 
and to architectures which require a human in the loop~\citep{parikh2011interactively}.
Existing solutions ~\citep{berg2010automatic,wang2009learning,yanai2005image} were typically developed for visual object recognition tasks. 
\cite{wang2009learning} proposes to mine pre-existing natural language resources. 
\cite{berg2010automatic} uses mutual information  to learn attributes relevant for e-commerce categories (handbags, shoes, earrings and ties).
\cite{duan2012discovering} uses latent CRF to discover detectable and discriminative attributes. 
\cite{donahue2011annotator} learned models for pre-determined nameable visual attributes
and applied them in scene and human attractiveness classification tasks.
Moreover, approaches such as \citep{rohrbach2010helps} use natural language text in the form of captions or surrounding image text. 
Only \cite{orendovici2010training} take into account text to devise aesthetic attributes, but the process is entirely manual. 

In contrast to the reviewed works which hand-pick aesthetic attributes, we aim to automatically discover them from textual data, with preference scores as supervisory information. We next describe the dataset, AVA, that we will leverage for training our models.

\section{AVA: A large-scale database for aesthetic visual analysis}\label{sec:AVA}

AVA (Aesthetic Visual Analysis) is a publicly available database for aesthetics analysis which we recently introduced in \citep{murray12}. 
In what follows, we first compare AVA to related databases,
and describe their limitations for our goal of automatic discovery of mid-level image representations for aesthetic analysis.
We then provide a detailed analysis of AVA, focusing on 3 key components:
(i) its images;
(ii) its real-valued score annotations; and
(iii) its textual comments.

\subsection{AVA and Related Databases}\label{subsec:other_dbs}

In addition to AVA, there exist several public image databases in current use which contain aesthetic annotations. In this section, we compare the properties of these databases to those of AVA and discuss the features that differentiate AVA from such databases. A summary of this comparison is shown in Table~\ref{table:comparison}.

\begin{figure}[!t]
\centering
\includegraphics[width=0.9\linewidth]{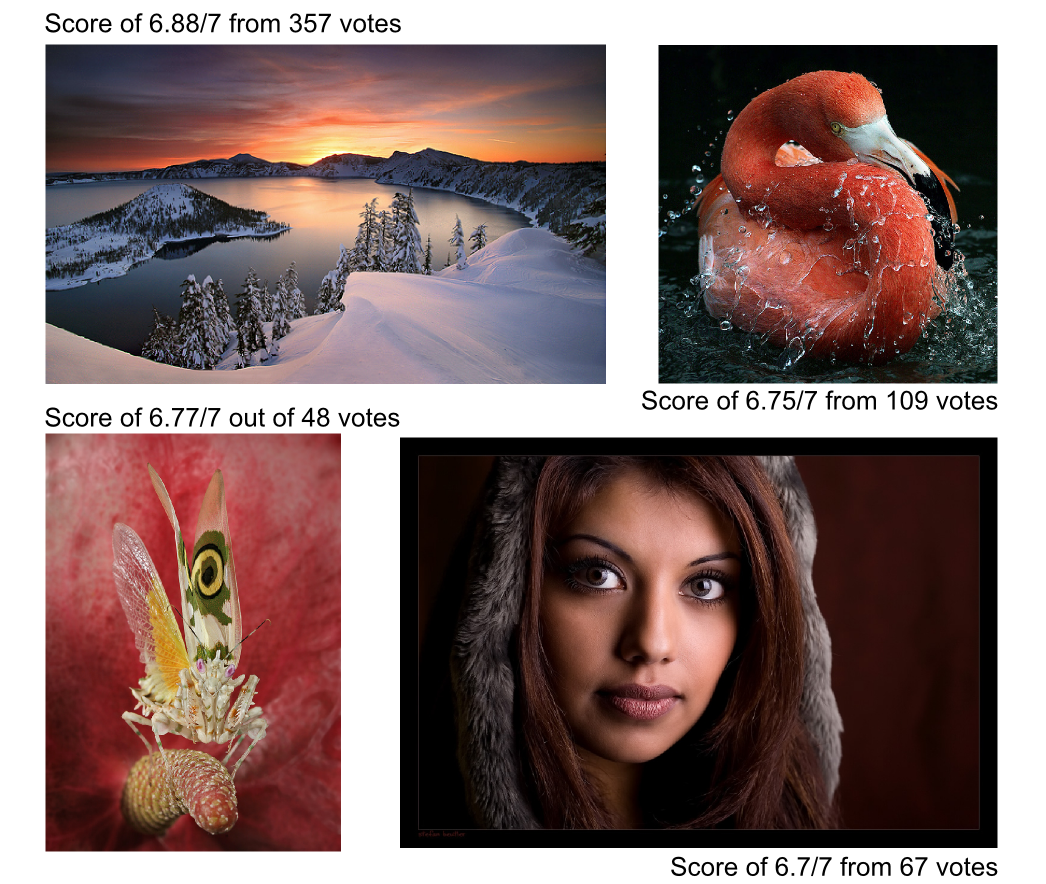}
\caption{Photos highly rated by peer voting in an on-line
  photo sharing community ({\em photo.net}).}
\label{fig:eyecatching}
\end{figure}

\noindent {\bf Photo.net, PN \citep{DJLW06}}: PN contains 3,581 images from the social network {\tt Photo.net}. In this online community, members are instructed to give two scores from 1 to 7 for an image. One score corresponds to the image's aesthetics and the other to the image's originality. The dataset includes the mean aesthetic score and the mean originality score for each image. As described in \citep{DJLW06}, the aesthetic and originality scores are highly correlated, with little disparity between these two scores for a given image. This is probably due to the difficulty of separating these two characteristics of an image. As the two scores are therefore virtually interchangeable, works using PN have restricted their analysis to the aesthetic scores. Figure \ref{fig:eyecatching} shows sample photos of high quality with their scores and number of votes.

\begin{table}[!t]
\scriptsize
\begin{tabularx}{1\linewidth}{|X|c|c|c|c|c|}
\hline
& AVA & PN & CUHK & CUHKPQ & CLEF \\\hline
\hline
Large scale & \color{green}{Y} & \color{red}{N} & \color{red}{N} & \color{red}{N} & \color{green}{Y} \\
\hline
Score distr. & \color{green}{Y} & \color{green}{Y} & \color{red}{N} & \color{red}{N} & \color{red}{N} \\
\hline
Rich \hspace{15mm} annotations & \color{green}{Y} & \color{red}{N} & \color{green}{Y} & \color{green}{Y} & \color{green}{Y} \\
\hline
Semantic labels & \color{green}{Y} & \color{red}{N} & \color{red}{N} & \color{green}{Y} & \color{green}{Y} \\
\hline
Style labels & \color{green}{Y} & \color{red}{N} & \color{red}{N} & \color{red}{N} & \color{green}{Y} \\
\hline
\end{tabularx}
\caption{Comparison of the properties of current databases containing aesthetic annotations. AVA is large-scale and contains score distributions, rich annotations, and semantic and style labels.}
\label{table:comparison}
\end{table}

\begin{figure}[!t]
\centering
\includegraphics[height=1.5cm]{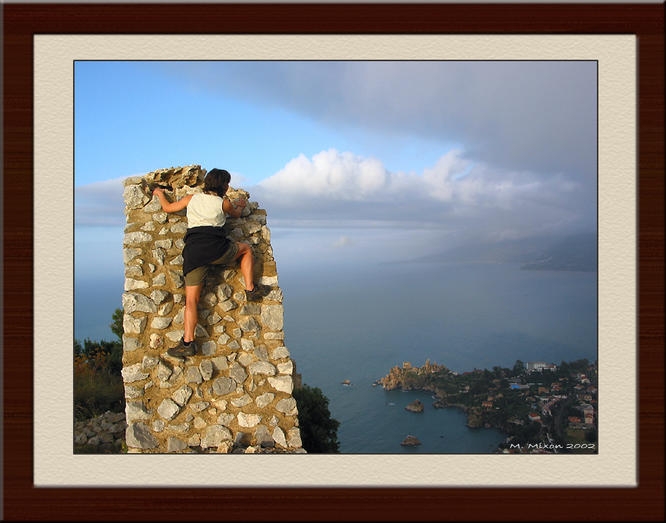}
\includegraphics[height=1.5cm]{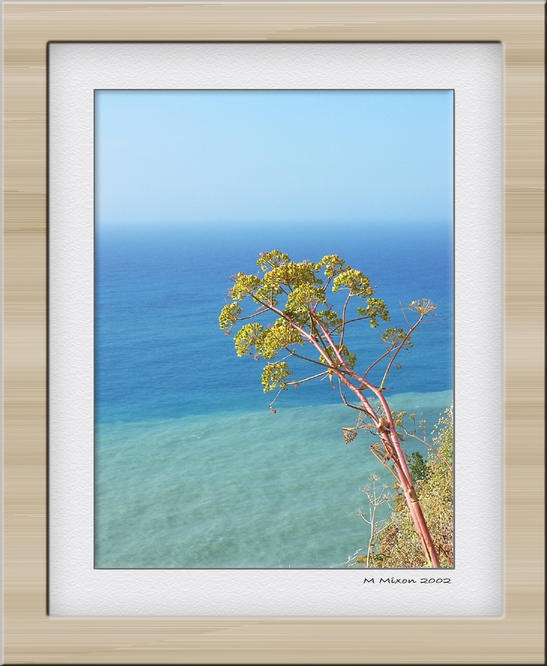}
\includegraphics[height=1.5cm]{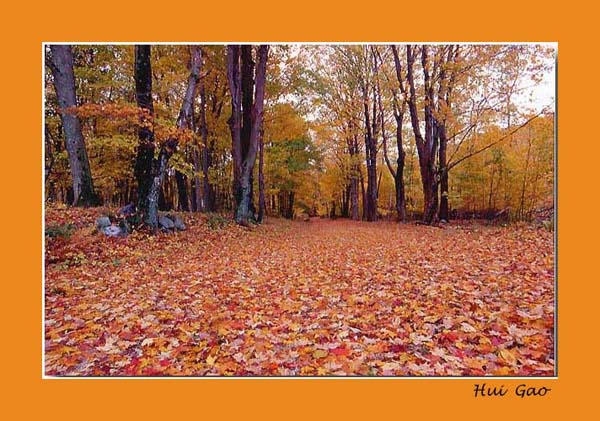}
\includegraphics[height=1.5cm]{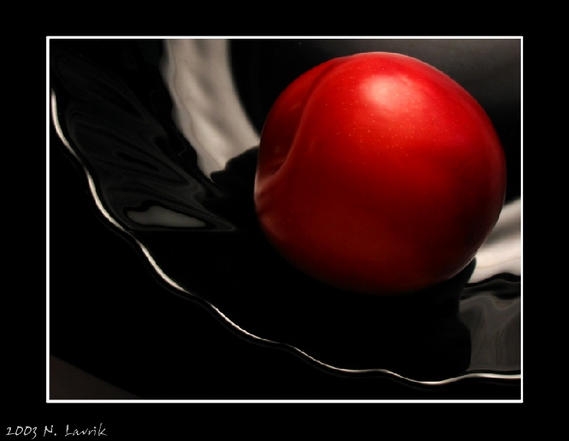}
 \caption{Sample images from PN with borders manually created by photographers to enhance the photo visual appearance.}
\label{fig:frames}
\end{figure}

Upon visual inspection of PN, we have noticed a correlation between images receiving a high grade and the presence of frames manually created by the owners to enhance the visual appearance (see examples in Figure \ref{fig:frames}). In fact, we manually detected that more than 30\% of the images are framed.

In addition to this bias, many images in PN have been scored by very few users. In fact, the images were included on the condition that they had received scores from at least two users. In contrast, each image included in AVA has at least 78 votes. In addition, AVA contains approximately 70 times as many images.\\[2mm]
\noindent {\bf CUHK \citep{Ke2006}}: 
CUHK contains 12,000 images, half of which are considered high quality and the rest labeled as low quality. \citep{Ke2006} observed the same bias for images with border as we did for PN, so they removed all the frames from the images they released. The images were obtained by retaining the top and bottom 10\% (in terms of mean scores) of 60,000 images randomly crawled from \texttt{www.dpchallenge.com}. Our dataset differs from CUHK in several ways. While AVA includes more ambiguous images, CUHK only contains images with a very clear consensus on their score. As a consequence, the images in CUHK are much less representative of the range of images, in terms of aesthetic quality, that one would find in a real-world application such as re-ranking images returned by a search on the web. In addition, CUHK is no longer a challenging dataset for classification; recent methods achieved accuracies superior to 90\% on this dataset \citep{Marches2011}. Finally, CUHK provides only binary labels (1=high quality images, 0=low quality images) whereas AVA provides an entire distribution of scores for each image.\\[2mm]
\noindent {\bf CUHKPQ \citep{Luo2011}}: CUHKPQ consists of 17,690 images obtained from a variety of on-line communities and divided into 7 semantic categories. Each image was labeled as either high or low quality by at least 8 out of 10 independent viewers. Therefore this dataset consists of very high consensus images and their binary labels. Like CUHK, it is not a challenging dataset for the problem of binary classification: the method of \citep{Luo2011} obtained Area under the ROC curve (AROC) values between 0.89 and 0.95 for all semantic categories. Also like CUHK, the images in the dataset do not span the full range of images, in terms of aesthetic quality, that one is likely to find in a real-world aesthetic prediction application. In addition, despite the fact that AVA shares similar semantic annotations, it differs in terms of scale and also in terms of consistency. In fact, CUHKPQ was created by mixing high quality images derived from photographic communities and low quality images provided by university students.\\[2mm]
\noindent {\bf MIRFLICKR/Image CLEF: Visual Concept Detection and Annotation Task 2011 \citep{clef}}: MIRFLICKR is a large dataset introduced in the community of multimedia retrieval. It contains 1 million images collected from Flickr, along with textual tags, aesthetic annotations (Flickr's interestingness flag) and EXIF meta-data. A sub-part of MIRFLICKR was used by CLEF (the Cross-Language Evaluation Forum) to organize two challenges on ``Visual Concept Detection''. For these challenges, the basic annotations were enriched with emotional annotations and with some tags related to photographic style. It is probably the dataset closest to AVA but it lacks rich aesthetic preference annotations. In fact, only the ``interestingness'' flag is available to describe aesthetic preference. Some of the 44 visual concepts available might be related to AVA photographic styles but they focus on two very specific aspects: exposure and blur. Only the following categories are available: neutral illumination, over-exposed, under-exposed, motion blur, no blur, out of focus, and partially blurred. In addition, the number of images with such style annotations is limited.

\subsection{AVA and its annotations}
\begin{figure}[!t]
 \centering
\includegraphics[width=\linewidth]{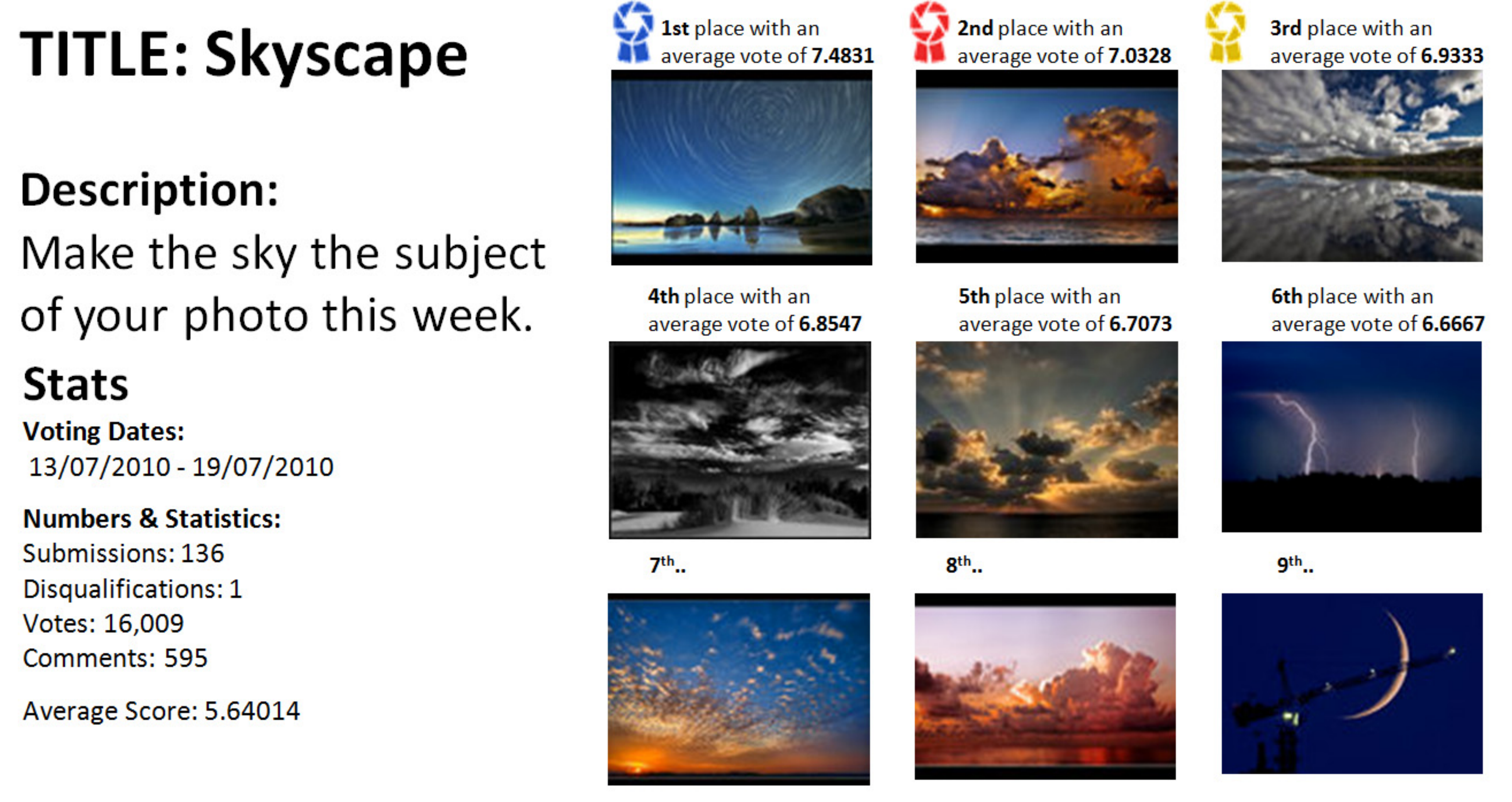}       
\caption{A sample challenge entitled ``Skyscape'' from the social network \texttt{www.dpchallenge.com}. Users submit images that should conform to the challenge description and be of high aesthetic quality. The submitted images are rated by members of the social network during a finite score period. After this period, the images are ranked by their average scores and the top three images are awarded ribbons.}
\label{fig:chall_eg}
\end{figure}

AVA contains photographic images and a rich variety of associated meta-data, derived from \texttt{www.dpchallenge} \texttt{.com}. To our knowledge, AVA represents the first attempt to create a large database containing a unique combination of heterogeneous annotations. The peculiarity of this database is that it is derived from a community where images are uploaded and scored in response to photographic challenges. Each challenge is defined by a title and a short description (see Fig.~\ref{fig:chall_eg} for a sample challenge).

\begin{figure}[!t]
 \centering
\includegraphics[width=\linewidth]{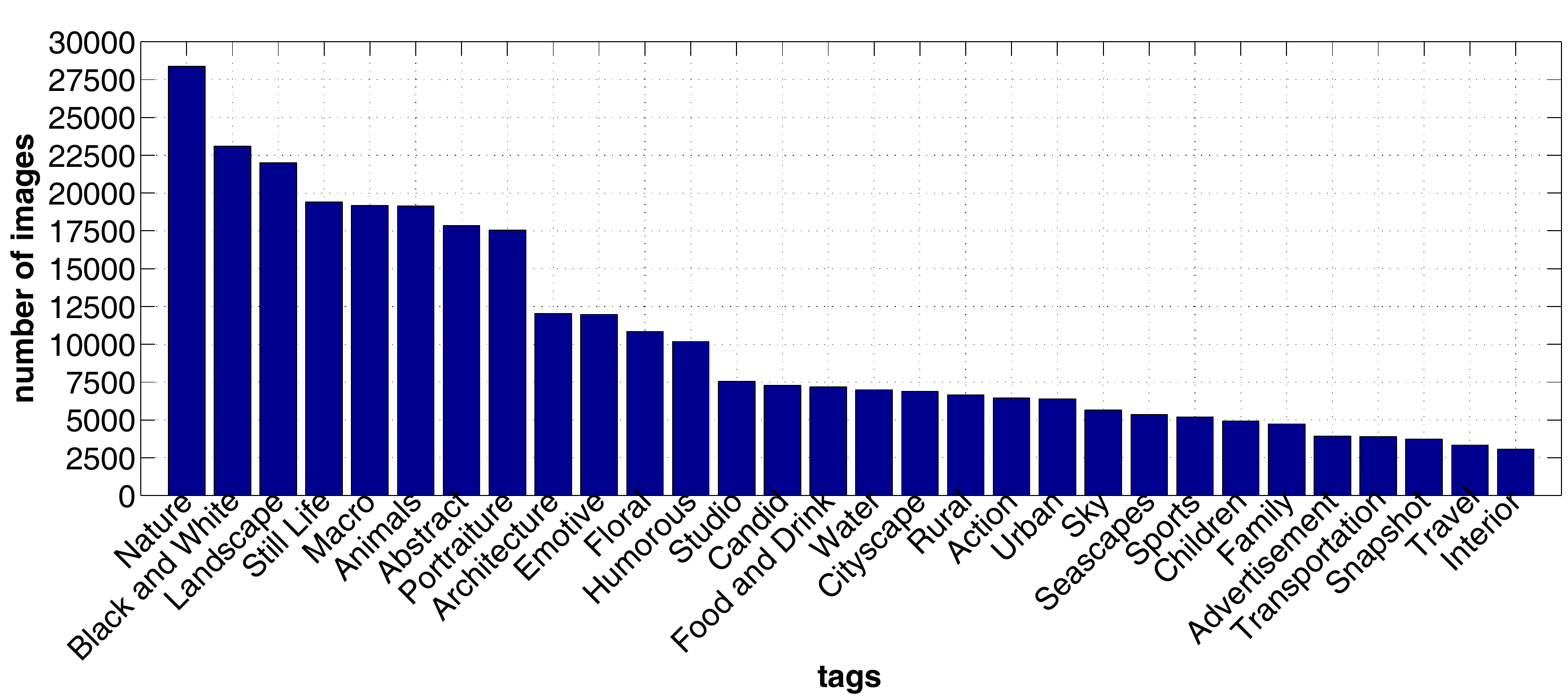}           
\caption{Frequency of the 30 most common semantic tags in AVA. The tags cover a wide range of content and styles. The most popular content-related tags are {\em nature} and {\em landscape}, while the most popular styles are {\em black and white} and {\em macro}.}
\label{fig:tag_hist}
\end{figure}

Using this interesting characteristic, we associated each image in AVA with the information of its corresponding challenge. This information can be exploited in combination with aesthetic scores or semantic tags to gain an understanding of the context in which such annotations were provided.
We created AVA by collecting approximately 255,000 images covering a wide variety of subjects on 1,447 challenges. We combined the challenges with identical titles and descriptions and we reduced them to 963. Each image is associated with a single challenge.\\[2mm]
In AVA we provide three types of annotations:\\[2mm]
{\bf Aesthetic annotations}: Each image is associated with a {\em distribution of scores}
which correspond to individual votes.
The number of votes per image ranges from 78 to 549, with an average of 210 votes. 
Such score distributions represent a gold mine of aesthetic judgments generated by hundreds of amateur and professional photographers with a practiced eye.
In addition, AVA contains rich {\em textual comments} given to users by other community members.
We believe that such annotations have a high intrinsic value because they capture the way hobbyists and professionals understand visual aesthetics.\\[2mm]
{\bf Semantic annotations}: We provide 66 textual tags describing the semantics of the images. Approximately 200,000 images contain at least one tag, and 150,000 images contain 2 tags. The frequency of the most common tags in the database can be observed in Fig.~\ref{fig:tag_hist}.\\[2mm]
{\bf Photographic style annotations}: Despite the lack of a formal definition, we understand photographic style as a consistent manner of shooting photographs achieved by manipulating camera configurations (such as shutter speed, exposure, or ISO level). 
We manually selected  72 Challenges corresponding to photographic styles and we identified three broad categories according to  a popular photography manual \citep{KODAK82}: {\em Light}, {\em Color}, {\em Composition}. 
We then merged similar challenges (\eg ``Duotones'' and ``Black \& White'') and we associated each style with one category. The 14 resulting photographic styles along with the number of associated images are: Complementary Colors (949), Duotones (1,301), High Dynamic Range (396), Image Grain (840), Light on White (1,199),  Long Exposure (845), Macro (1,698), Motion Blur (609), Negative Image (959), Rule of Thirds (1,031), Shallow DOF (710), Silhouettes (1,389), Soft Focus (1,479), Vanishing Point (674).

In the next two sections we focus on the key AVA annotations necessary for our goal of learning aesthetic attributes, namely the score distributions and textual comments.

\subsection{Aesthetic preference as real-valued scores}\label{sec:ava_analysis}
\begin{figure*}[!t]
  \centering
  \subfigure[]{\includegraphics[width=0.24\linewidth]{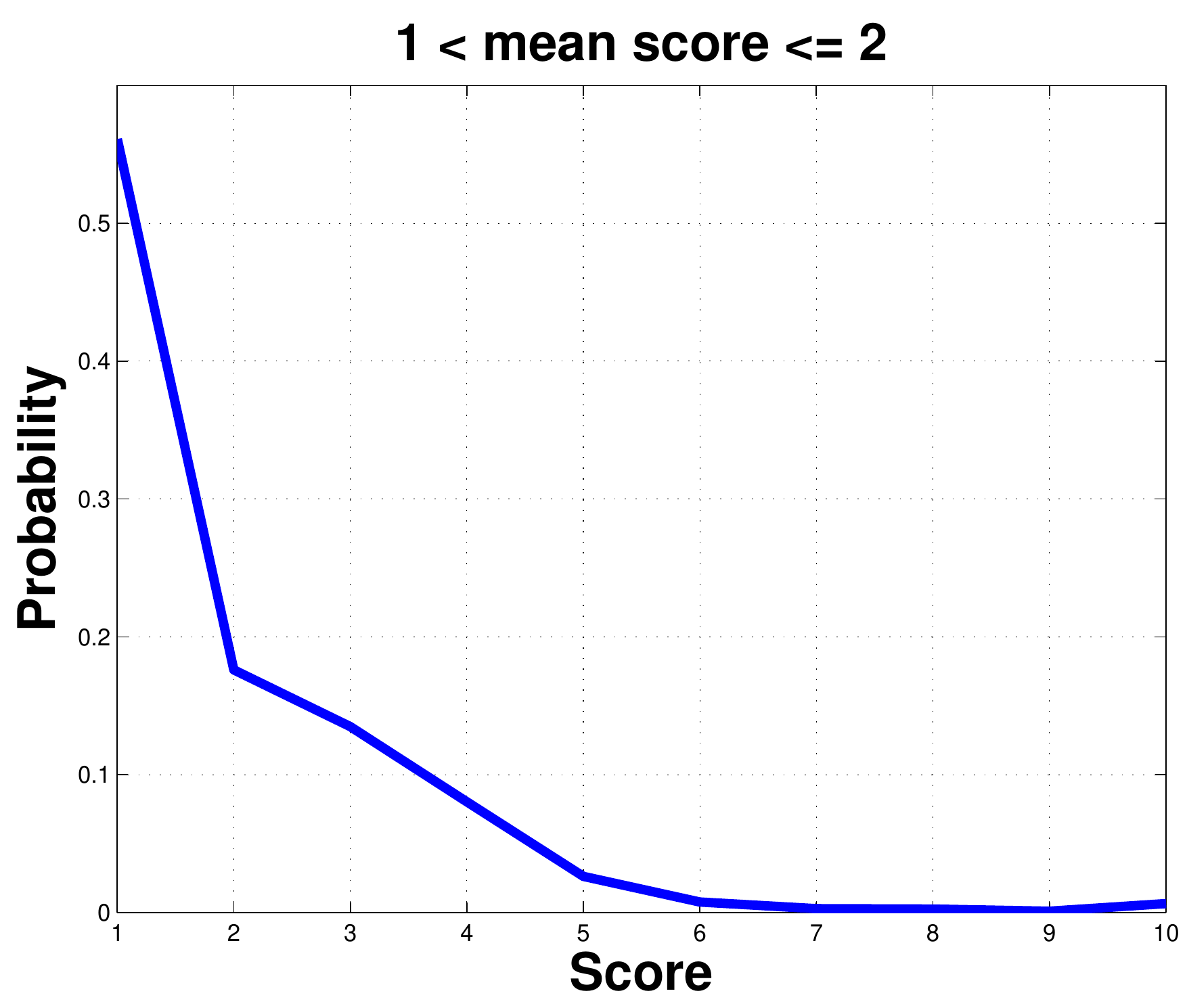}\label{fig:distro_clusters1}}
  \subfigure[]{\includegraphics[width=0.24\linewidth]{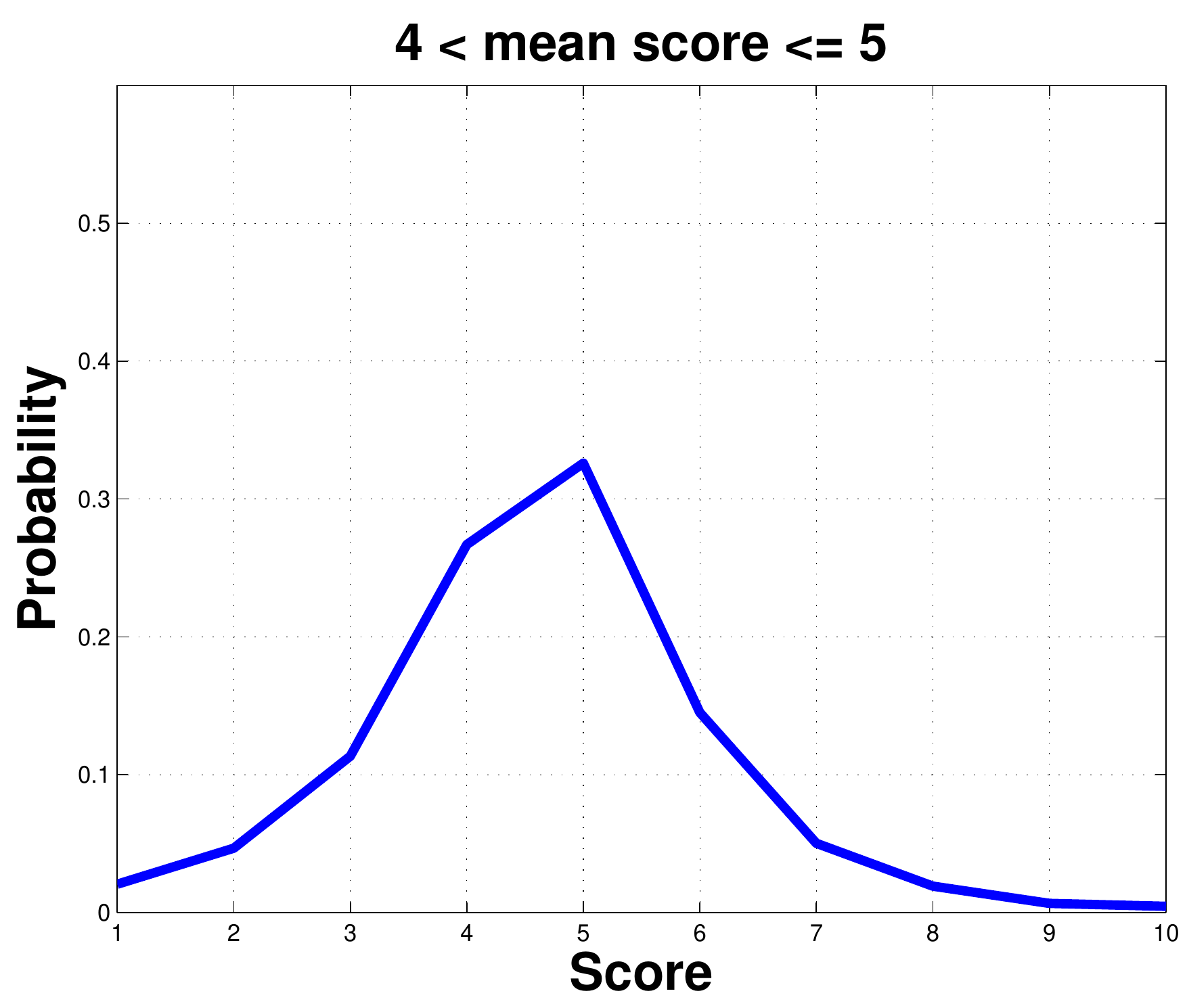}\label{fig:distro_clusters2}}
  \subfigure[]{\includegraphics[width=0.24\linewidth]{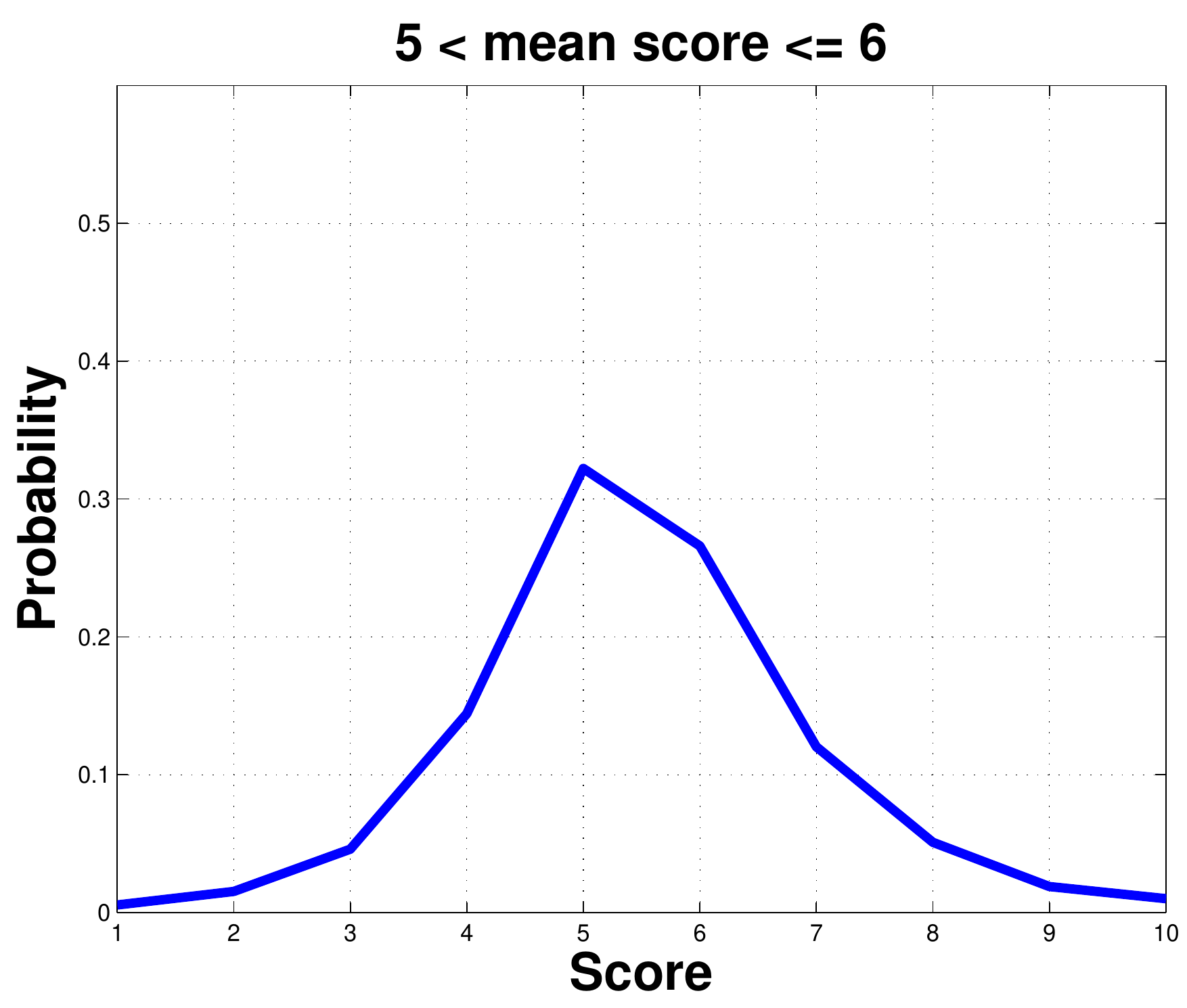}\label{fig:distro_clusters3}}
  \subfigure[]{\includegraphics[width=0.24\linewidth]{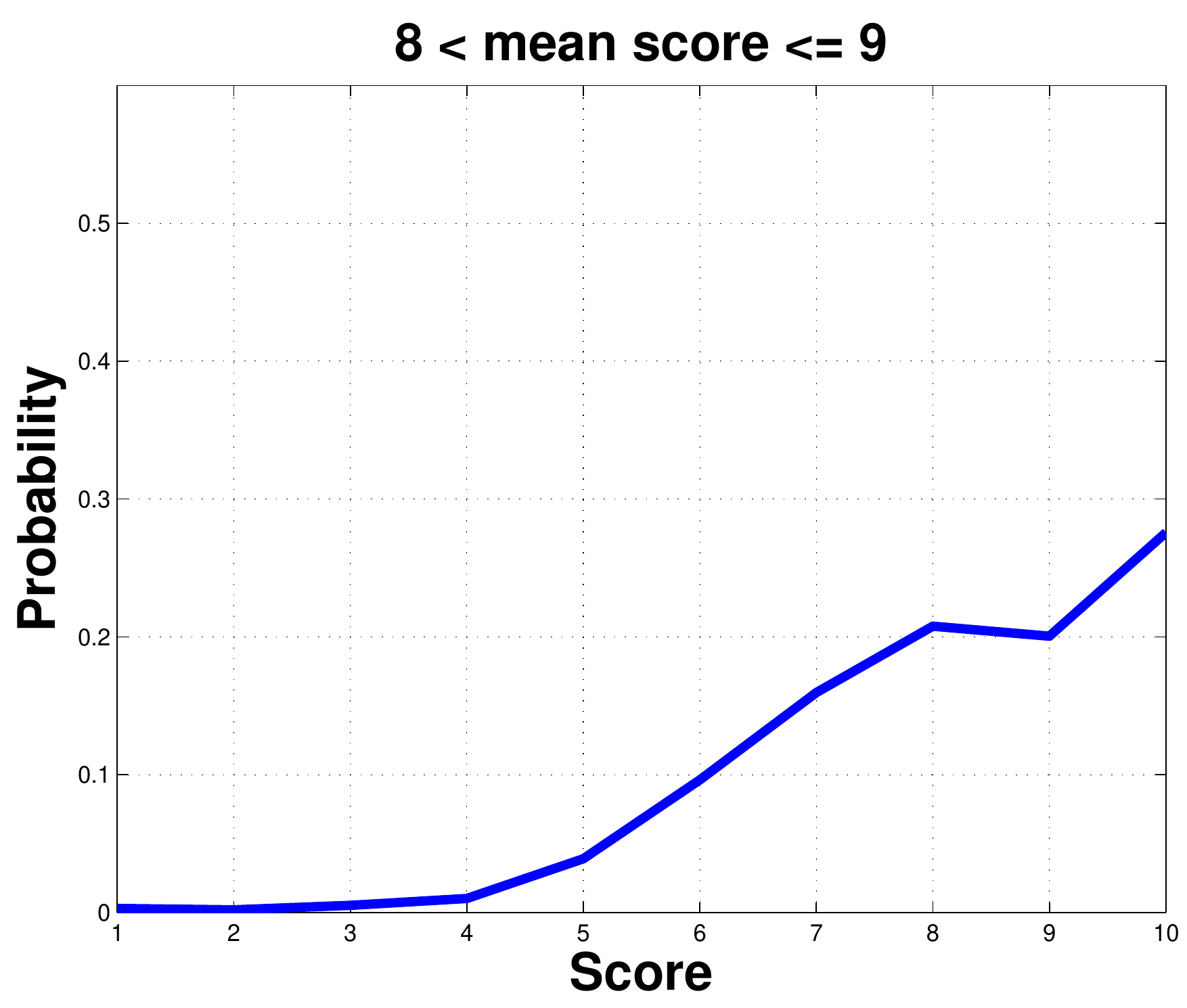}\label{fig:distro_clusters4}}
\caption{Averaged distributions for images with different mean scores. Distributions with mean scores close to the mid-point of the score scale tend to be Gaussian, with highly-skewed distributions appearing at the end-points of the scale.}
\label{fig:distro_clusters}
\end{figure*}

Annotations of aesthetic preference are typically in the form of real-valued scores. When multiple scores are given to an image, as it is the case with images derived from social networking sites like \texttt{www.dpchallenge.com}, a score distribution is formed. In this section, we analyze the rich score distributions (consisting on average of approximately 200 scores) available in AVA in order to gain a deeper understanding of such distributions and of what kind of information can be deduced from them.\\[2mm]
\noindent {\bf Score distributions are largely Gaussian}. 
Table~\ref{tab:distro_gof} shows a comparison of Goodness-of-Fit (GoF), as measured by RMSE, between top performing distributions we used to  model the score distributions of AVA. 
One sees that Gaussian functions perform adequately for images with mean scores between 2 and 8, which constitute 99.77\% of all the images in the dataset. In fact, the RMSEs for Gaussian models are rarely higher than 0.06. This is illustrated in Fig.~\ref{fig:distro_clusters}. Each plot shows a density function obtained by averaging the score distributions of images whose mean score lies within a specified range. The averaged score distributions are usually well approximated by Gaussian functions (see Figures~\ref{fig:distro_clusters2} and \ref{fig:distro_clusters3}). We also fitted Gaussian Mixture Models with three Gaussians to the distributions but we only found minor improvement with respect to one Gaussian. Beta, Weibull and Generalized Extreme Value distributions were also fitted to the score distributions, but gave poor RMSE results.

\begin{figure}[!t]
 \centering
\includegraphics[width=0.7\linewidth]{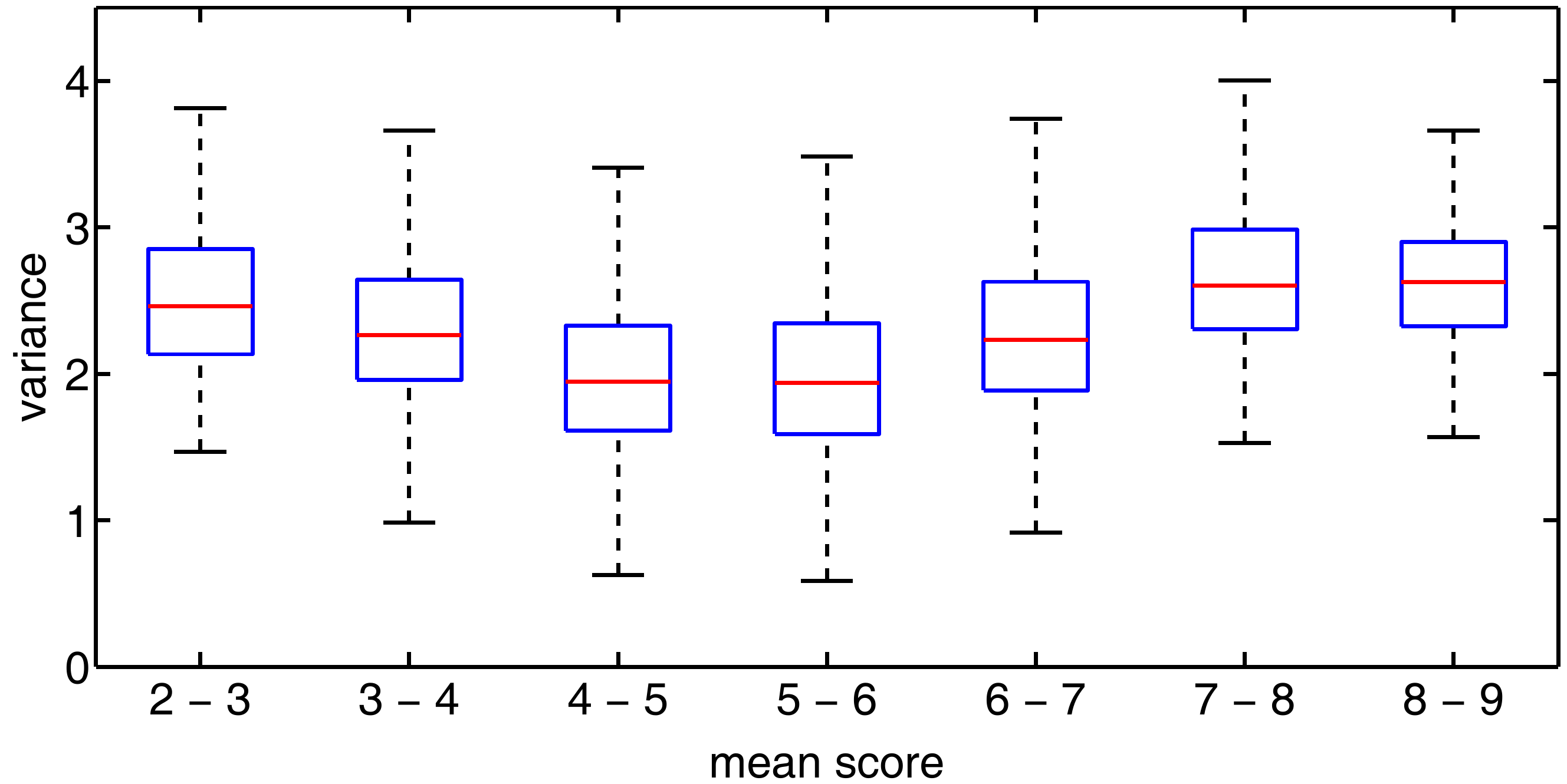}      
\caption{Distributions of variances of score distributions, for images with different mean scores. The variance tends to increase with the distance between the mean score and the mid-point of the score scale.}
\label{fig:var_w_mean}
\end{figure}

\begin{figure}[!t]
  \setlength\fboxsep{-0mm}
  \centering
  \begin{tabular}{m{0.01\textwidth}m{0.98\textwidth}}
  \begin{sideways}
{\small low variance}
  \end{sideways} &
\fcolorbox{white}{white}{\includegraphics[height=0.115\textwidth]{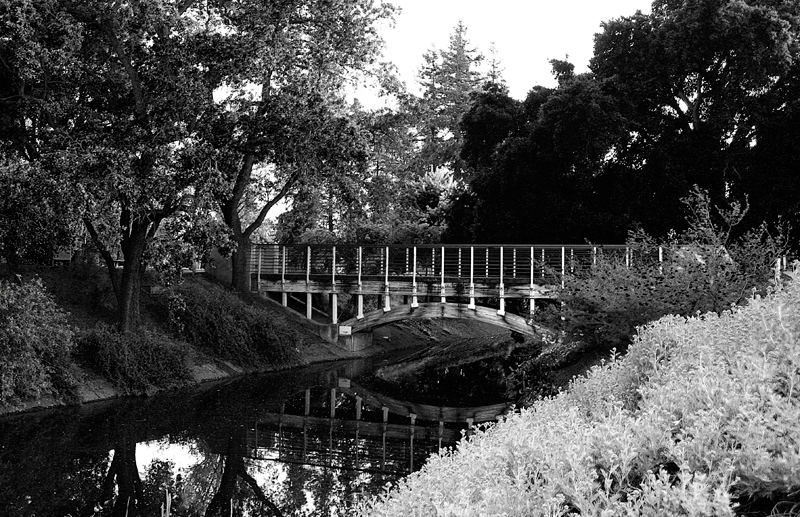}} 
\fcolorbox{white}{white}{\includegraphics[height=0.115\textwidth]{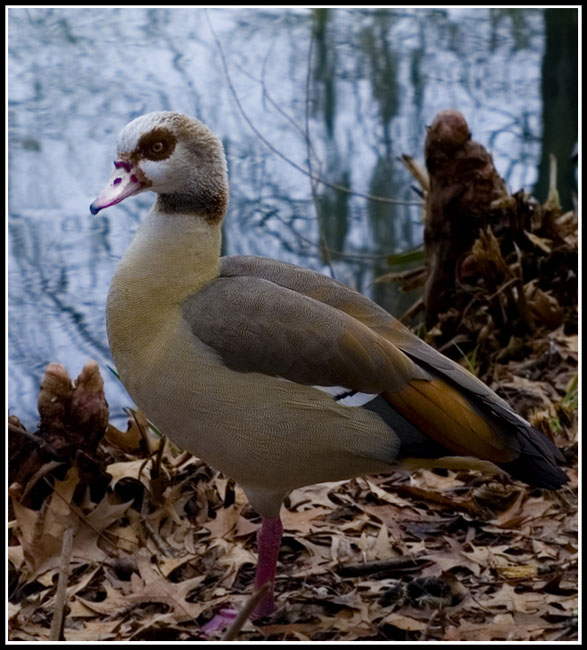}} 
\fcolorbox{white}{white}{\includegraphics[height=0.115\textwidth]{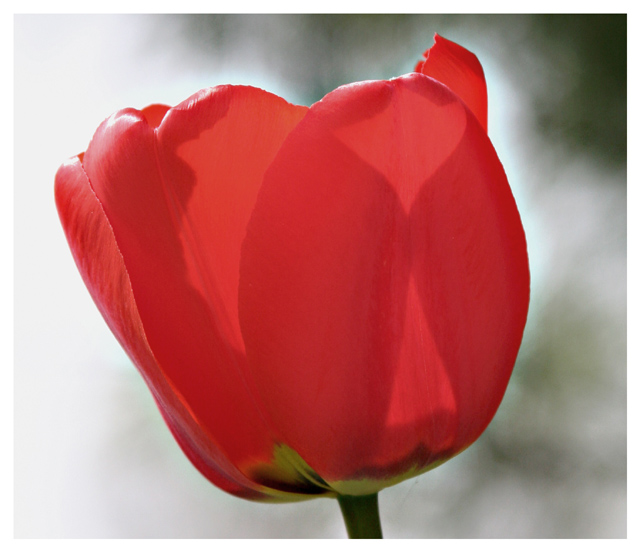}}\\[2mm]
  \cline{1-1} 
 \begin{sideways}
{\small high variance}
 \end{sideways}&
\fcolorbox{white}{white}{\includegraphics[height=0.105\textwidth]{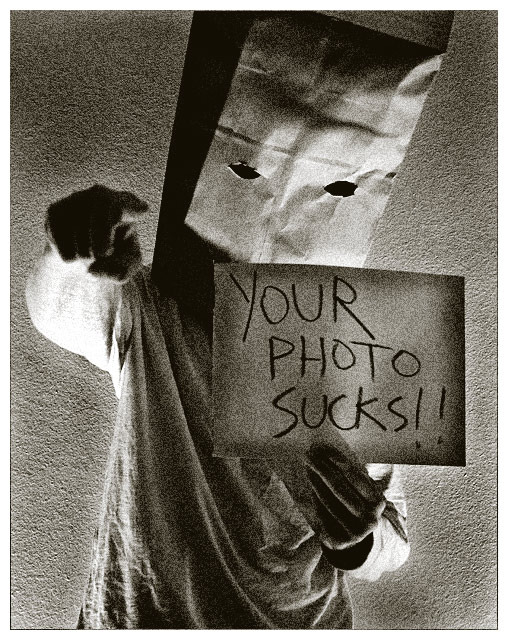}}
\fcolorbox{white}{white}{\includegraphics[height=0.105\textwidth]{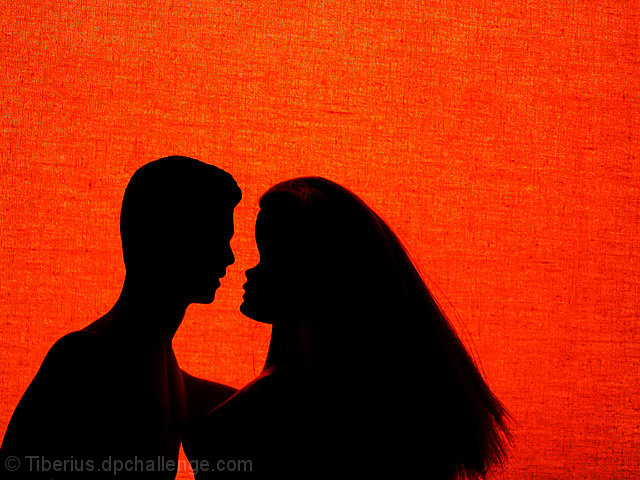}}
\fcolorbox{white}{white}{\includegraphics[height=0.105\textwidth]{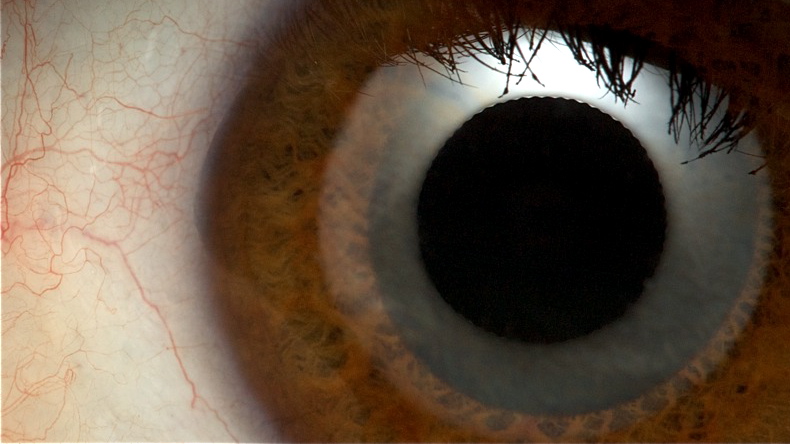}}\\[2mm]
  \end{tabular}
  \caption{Examples of images with mean scores around 5 but with different score variances. High-variance images have non-conventional styles or subjects.}
  \label{fig:qualitative_var_w_mean}
\end{figure}

Non-Gaussian distributions tend to be highly-skewed. This skew can be attributed to a floor and ceiling effect \citep{Cramer2004}, occurring at the low and high extremes of the score scale. This can be observed in Figures~\ref{fig:distro_clusters1} and \ref{fig:distro_clusters4}. Images with positively-skewed distributions are better modeled by a Gamma distribution $\Gamma(s)$, which may also model negatively-skewed distributions using the transformation $\Gamma'(s)=\Gamma((s_{min} + s_{max}) - s)$, where $s_{min}$ and $s_{max}$ are the minimum and maximum scores of the score scale.\\[2mm]
\begin{table}[!t]
\centering
\small
\begin{tabular}{p{2.5cm}ccc}
\toprule \textbf{Mean score} & \multicolumn{3}{c}{\textbf{Average RMSE}} \\
& Gaussian & $\Gamma$ & $\Gamma'$\\
\midrule 1-2 & 0.1138 & {\bf 0.0717} & 0.1249\\
\midrule 2-3 & 0.0579 & {\bf 0.0460} & 0.0633\\
3-4 & {\bf 0.0279} & 0.0444 & 0.0325\\
4-5 & {\bf 0.0291} & 0.0412 & 0.0389\\
5-6 & {\bf 0.0288} & 0.0321 & 0.0445\\
6-7 & 0.0260 & {\bf 0.0250} & 0.0455\\
7-8 & {\bf 0.0268} & 0.0273 & 0.0424\\
8-9 & 0.0532 & 0.0591 & {\bf 0.0403}\\
\midrule Average RMSE & {\bf 0.0284} & 0.0335 & 0.0429\\
\bottomrule\\
\end{tabular}
\caption{Goodness-of-Fit per distribution with respect to mean score: the last row shows the average RMSE for all images in the dataset. The Gaussian distribution was the best-performing model for 62\% of images in AVA.}
\label{tab:distro_gof}
\end{table}
{\bf Standard Deviation is a function of mean score}.
Box-plots of the variance of scores for images with mean scores within a specified range are shown in Fig.~\ref{fig:var_w_mean}. It can be seen that images with ``average'' scores (scores around 4, 5 and 6) tend to have a lower variance than images with scores greater than 6.6 or less than 4.5. Indeed, the closer the mean score gets to the extreme scores of 1 or 10, the higher the probability of a greater variance in the scores. This is likely due to the non-Gaussian nature of score distributions at the extremes of the score scale.\\[2mm]
\noindent {\bf Images with high variance are often non-conven-} {\bf tional}. To gain an understanding of the additional information a distribution of scores may provide, we performed a qualitative inspection of images with low and high variance. Table~\ref{tab:mv_matrix} displays our findings. The styles and photographic techniques employed to shoot seem to correlate with the mean score photographs receive. For a given mean value however, images with a high variance seem more likely to be edgy or subject to interpretation, while images with a low variance tend to use conventional styles or depict conventional subject matter. This is consistent with our intuition that an innovative application of photographic techniques and/or a creative interpretation of a challenge description is more likely to result in a divergence of opinion among voters. Examples of images with low and high score variances are shown in Fig.~\ref{fig:qualitative_var_w_mean}. The bottom-left photo in particular, submitted to the challenge ``Faceless'', had an average score of 5.46 but a very high variance of 5.27. The comments it received indicate that while many voters found the photo humorous, others may have found it rude.

\begin{table}[!t]\small
\centering
\begin{tabular}{p{0.25cm}p{0.25cm}|p{2.5cm}|p{3cm}|}
\cline{3-4}
& & \multicolumn{2}{c|}{{\bf variance}} \\
\cline{3-4}
& & low & high \\
\cline{1-4}
\multicolumn{1}{|c|}{\multirow{2}{*}{{\bf mean}}} &
\multicolumn{1}{c|}{low} & poor, conventional technique and/or subject matter & poor, non-conventional technique and/or subject matter\\
\cline{2-4}
\multicolumn{1}{|c|}{} &
\multicolumn{1}{c|}{high} &good, conventional technique and/or subject matter & good, non-conventional technique and/or subject matter\\
\cline{1-4}\\
\end{tabular}
\caption{By qualitatively inspecting images with different means and variances, we identified 4 categories of images with shared patterns, common quality features and subjects.}
\label{tab:mv_matrix}
\end{table}

\begin{figure}[!t]
 \centering
\includegraphics[width=\linewidth]{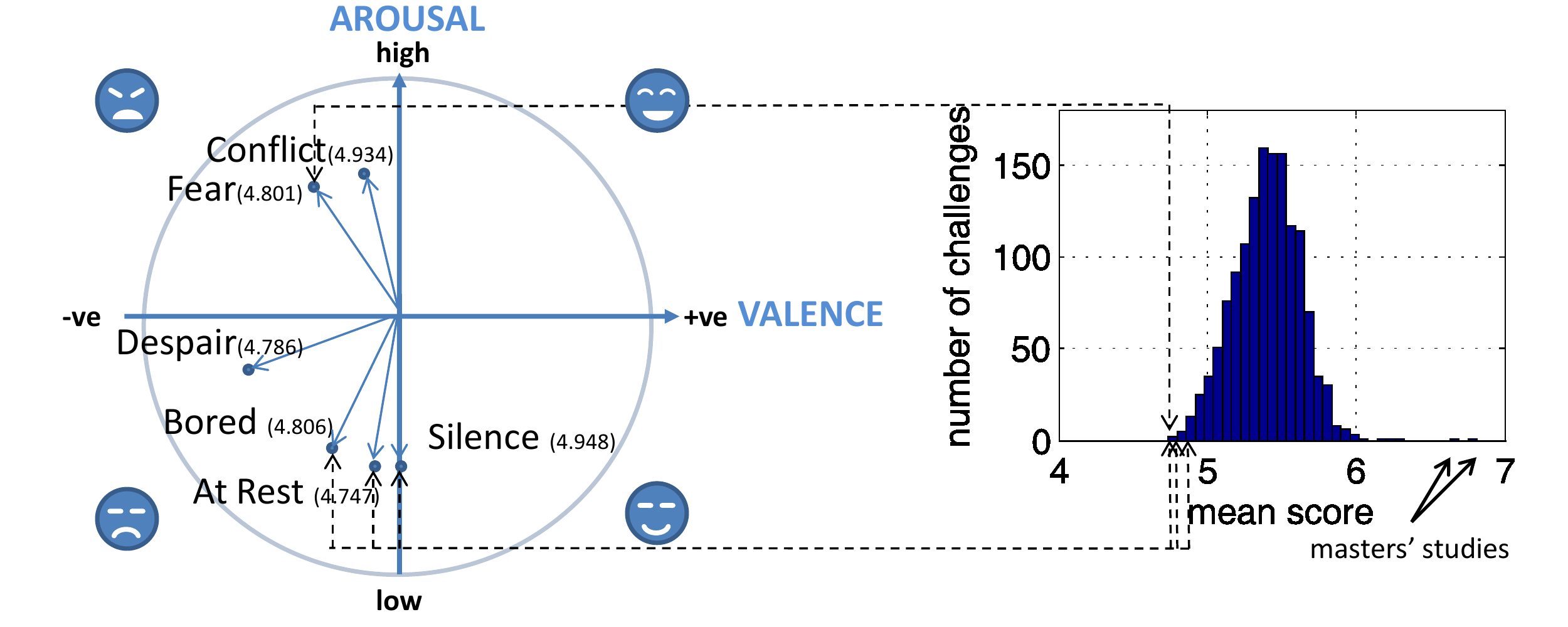}

\caption{Challenges with a lower-than-normal average vote are often in the left quadrants of the  arousal-valence plane. The two outliers on the right are masters' studies challenges.}
\label{fig:challs_mean_score}
\end{figure}

\noindent\textit{Semantic content and aesthetic preference}\\[2mm]~
We evaluated aggregated statistics for each challenge using the score distributions of the images that were submitted.
Fig.~\ref{fig:challs_mean_score} shows a histogram of the mean score of all challenges. 
As expected, the mean scores are approximately normally distributed around the mid-point of the score scale. 
We inspected the titles and associated descriptions of the challenges at the two extremes of this distribution. 
We did not observe any semantic coherence between the challenges in the right-most part of the distribution. 
However, it is worth noticing that two ``masters' studies'' (where only members who have won awards in previous challenges are allowed to participate) were among the top 5 scoring challenges.
We use the arousal-valence emotional plane \citep{russell1980circumplex} to plot the challenges on the left of the distribution (the low-scoring tail).  The dimension of valence ranges from highly positive to highly negative, whereas the dimension of arousal ranges from passive to active.
In particular, among the lowest-scoring challenges we identified: \#1 ``At Rest'' (av. vote= 4.747), 
\#2 ``Despair'' (av. vote=4.786), 
\#3 ``Fear'' (av.vote=4.801), 
\#4 ``Bored'' (av. vote=4.8060), 
\# 6 ``Pain'' (av. vote=4.818), 
\#23 ``Conflict'' (av. vote= 4.934), 
\#25 ``Silence'' (av. vote= 4.948), 
\#30 ``Shadows'' (av. vote= 4.953), 
\#32 ``Waiting'' (av. vote.=4.953), 
\#39 ``Obsolete'' (av.vote= 4.9740). In each case, the photographers were instructed to depict or interpret the emotion or concept of the challenge's title. This suggests that themes in the left quadrants of the arousal-valence plane (see Fig.~\ref{fig:challs_mean_score}) bias the aesthetic judgments towards lower scores. 

We investigated the relationship between the title and description of a challenge and the mean of the variance of the score distributions of images submitted to that challenge. We found that the majority of free study challenges were among the bottom 100 challenges by variance, with 11 free studies among the bottom 20 challenges. Free study challenges have no restrictions or requirements as to the subject matter of the submitted photographs. The low variance of these types of challenges suggests that challenges with specific requirements tend to lead to a greater variance of opinion, probably with respect to how well entries adhere to these requirements.

\subsection{Aesthetic preference as textual comments}\label{sec:comments}
Of the 255,530 images in AVA, most of them (253,903) received at least one comment from a member of the social network. There are two phases in which comments may be given. In the first phase, the challenge is ongoing and the comments and votes given to images are not yet visible to the community. In this phase, a user is allowed to give a comment to an image after giving that image a score. Comments given in this phase should therefore be unbiased with respect to the opinions of other members. In the second phase, the challenge has been completed and the results are public. Comments given in this phase are therefore likely to be biased in at least two ways. First, images which performed well during the challenge are likely to have a greater number of comments as they are more visible, being high in the rankings for that challenge. Second, the comments given to an image in this period may be influenced by the results of the challenge and the comments it has already received.

The guidelines for commenting\footnote{http://www.dpchallenge.com/help\_faq.php\#howcomments} encourage the users to leave comments when voting and to include advice for improving the work. As such, comments typically express the member's opinion on the quality of the photograph, their justifications for giving a certain score, as well as critiques of the strengths and weaknesses of the photograph. For example, the top right image in Fig.~\ref{fig:qualitative_var_w_mean} received the following comment:

\begin{myQuote}
"Like the shot. One thing I think it could be helped by is a bit more contrast, make the colors more rich and stand out that much more. I like the [square] crop...good choice."
\end{myQuote}

These comments are a rich source of information about the {\em reasons} for which an individual may assign a particular aesthetic score to an image.\\

We investigated several properties of the comments given to images in AVA: the number of available comments; the commentators' activity; and the quality of available comments.\\[2mm]
\textbf{Number of comments}: Statistics on the number and length of comments given to images are shown in Table~\ref{tab:comment_stats}. On average, an image tends to have about 11 comments, with a comment having about 18 words on average. However, the mean number of comments given during a challenge is greater than the mean number of comments given after. Interestingly, the length of comments given during a challenge is on average much shorter than those given after the challenge. Our observations lead us to believe that this is due to a "critique club" effect. The critique club comprises volunteer members who give a detailed critique of images which they have been assigned to review. The website states that\footnote{http://www.dpchallenge.com/forum.php?action=read\\\&FORUM\_THREAD\_ID=19842
}
\begin{myQuote}
"...the Critique Club critiques should be significantly longer than your average challenge comment and they should contain details about why the viewer feels a certain way about a photograph."
\end{myQuote}
For an image to be critiqued, its author must request a critique when submitting the image. These critiques are then posted to the image's page {\em after} voting has finished. As such comments are detailed and long, they likely increase the average length of comments given after challenge completion.\\

\begin{figure}[!t]
 \centering
\includegraphics[width=0.8\linewidth]{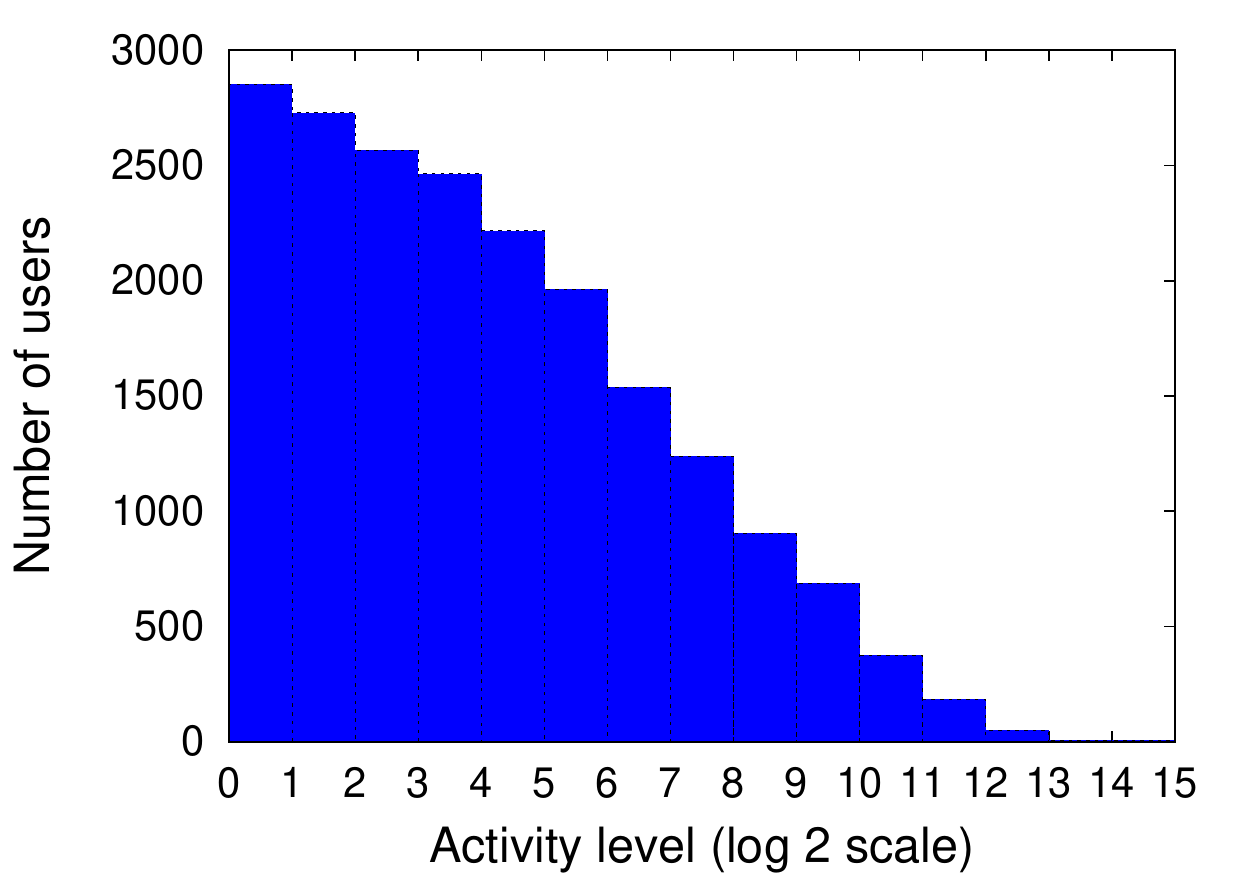}
\caption{Histogram of number of users for different activity levels, where activity level is denoted by number of comments made. The activity level ranges from 1 to 24,232 comments.}
\label{fig:comm_activity}
\end{figure}

As shown in Table~\ref{fig:ava_comment_hist}, the number of comments made about an image varies significantly with respect to the mean score given to that image. Unsurprisingly, high-scoring images have a large number of comments compared to other images. This bias is more pronounced when comparing the number of comments given during voting to the number of comments given after. Images with mean scores close to the midpoint of the score scale tend to have very few comments, perhaps because it is difficult to form an opinion about an image that is neither clearly bad nor clearly good. However, the mean length of the comments given to such images is much higher than the global average. This may be because critique club comments are often one of the few comments given to such images, and bias the mean length towards a higher number.\\[2mm]
\begin{table*}[t]
\small
\centering
\begin{tabular}{cccc}
\toprule
Statistics & During challenge & After challenge & Overall \\
\midrule
comments per image ($\mu$ and $\sigma$) & 9.99 (8.41) & 1.49 (4.77)& 11.49 (11.12)\\
words per comment ($\mu$ and $\sigma$) & 16.10 (8.24) & 43.51 (61.74)& 18.12 (11.55) \\
\bottomrule
\end{tabular}
\caption{Statistics on comments in AVA. 
On average, an image tends to have about 11 comments, with a comment having about 18 words on average. 
As the statistics in columns 2 and 3 attest however, commenting behavior is quite different during and after challenges.}
\label{tab:comment_stats}
\end{table*}
\begin{table*}[!t]
\centering
\begin{tabular}{MMMM}
\toprule
Statistic & During challenge & After challenge & Overall \\
\midrule
Mean number of comments per image & 
\includegraphics[width=0.23\textwidth, clip=true, trim=5mm 0cm 0cm 0cm]{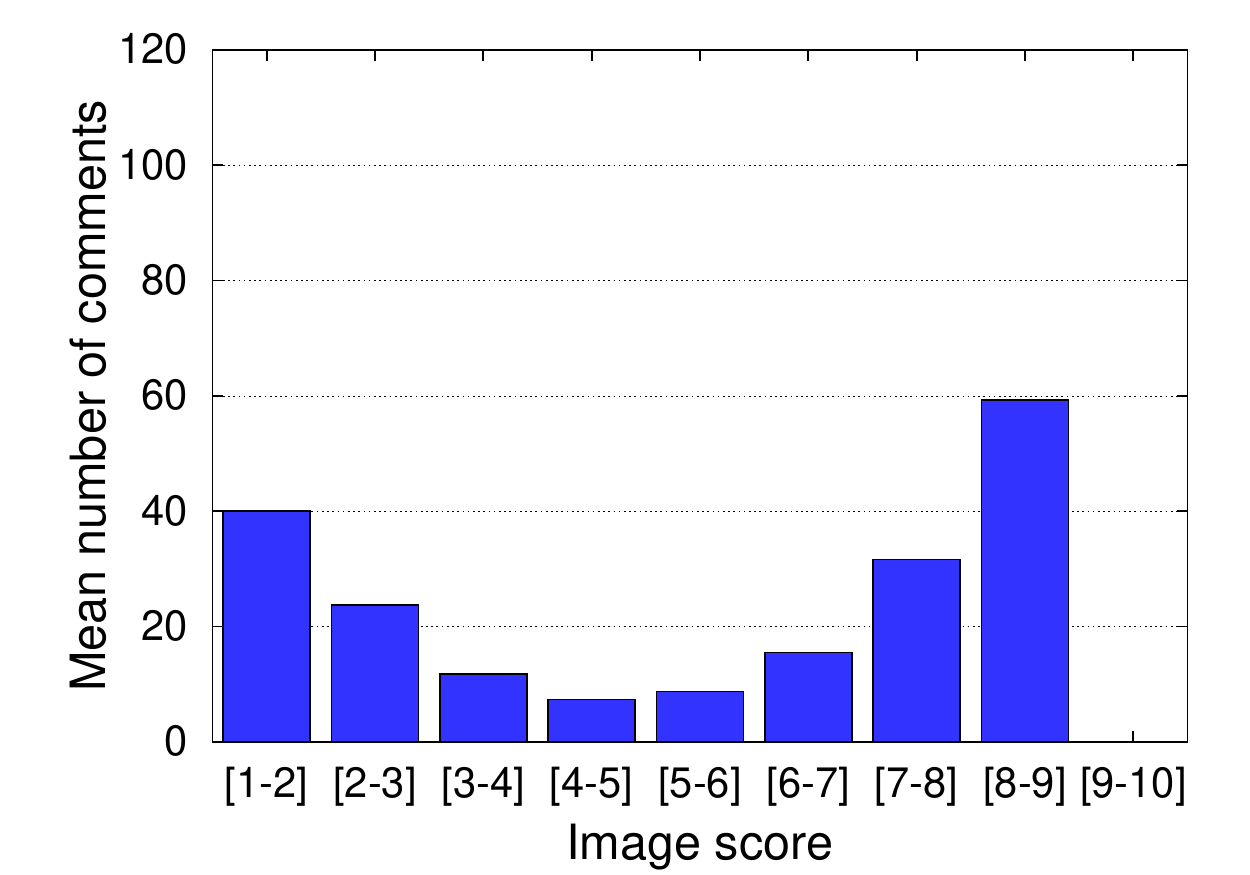} &
\includegraphics[width=0.23\textwidth, clip=true, trim=5mm 0cm 0cm 0cm]{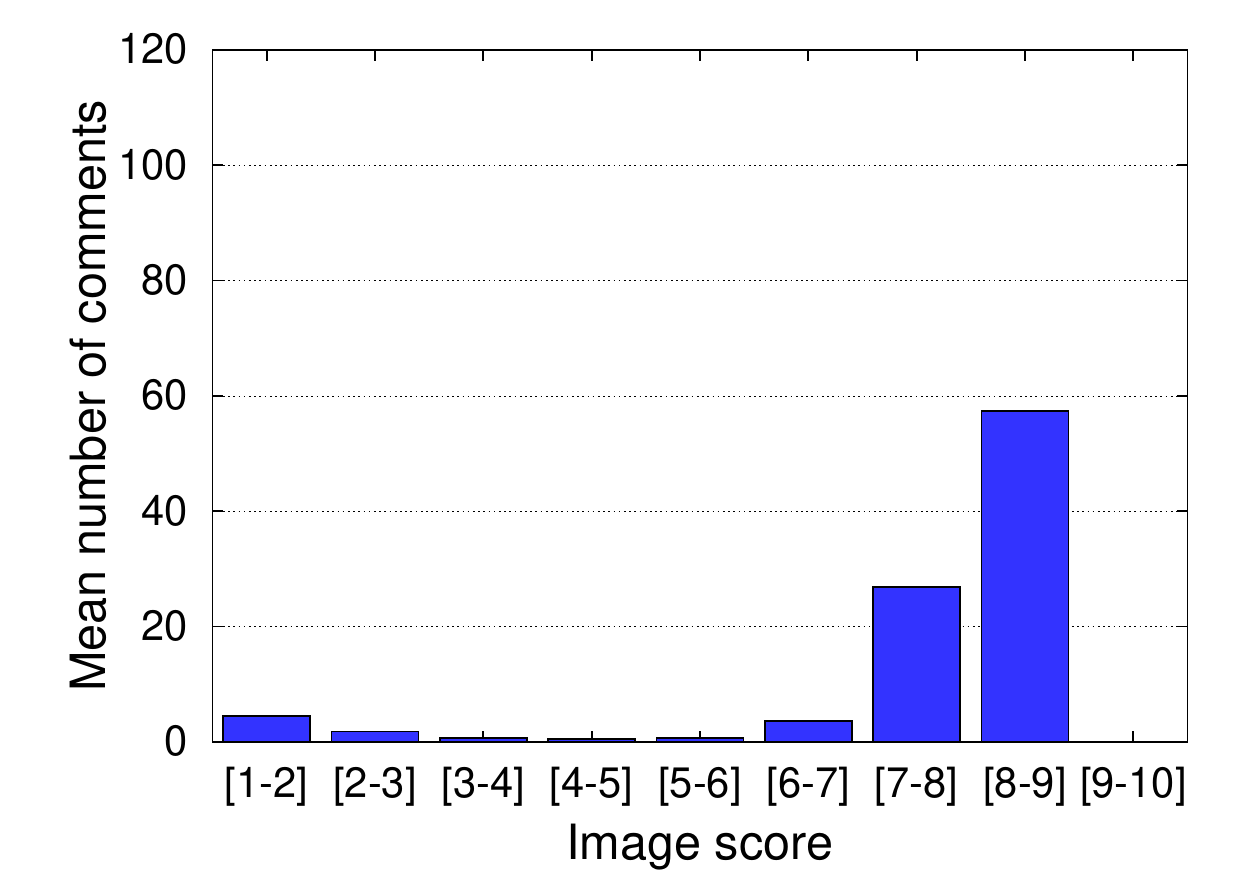} &
\includegraphics[width=0.23\textwidth, clip=true, trim=5mm 0cm 0cm 0cm]{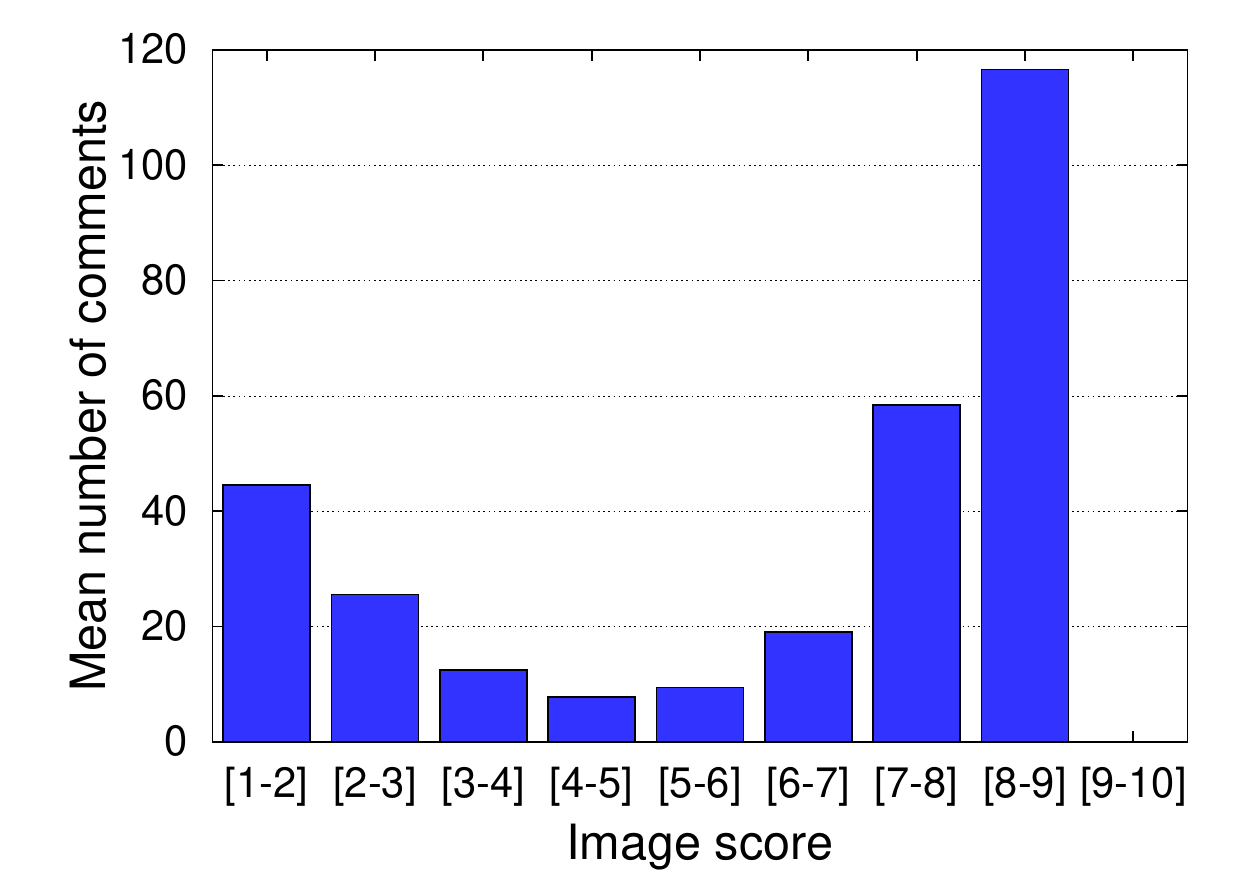}
\\
Std. dev. of number of comments per image & 
\includegraphics[width=0.23\textwidth, clip=true, trim=5mm 0cm 0cm 0cm]{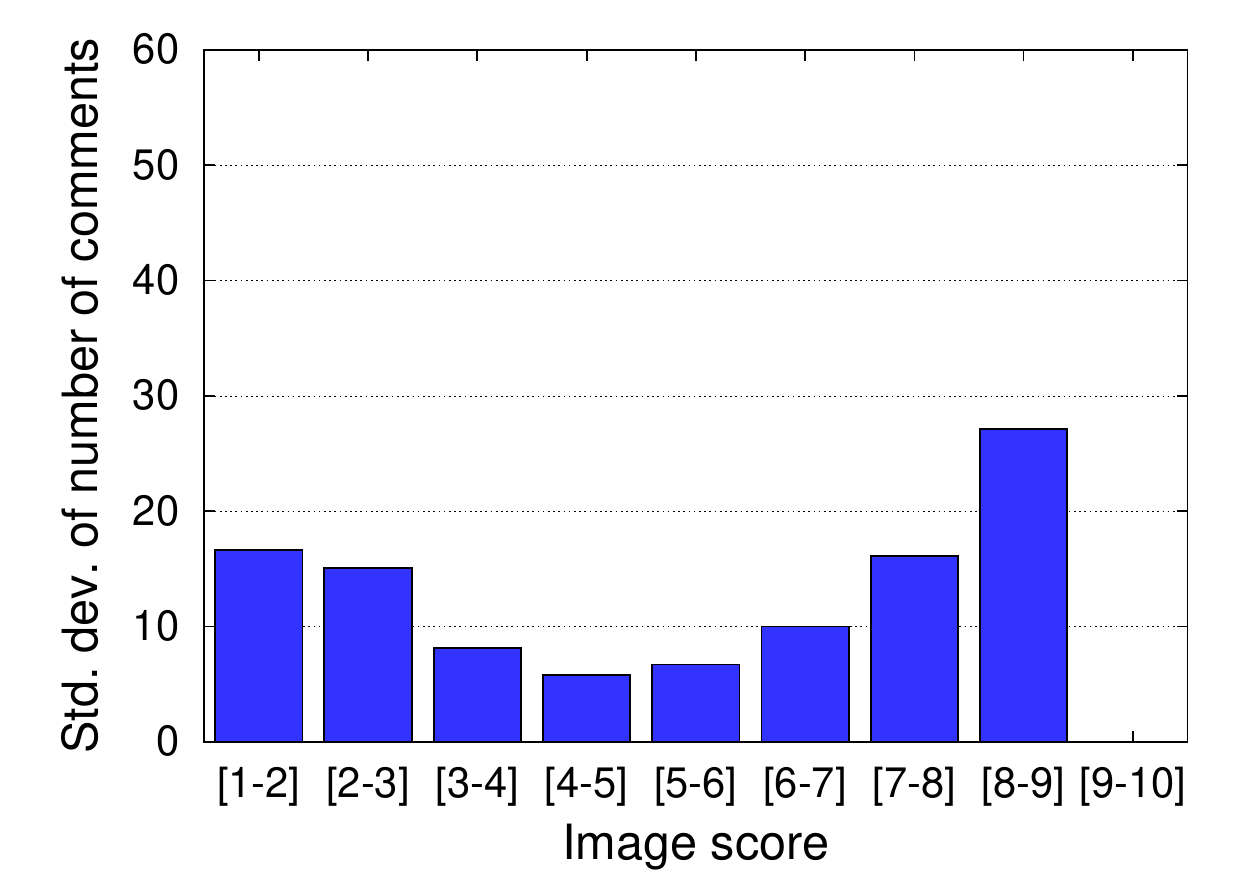} &
\includegraphics[width=0.23\textwidth, clip=true, trim=5mm 0cm 0cm 0cm]{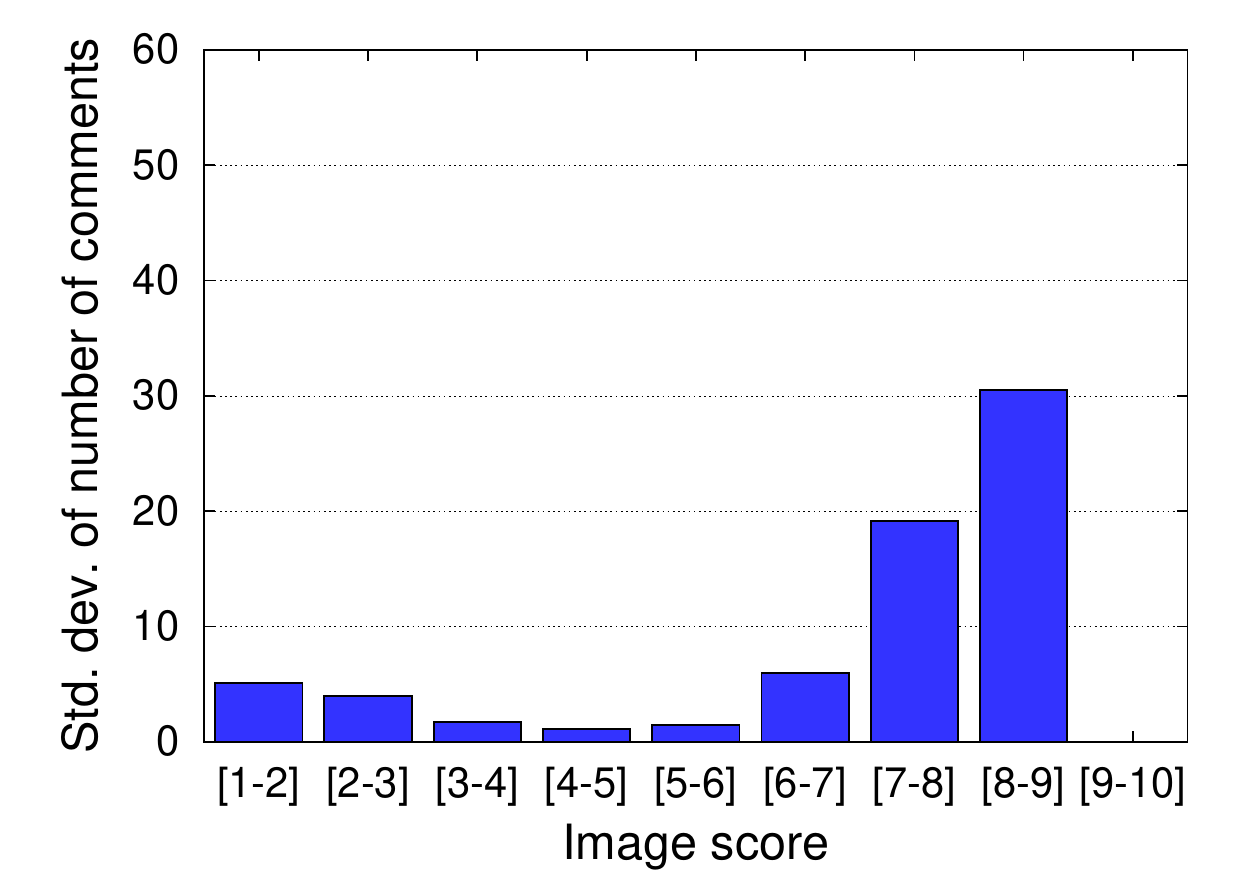} &
\includegraphics[width=0.23\textwidth, clip=true, trim=5mm 0cm 0cm 0cm]{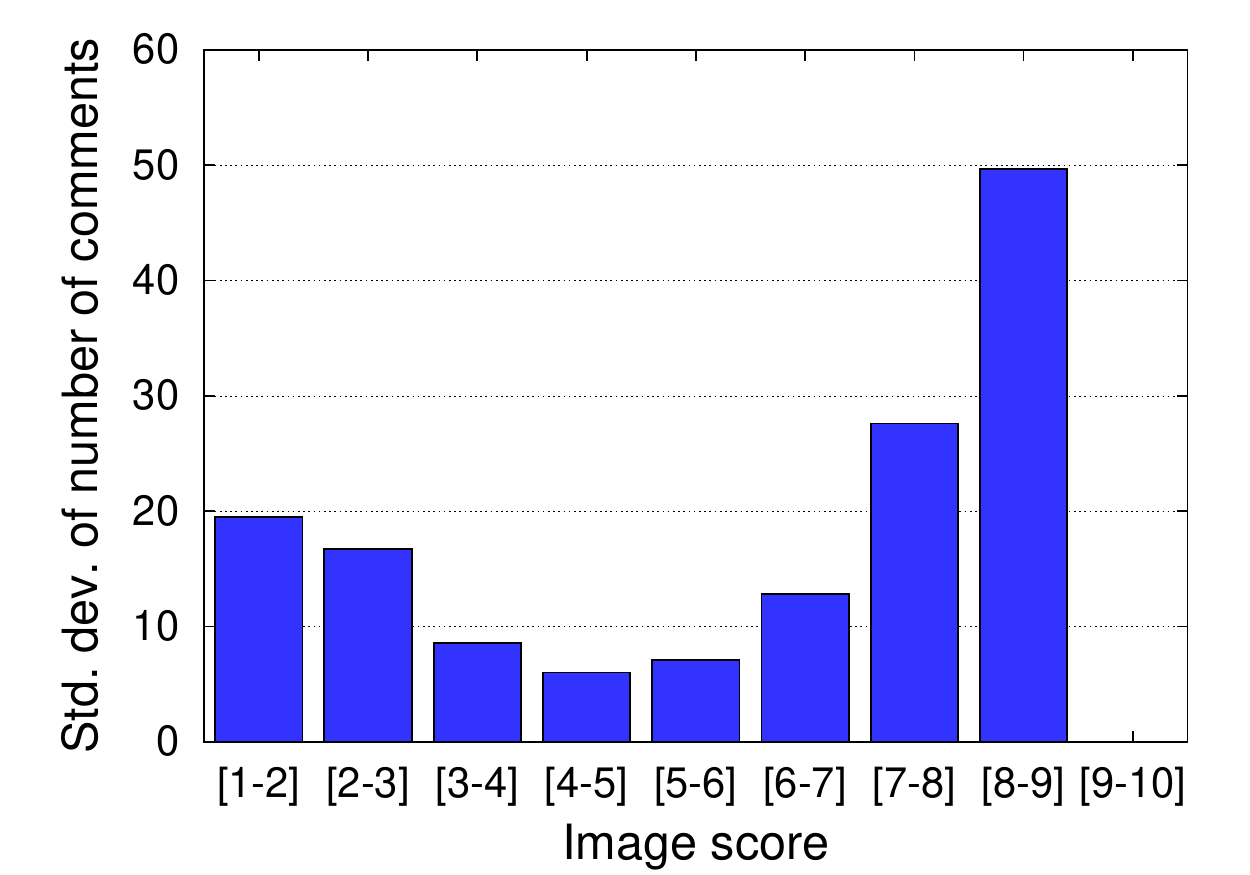}
\\
Mean comment length &
\includegraphics[width=0.23\textwidth, clip=true, trim=5mm 0cm 0cm 0cm]{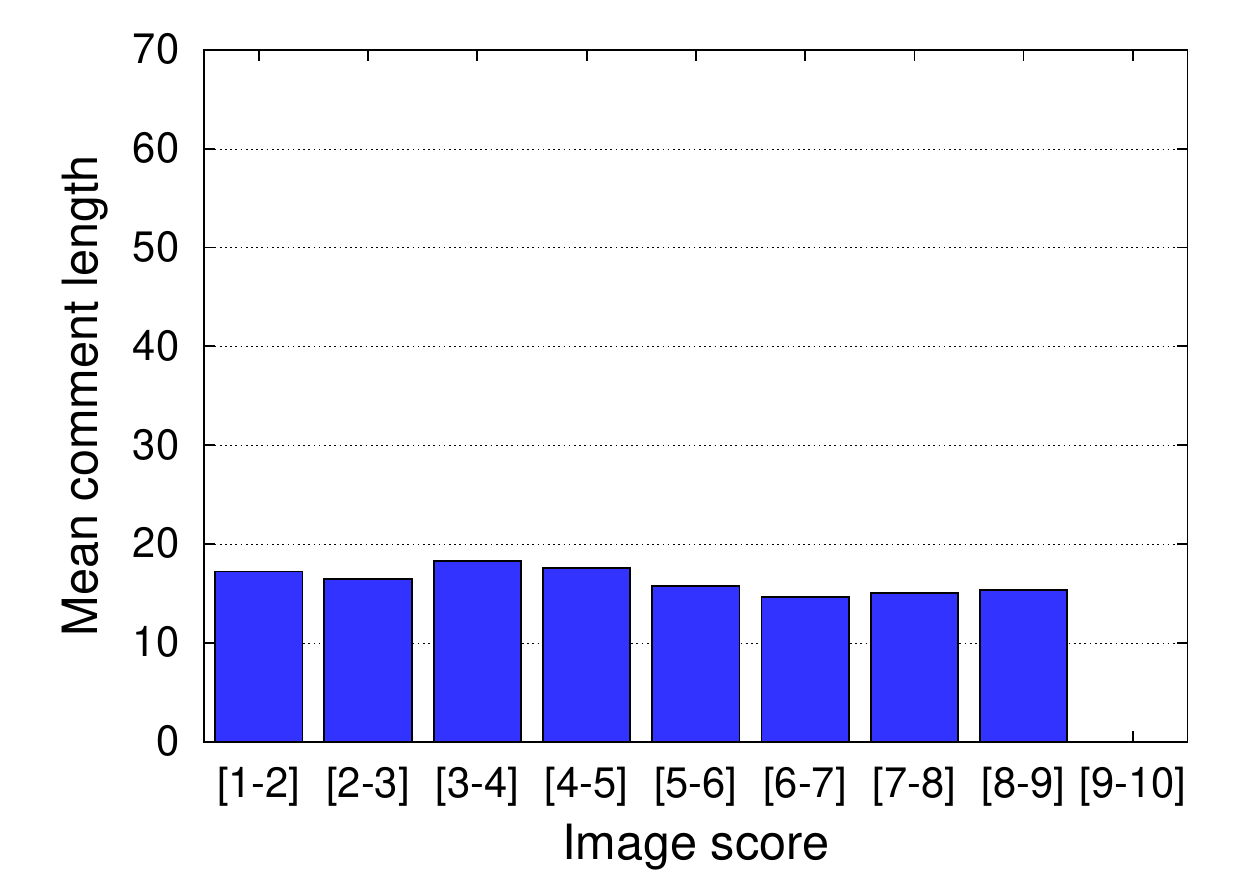} &
\includegraphics[width=0.23\textwidth, clip=true, trim=5mm 0cm 0cm 0cm]{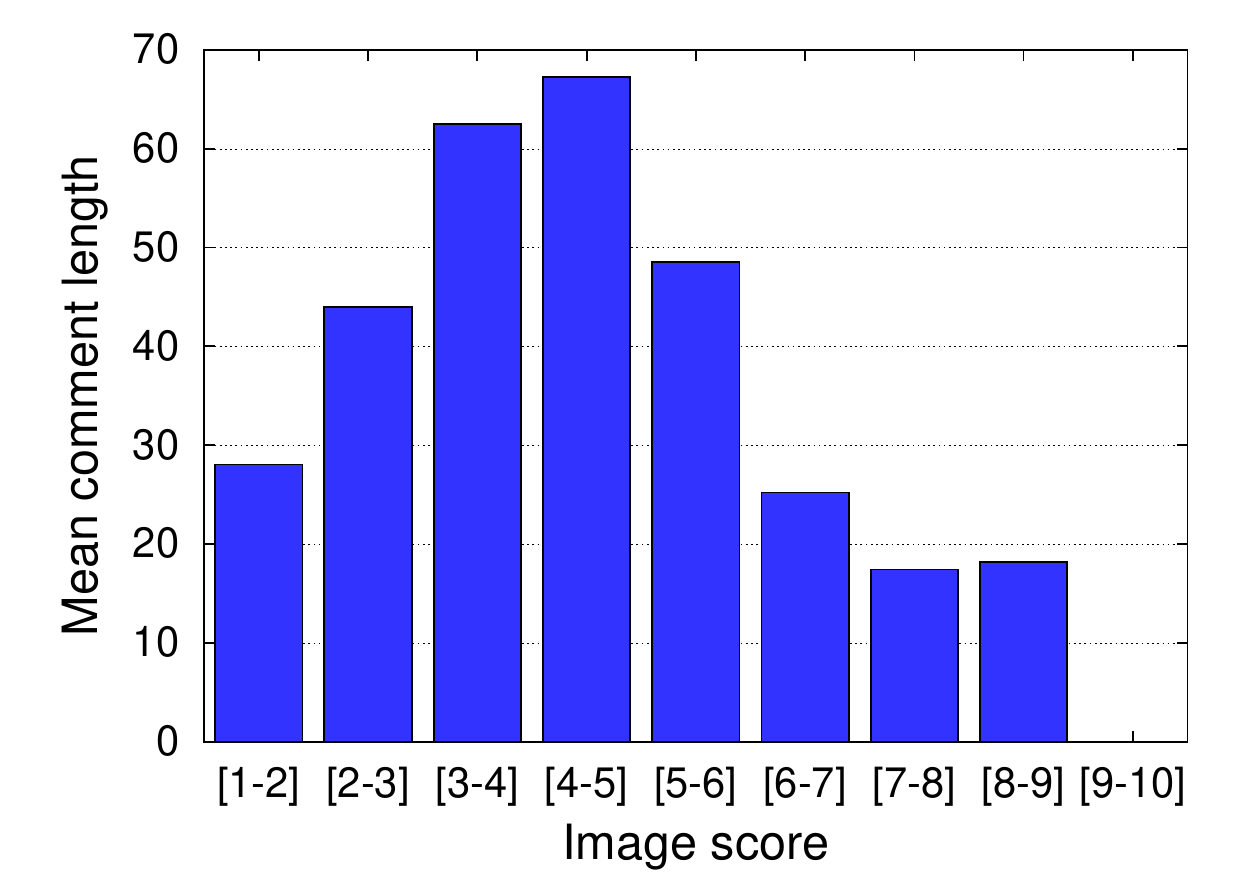} &
\includegraphics[width=0.23\textwidth, clip=true, trim=5mm 0cm 0cm 0cm]{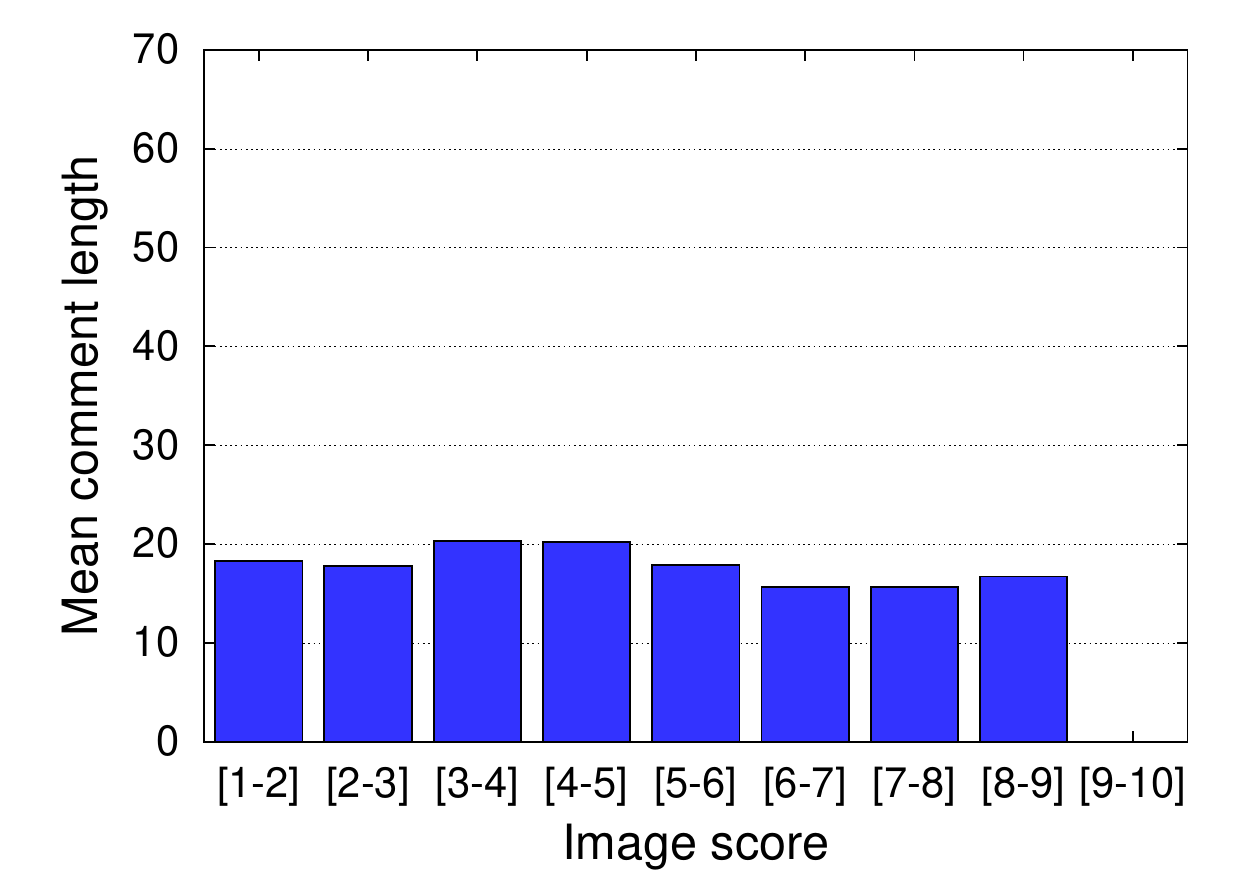}
\\
Std. dev. of comment length &
\includegraphics[width=0.23\textwidth, clip=true, trim=5mm 0cm 0cm 0cm]{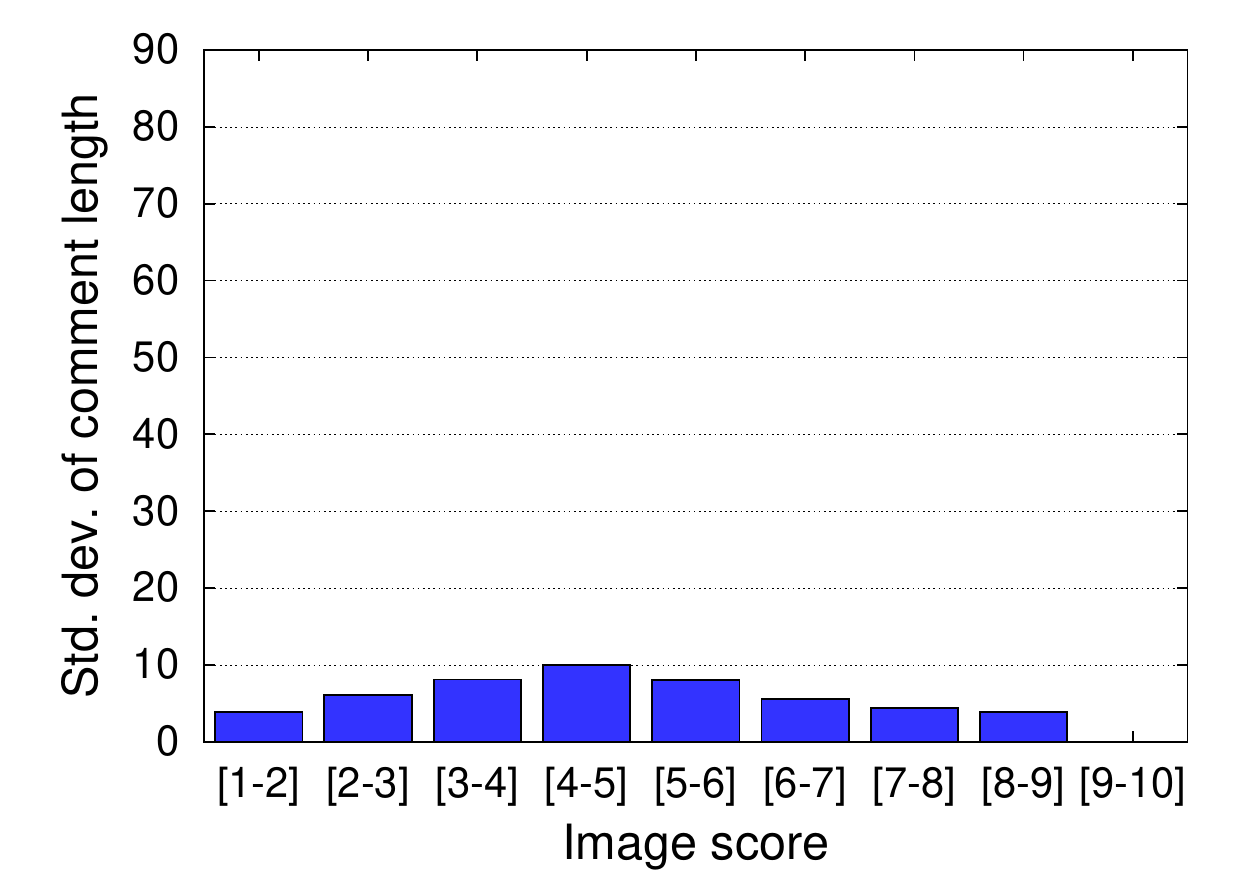} & 
\includegraphics[width=0.23\textwidth, clip=true, trim=5mm 0cm 0cm 0cm]{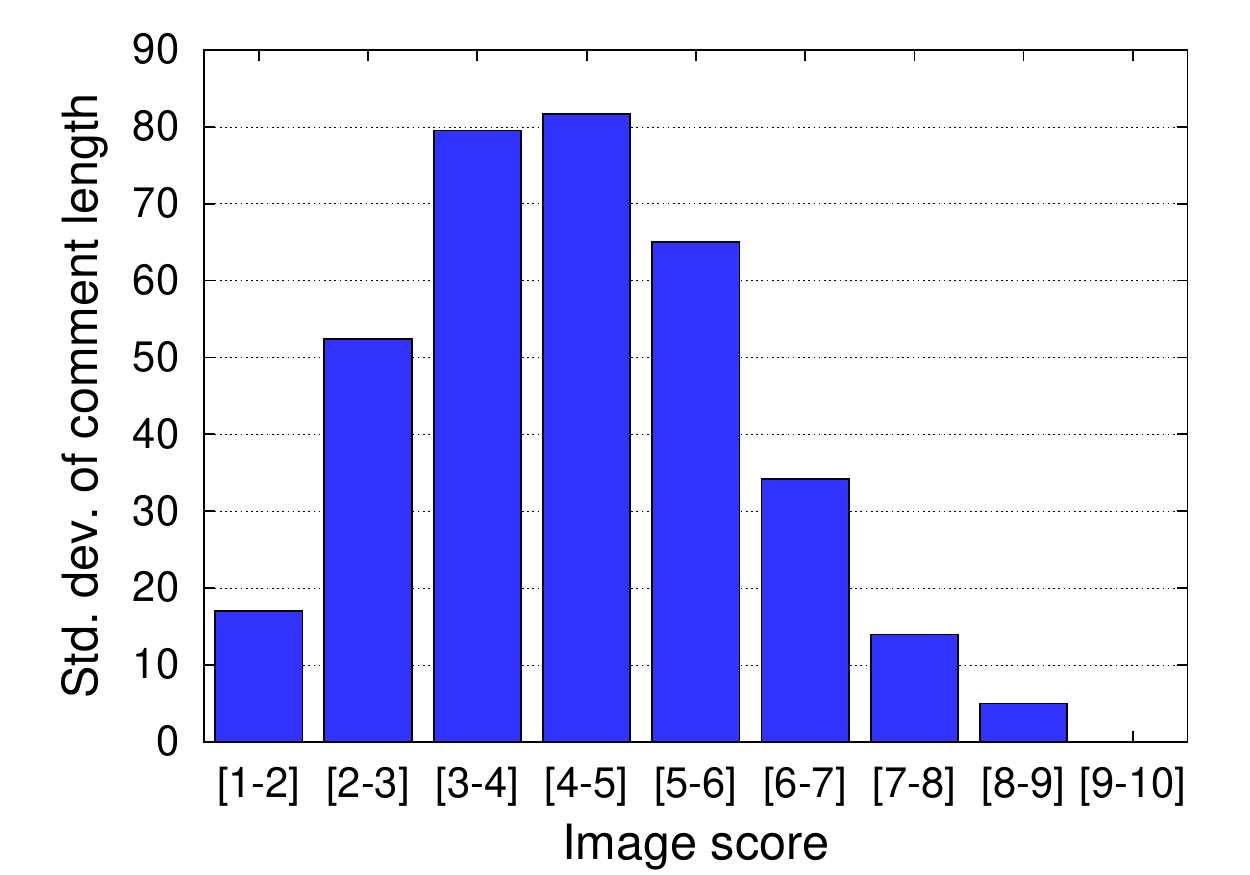} & 
\includegraphics[width=0.23\textwidth, clip=true, trim=5mm 0cm 0cm 0cm]{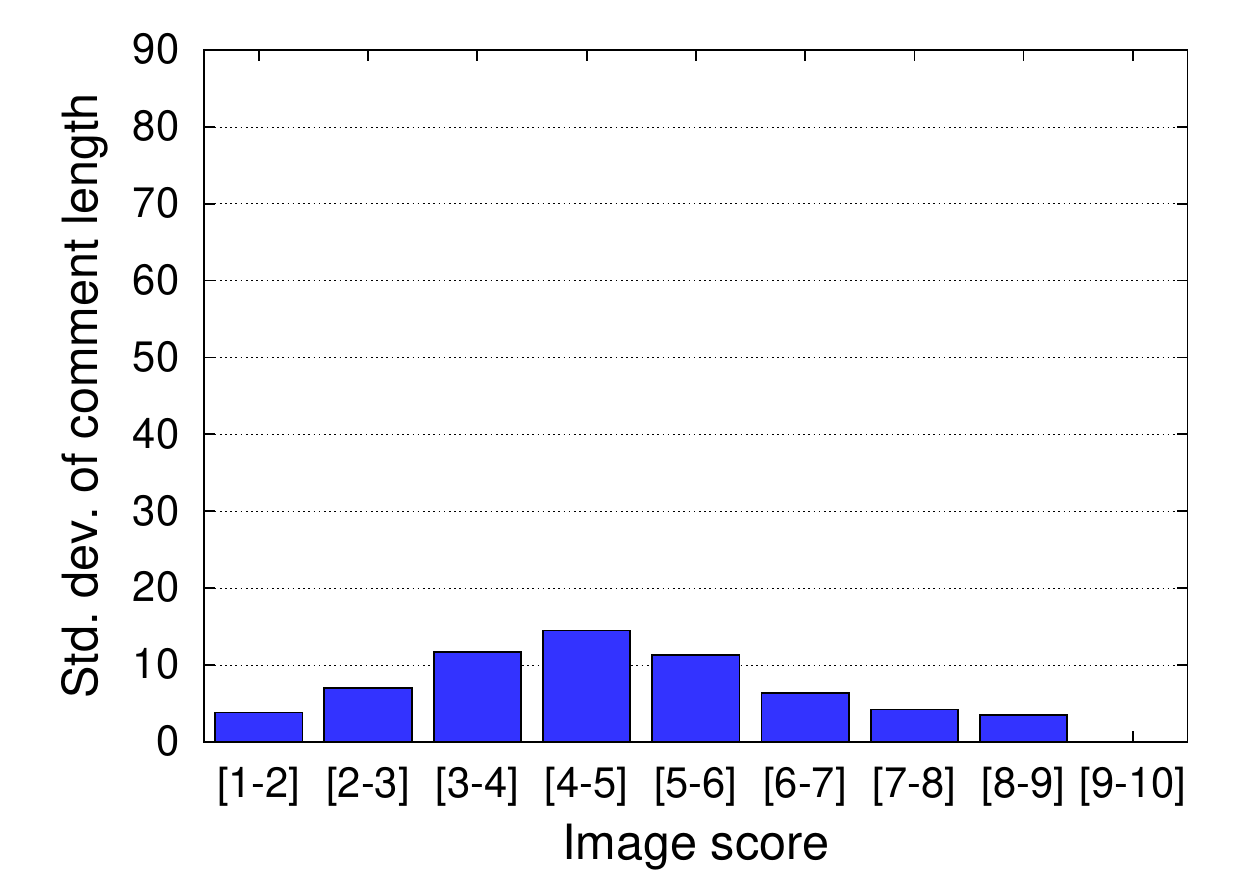}
\\
\bottomrule
\end{tabular}
\caption{Number of comments in the AVA database and their length (in number of words) for images within the given score range. More and longer comments are made during challenges than afterwards. Overall, high-scoring images have a large number of comments compared to other images.}
\label{fig:ava_comment_hist}
\end{table*}
\noindent \textbf{Commentators' activity:} For the images in AVA, 27,557 unique members made 2,934,728 comments. Fig.~\ref{fig:comm_activity} shows the commenting activity of these commentators. We found that approximately 86\% of users write comments only occasionally, while the remaining 3,983 users are regular commentators who have authored at least 100 comments.\\[2mm]
\noindent\textbf{Technical content in comments}: We investigated the words present in comments to determine how many comments contained technical content related to photographic techniques and aesthetic quality. We manually selected the technical words found among the 1,000 most frequently used words in the set of comments. We found 149 such words, examples of which are ``exposure", ``lighting", ``vivid" and ``texture". We note that this was a non-exhaustive list of the technical terms included in the corpus of comments. Even so, we found that 77\% of comments include at least one of these technical words, and among these comments, 2.8 words were used on average.

We next describe how AVA's textual comments, used in conjunction with its real-valued scores can be leveraged to automatically discover visual attributes.
\section{Discovering textual attributes}
\label{sec:mining}

As stated earlier, we aim to use the user comments of the AVA dataset as a textual resource, since they contain very rich information about aesthetics.
However, such comments are quite noisy: they can be very short as shown in the previous section and they are written in a very spontaneous manner.
This makes our task particularly challenging.

In this section we first describe how the comments found in AVA can be used to obtain textual features for image aesthetics (section~\ref{sec:textfeats}).
We then describe in section~\ref{sec:unsupdisc} a first approach to attribute discovery which is fully unsupervised as it only relies on comments. 
We show its limitations and then propose in section~\ref{sec:supdisc} a supervised approach to attribute discovery 
which relies on the user scores. 

\subsection{Textual features for image aesthetics}\label{sec:textfeats}
Textual information has only recently been used to infer the aesthetics of images. \citep{geng2011} created several bags-of-textual-words from different text sources related to web images. These sources included the image url and the title of the page on which the image is found. The textual vocabulary consisted of the 8 words in their dataset with the most information gain. \citep{SanPedro2012} used a sentiment analysis method to extract features from textual comments given to images by users. These features were the 49 most frequents words used in comments to refer to visual characteristics of images. Examples include ``color", ``composition" and ``lighting".

As mentioned previously, the textual comments in AVA contain detailed opinions of users on the aesthetic properties of images. We used these comments to create descriptors comprised of term frequency-inverse document frequency (tf-idf) weights. Such descriptors have been very successful in information retrieval applications \citep{joachims1998text}.

We first created a tokenized corpus using the comments of all images in AVA.
The terms in the corpus which are repeated at least 10 times are used to create a vocabulary.
We merge all the critiques related to an image into a single textual document. 
Merging the generally very short and noisy comments averages noise and thus leads to a more robust representation.
We tokenize and spell-check each document and we remove stop-words and numbers.
Each document is represented as a bag-of-words (BOW) histogram using the 
term frequency-inverse document frequency weighting (tf-idf).
Hence, each commented image is associated with a bag-of-words vector.

We constructed vocabularies comprising: (i) unigrams or single word terms; (ii) bigrams or terms compromising two words that appear consecutively in a comment; or (iii) unigrams and bigrams. We chose to investigate these particular vocabulary compositions as they achieved good performance in the text categorization literature \citep{bekkerman2004using}.

Our unigram, bigram and unigram+bigram vocabularies contained 30,595, 138,993 and 169,560 terms respectively. Bigrams retain some of the semantic relations between words, while this is completely lost in the case of unigrams. On the other hand, unigrams which are highly informative of aesthetic impressions are not present in the bigram feature representation.

We evaluate our TF-IDF vectors using a subset of AVA which we will call sAVA. This subset of 70,000 images was created by \citep{SanPedro2012} for evaluating textual features derived from user comments. We randomly select from sAVA 30,000 images for training, 10,000 for validation, and 30,000 images for testing. To evaluate on an aesthetics classification we must derive binary labels from the user scores. To do this, we follow \citep{DJLW08} and set two thresholds $\theta_1=5+\delta/2$ and $\theta_2=5-\delta/2$.  We then annotate each image with the label ``beautiful'' if $q_{av}(i) \geq \theta_1$ and ``bad'' if $q_{av}(i) \leq \theta_2$. $\delta$ is used to artificially create a gap between high and low quality images, as pictures lying in this gap are likely to represent noisy data in the peer-score process. As in~\citep{DJLW06} we vary this $\delta$ value in our experiments. Increasing the value $\delta$ obviously makes the classification task easier. Note that images belonging to the ``bad'' class are not necessarily bad {\it per se}. They only correspond to images that received lower scores.

\begin{figure}[!t]
 \begin{center}
  \includegraphics[width=1.0\linewidth]{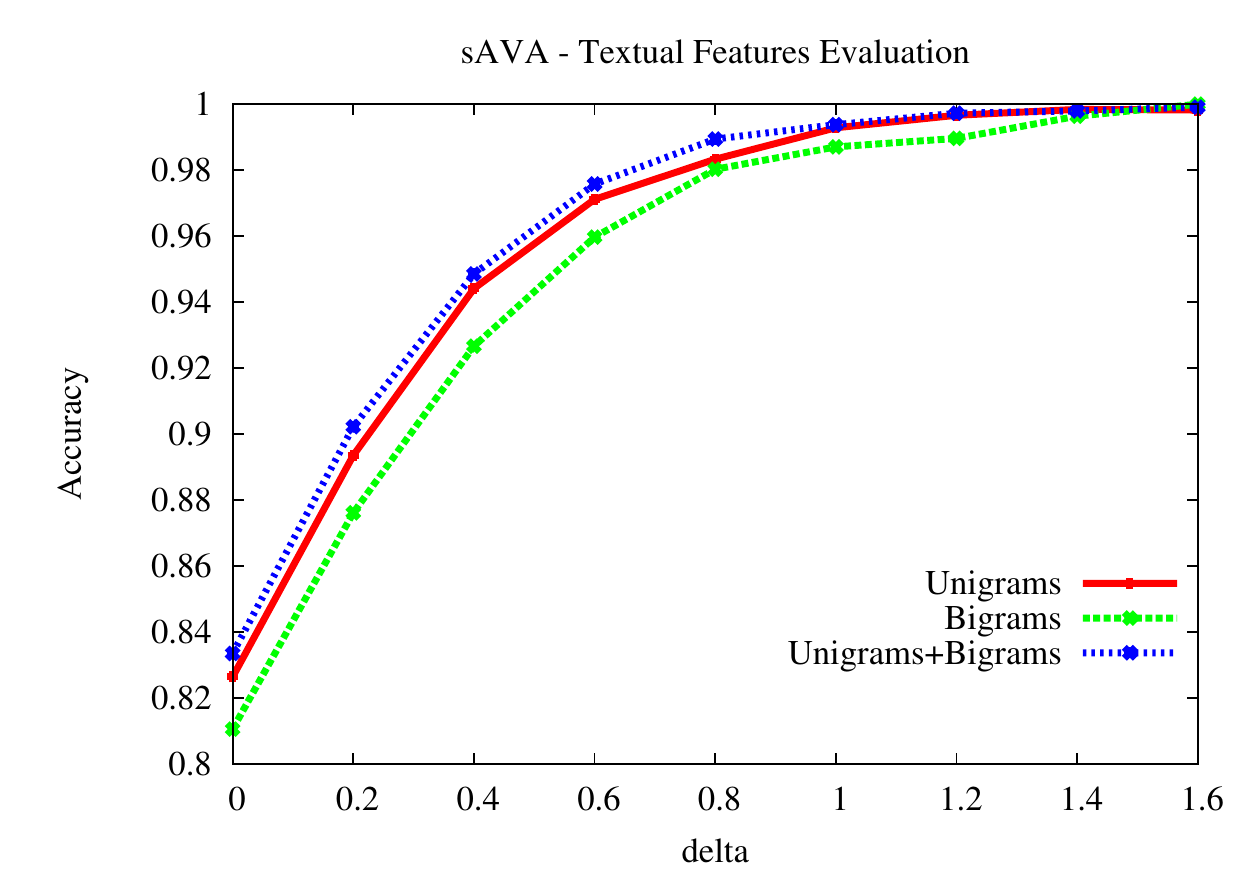}
 \end{center}
\caption{Classification accuracy on sAVA using different flavors of textual descriptors. Unigrams+Bigrams outperform Unigrams which outperform Bigrams.
\label{fig:comp_txt_sAVA}}
\end{figure}

Results are shown in Fig.~\ref{fig:comp_txt_sAVA}. We found that vectors constructed from a unigram vocabulary performed better than those formed from a bigram vocabulary, while vectors formed from a vocabulary of unigrams and bigrams out-performed both, findings which are consistent with text categorization problems \citep{bekkerman2004using}. However, the gain in performance was modest and unlikely to justify the increase in training time and storage requirements due to the increased vocabulary size.

We also measured the correlation between our classifier scores and the scores of the test images. 
As shown in Table~\ref{tab:text_results}, our textual features out-perform the text and visual-based features of \citep{SanPedro2012}, 
and the state-of-the-art visual features of \citep{Marches2011}. 
This shows that our textual features can be used to predict attractiveness,
thus validating their usefulness for our task. 
We next describe how we automatically discover attributes from these features.

%
%

\begin{table}[!t]
\centering
\begin{tabular}{lc}
\toprule
Method & Spearman's $\rho$\\
\midrule
San Pedro {\em et al.}, visual-based & 0.3133\\
Marchesotti {\em et al.}, visual-based & 0.4524 \\
San Pedro {\em et al.}, comment-based & 0.5839\\
San Pedro {\em et al.}, visual+comment-based & 0.6107\\
Unigrams & 0.8335\\
Bigrams & 0.8209\\
Unigrams + Bigrams & \textbf{0.8433}\\
\bottomrule
\end{tabular}
\caption{Regression performance on sAVA dataset. Our proposed textual features outperform the state-of-the-art feature extraction schemes.}
\label{tab:text_results}
\end{table}


\begin{table*}[!t]
\footnotesize
\centering
  \begin{tabular}[c]{|p{.5cm}|p{16cm}|}
\hline
\textbf{T3:}& ribbon, congrats, congratulations, deserved, first, red, well, awesome, yellow, great, glad, fantastic, excellent, page, wonderful, happy\\
\textbf{T11:}& beautiful, wow, amazing, congratulations, top, congrats, finish, love, stunning, great, wonderful, excellent, awesome, perfect, fantastic, gorgeous, absolutely, capture\\
\textbf{T28:}& idea, creative, clever, concept, cool, executed, execution, original, well, great, pencil, job, creativity, thought, top, work, shannon, interesting, good\\
\textbf{T20:}& funny, lol, laugh, hilarious, humor, expression, haha, title, fun, made, oh,love,smile, hahaha, great\\
\hline
\textbf{T35:}& motion, panning, blur, speed, movement, shutter, moving, blurred, abstract, blurry, effect, pan, stopped, sense, camera, fast, train, slow, background, exposure\\
\textbf{T27:}& colors, red, colours, green, abstract, color, yellow, orange, beautiful, colour, border, vibrant, complementary, composition, leaf, lovely, love, background, bright, purple\\
\textbf{T49:}& selective, desat, desaturation, red, use, color, works, processing, desaturated, saturation, editing, fan\\
\hline
\textbf{T8:}& portrait, eyes, face, expression, beautiful, skin, hair, character, portraits, eye, smile, nose, lovely, self, girl, look, wonderful, great, lighting, crop\\
\textbf{T14:}& cat, cats, kitty, eyes, fur, pet\\
\textbf{T37:}& sign, road, signs, street, stop\\
\hline
  \end{tabular}
  \caption{Sample topics generated by pLSA for $K=$50 topics.}
  \label{tab:plsatopics}
\end{table*}

\subsection{Attributes discovery from textual features}
We aim to use the features, or terms, in our textual vocabulary as aesthetic attributes.
We next present two approaches to this task: one without and one with supervisory data.

\subsubsection{Unsupervised attributes discovery}\label{sec:unsupdisc}
As a first attempt to discover attributes, we use the unsupervised 
probabilistic Latent Semantic Analysis (pLSA)~\citep{hofmann2001unsupervised} algorithm
on the BOW histograms. 
The hope is that the learned topics correlate with photographic techniques and therefore they are interpretable as attributes.
In Table \ref{tab:plsatopics}, we report some of the most interpretable topics discovered by pLSA with $K=$50 hidden topics.  
We can see that some topics relate to general appreciation and mood (T3, T11, T28, T20), 
to photographic techniques and colors (T35, T27, T49) or to semantic labels (T8, T14, T37).
Despite the relevance of these topics to visual attractiveness, we cannot directly use them as attributes:
they are too vague (i.e. not granular enough) and much manual post-processing would be needed to extract something useful.
Experiments with different numbers of topics $K$ did not lead to more convincing results.

\subsubsection{Supervised attributes discovery}\label{sec:supdisc}
We devise an alternative strategy based on the following approach: 
we use the attractiveness scores as \textit{supervisory information} to mitigate the noise of textual labels. 
The hope is that by using attractiveness scores we will be able to identify
interpretable textual features that are highly correlated with aesthetic
preference and use them to predict aesthetic scores.\\[2mm]
{\bf Selecting discriminative textual features}.
We mine beautiful and ugly attributes by discovering which terms 
can predict the aesthetic score of an image.

For this purpose, we train an elastic net \citep{zou2005regularization} support vector regressor to predict aesthetic scores and, at the same time, select textual features.
It is a regularized regression method that combines an $\ell_2$-norm and a sparsity-inducing $\ell_1$-norm. 
Let $N$ be the number of textual documents. 
Let $D$ be the dimensionality of the BOW histograms.
Let $\mathbf{X}$ be the $N \times D$ matrix of documents.
Let $y$ be the $N \times 1$ vector of scores of aesthetic preference (the score of an image is the average of the scores it received). 
Our goal if to learn a $D$-dimensional vector $\hat{\beta}$ that reflects the contribution of each BOW entry to the aethetic preference.
Toward this purpose, we optimize the following objective:
\begin{equation}
  \label{eq:ridge}
  \hat{\beta}=\arg\min_{\beta}||\mathbf{y}-\mathbf{X}\mathbf{\beta}||^2+\lambda_1||\beta||_1+\lambda_2||\beta||^2 
\end{equation}
where $\lambda_1$ and $\lambda_2$ are the regularization parameters.\\[2mm]
We first experiment with the same vocabulary of $D\approx$30,000 unigrams described in section~\ref{sec:textfeats}. 
We cross-validated the regularization parameters using Spearman's $\rho$
correlation coefficient and we selected the values of $\lambda_1$ and
$\lambda_2$ providing highest performances with $1,500$ non-zero $\beta$
coefficients.  
We analyze the candidate attributes by sorting them according to $|\beta|$ (see Table \ref{tab:ranksFeatures}) to verify their interpretability.
By inspecting the most discriminant \textit{unigrams}, we can see that the ones at the top of each rank relate to specific visual attributes (e.g. grainy, blurry).
But others can be ambiguous (e.g. not, doesn't, poor) and interpreting them is rather problematic.

\begin{table*}[!t]
\footnotesize
  \centering
  \begin{tabular}[c]{|p{1.4cm}|p{15cm}|}
\hline
{\bf\textsc{Unigrams+}}&
great  (0.4351), like  (0.3301), excellent  (0.2943), love  (0.2911), beautiful  (0.2704), done  (0.2609), very  (0.2515), well  (0.2465), shot  (0.2228), congratulations  (0.2223), perfect  (0.2142), congrats  (0.2114), wonderful  (0.2099), nice  (0.1984), wow  (0.1942), one  (0.1664), top  (0.1651), good  (0.1639), awesome  (0.1636),
\\
\hline
{\bf \textsc{Unigrams-}}& sorry  (-0.2767), focus  (-0.2345), blurry  (-0.2066), small  (-0.1950), not  (-0.1947), don  (-0.1881), doesn  (-0.1651), flash  (-0.1326), snapshot  (-0.1292), too  (-0.1263), grainy  (-0.1176), meet  (-0.1122), out  (-0.1054), try  (-0.1041), low  (-0.1013), poor  (-0.0978), distracting  (-0.0724), 
 \\
\hline
{\bf\textsc{Bigrams+}}&  well done  (0.6198), very nice  (0.6073), great shot  (0.5706), very good  (0.3479), great job  (0.3287), your top  (0.3262), my favorites  (0.3207), top quality  (0.3198), great capture  (0.3051), lovely composition  (0.3014), my top  (0.2942), 
nice shot  (0.2360), th placing  (0.2330), great lighting  (0.2302), great color  (0.2245), excellent shot  (0.2221), good work  (0.2218), well executed  (0.2069), great composition  (0.2047), my only  (0.2032)\\
\hline
{\bf \textsc{Bigrams-}}& too small  (-0.3447), too blurry  (-0.3237), not very  (-0.3007), does not  (-0.2917), not meet  (-0.2697), wrong challenge  (-0.2561), better focus  (-0.2280), not really  (-0.2279), sorry but  (-0.2106), really see  (-0.2103), poor focus  (-0.2068), too out  (-0.2055), keep trying  (-0.2026), see any  (-0.2021), , not sure  (-0.2017), too dark  (-0.2007), next time  (-0.1865), missing something  (-0.1862), just don  (-0.1857), not seeing  (-0.1785)
\\
\hline
  \end{tabular}
  \caption{Most discriminant unigrams and bigrams with their regression coefficient $\beta$. Bigrams are in general more interpretable than unigrams since they can capture the polarity of comments and critiques.}
  \label{tab:ranksFeatures}
\end{table*}

To resolve these ambiguities we turn to bigrams. 
As mentioned in section~\ref{sec:textfeats}, bigrams preserve some of the semantic relations between neighboring words, 
which is essential for our purpose of obtaining human-interpretable attributes.
In particular, bigrams capture \textit{non-compositional} 
meanings that a simpler feature does not~\citep{riloff2006feature}.
For instance the word ``lighting'' does not have an intrinsic polarity while a
bigram composed of ``great'' and ``lighting'' can successfully clarify the
meaning.  
As such, the use of bigrams is a popular choice in opinion mining \citep{pang2002thumbs}. 

We performed regression on the 90,000 most frequent bigrams among those described in section~\ref{sec:textfeats}, 
using the same procedure employed for unigrams.
The bottom rows of Table \ref{tab:ranksFeatures} show
the bigrams which receive the highest/lowest regression weights.
As expected, regression weights implicitly select those features as the most discriminant ones for predicting attractiveness.
The highest weights correspond to ``beautiful'' attributes while the lowest
weights correspond to ``ugly'' attributes. It is noteworthy that we use an Elastic Net to overcome the limitations of  other sparsity-inducing norms like LASSO \citep{tibshirani1996regression} in the feature selection tasks: if there is a group of features among which the pairwise correlations are very high, then the LASSO tends to select only one random feature from the group \citep{zou2005regularization}. 
In our case, LASSO produces a compact vocabulary of uncorrelated attribute labels, but also a very small number of labeled images. This is problematic because we need as many annotated images as possible at a later stage to train one visual classifiers for each attribute.\\[2mm]
{\bf Clustering bigrams.}
The effect of the Elastic Net on correlated features
can be seen by looking at table \ref{tab:ranksFeatures}: as expected, the
Elastic Net tolerates correlated features (synonym bigrams) such ``well done'' or ``very
nice'', ``beautiful colors'' and ``great colors''. This augments the number of
annotated images, but it requires us to handle synonyms in the vocabulary of
attributes. For this reason, we compact the list of 3,000 candidate bigrams
(1,500 for Beautiful attributes and 1,500 for Ugly attributes) with Spectral
Clustering \citep{ng2002spectral}. We cluster the beautiful and ugly bigrams
separately. We heuristically set the number of clusters to 200 (100 Beautiful
and 100 Ugly clusters) and we create the similarity matrices with a simple but
very effective measure of bigram similarity: we calculate the Levenshtein
distance among the second term within each bigram and we discard the first
term. This approach is based on the following intuition: most bigrams are composed of a first term which indicates the polarity and a second
term which describes the visual attribute \eg ``lovely composition'', ``too
dark'', ``poor focus''.  
What we obtain is an almost duplicate-free set of attributes, and a richer set of images associated with them.
Some sample clusters are reported here below:\\{\small C18:
  ['beautiful', 'colors'] ['great', 'colors'] ['great', 'colours'] ['nice',
  'colors']\\ C56: ['challenge', 'perfectly'] ['just', 'perfect']\\ C67:
  ['nicely', 'captured'] ['well', 'captured'] ['you', 'captured']\\ C89:
  ['excellent', 'detail'] ['great', 'detail'] ['nice',
  'detail'])}.
We randomly draw a bigram from each cluster to name the corresponding attribute.

\section{Learning visual attributes}
\label{sec:learning}

The goal is now to learn one visual attribute model for each discovered textual attribute.
However, it is difficult to hand-design a different visual model for each of our 200 attributes. 
Therefore we propose to learn such attribute models from generic visual features, in the same manner that \cite{Marches2011} proposed to use generic visual features to learn preference models.
In this section, we first describe the chosen generic visual features that we use to represent our images.
We then explain how attribute models are learned and then re-ranked based on visualness.

\subsection{Visual features for image aesthetics}
We extract 128-dim SIFT~\citep{Lowe2004} and 96-dim color descriptors \citep{CCP07} 
from 24x24 patches on dense grids every 4 pixels at 5 scales. We reduce dimensionality by using a 64-dim PCA.
These low-level descriptors are aggregated into an image-level signature using the Fisher Vector (FV) \citep{CLV11}.
We use visual vocabularies of 64 Gaussians and encode some rough image layout information by 
concatenating FVs extracted from the whole image, its 4 quadrants and three equally-sized horizontal image strips.
We chose this image representation as it has been shown to result in state-of-the-art performance for semantic \citep{CLV11} 
as well as aesthetic tasks \citep{Marches2011}.
We compute one SIFT-based and one color-based representation per image and we concatenate them.
This leads to a combined 131,072-dim representation
which is PQ-compressed \citep{JDS11} to reduce the memory footprint and 
to enable all images to be kept in RAM.

\subsection{Attribute learning from visual features}\label{sec:att_learning}
A categorization problem is considered to be large-scale if either
(i) the size of the feature space;
(ii) the number of classes; or
(iii) the number of training samples
is large.
Given 
the high dimensionality of the FVs,
the large number of images available in AVA (approx. 250,000), and
the large number of attribute classifiers to be learned, 
our classification problem resides squarely in the large-scale paradigm.
It is therefore fundamental to employ a scalable solution.

\begin{figure*}[t!]
  \centering
 \includegraphics[trim=.15cm 0cm 0.cm 0cm, clip=true,width=\linewidth]{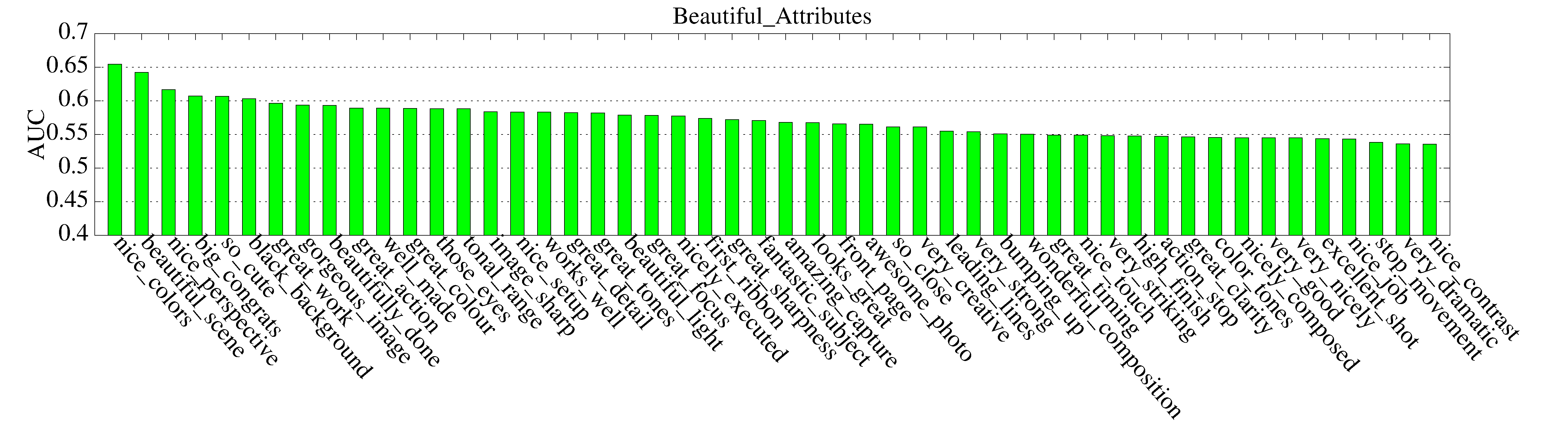}
 \includegraphics[trim=.15cm 0cm 0.cm 0cm, clip=true,width=\linewidth]{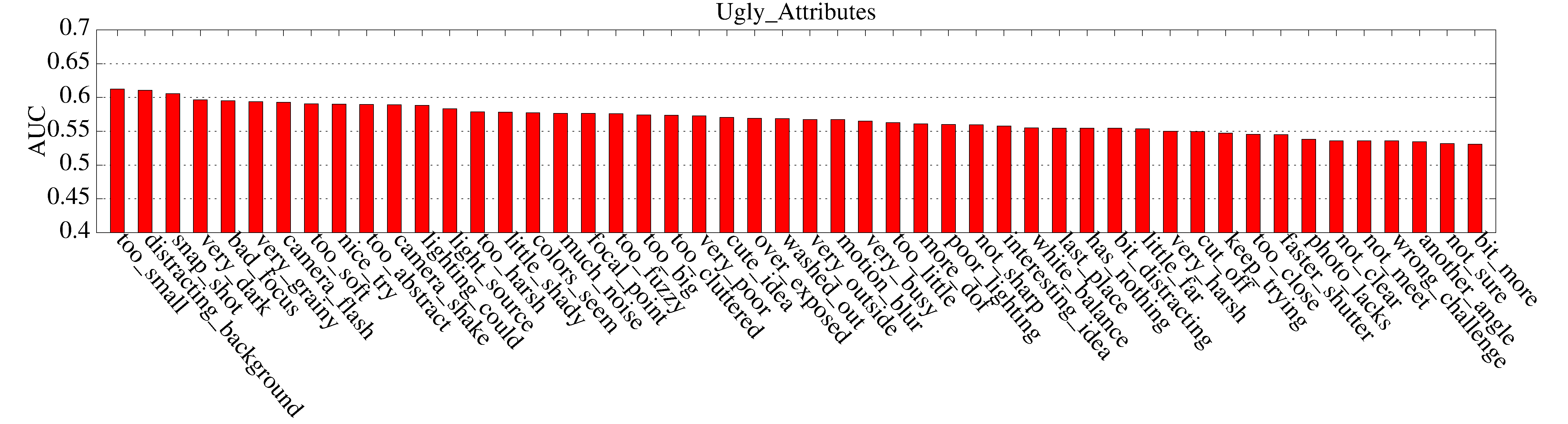}
\caption{Area Under the Curve (AUC) calculated for the top 50 Beautiful and Ugly attributes.}
  \label{fig:apbu}
\end{figure*}

For this reason, we use linear classifiers optimized with an online learning algorithm,
namely Stochastic Gradient Descent (SGD)~\citep{BB07}.
We use a regularized logistic regression objective function.
Using logistic loss (rather than a hinge loss for instance)
provides a probabilistic interpretation of the classification scores, 
which is a desirable property since we are training attributes.
The resultant linear classifiers are our visual attribute models.
%
%
In the previous section, we enforced interpretability and discriminability of the attribute labels using attractiveness scores as a supervision mechanism. 
However, this choice does not ensure that all these attributes can be recognized by a computer.
This is the reason why we measure ``visualness'' using the Area Under the ROC Curve
(AUC) calculated for each individual attribute. In particular, we benchmark the
classification performances of each attribute (1-vs-all) and we rank them using
AUC. We show the top 50 attributes in Figure \ref{fig:apbu} for Ugly and
Beautiful attributes. 
Our first observation is that beautiful attributes perform better than ugly attributes do.
This is not surprising since the latter attributes were trained with fewer images: 
as shown in Table~\ref{fig:ava_comment_hist}, people are less likely to comment on low-quality images, limiting the training set for ugly attributes.
Second, we notice that attributes which detect lighting conditions and colors
(e.g. {\em too dark}, \textit{great colour}, {\em too harsh}) perform better
than more complex visual concepts such as {\em interesting idea}, {\em bit
  distracting}, {\em very dramatic}. 

\begin{figure}
 \centering
\subfigure[Beautiful Attributes]{
    \includegraphics[width=0.49\linewidth]{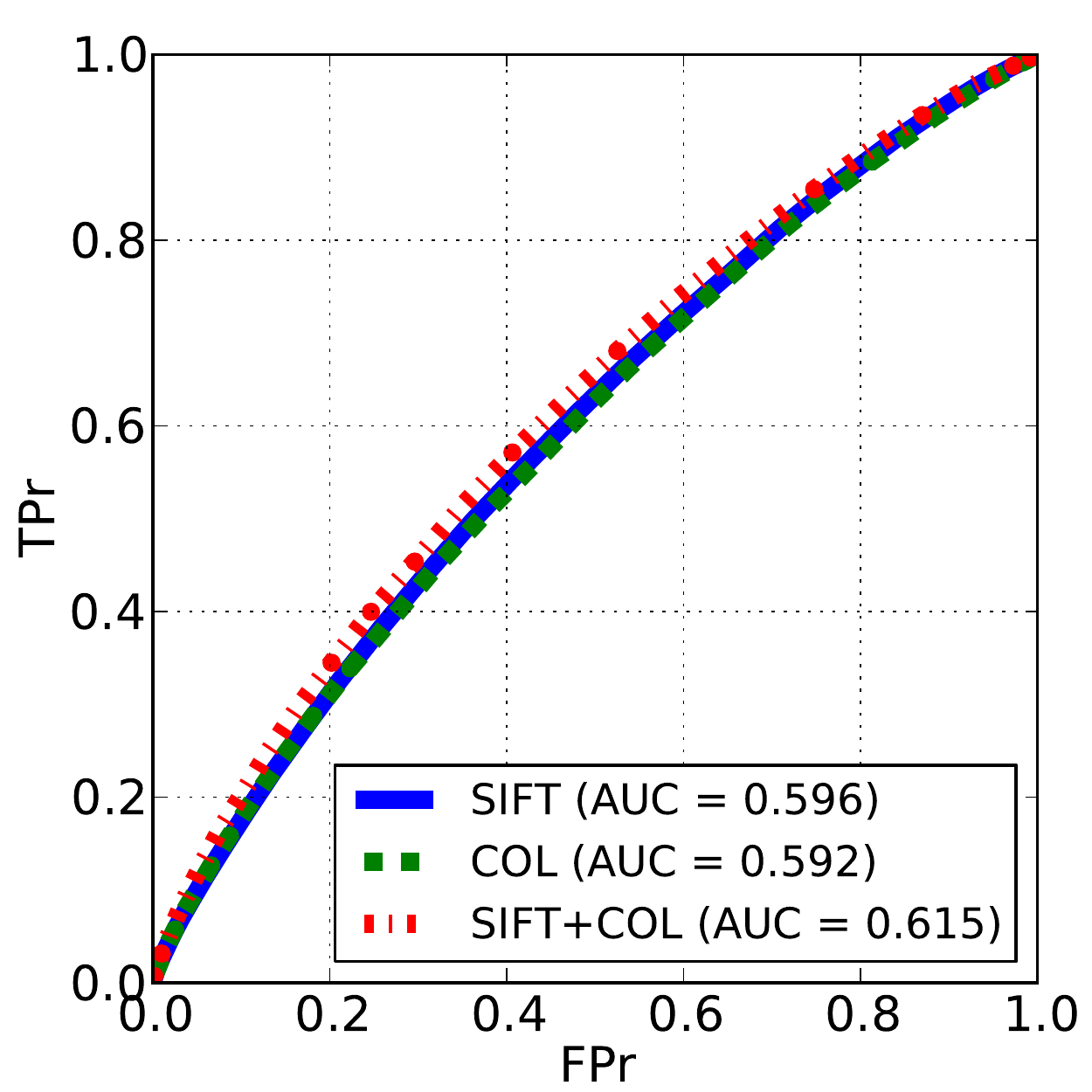}
}\subfigure[Ugly Attributes]{
    \includegraphics[width=0.49\linewidth]{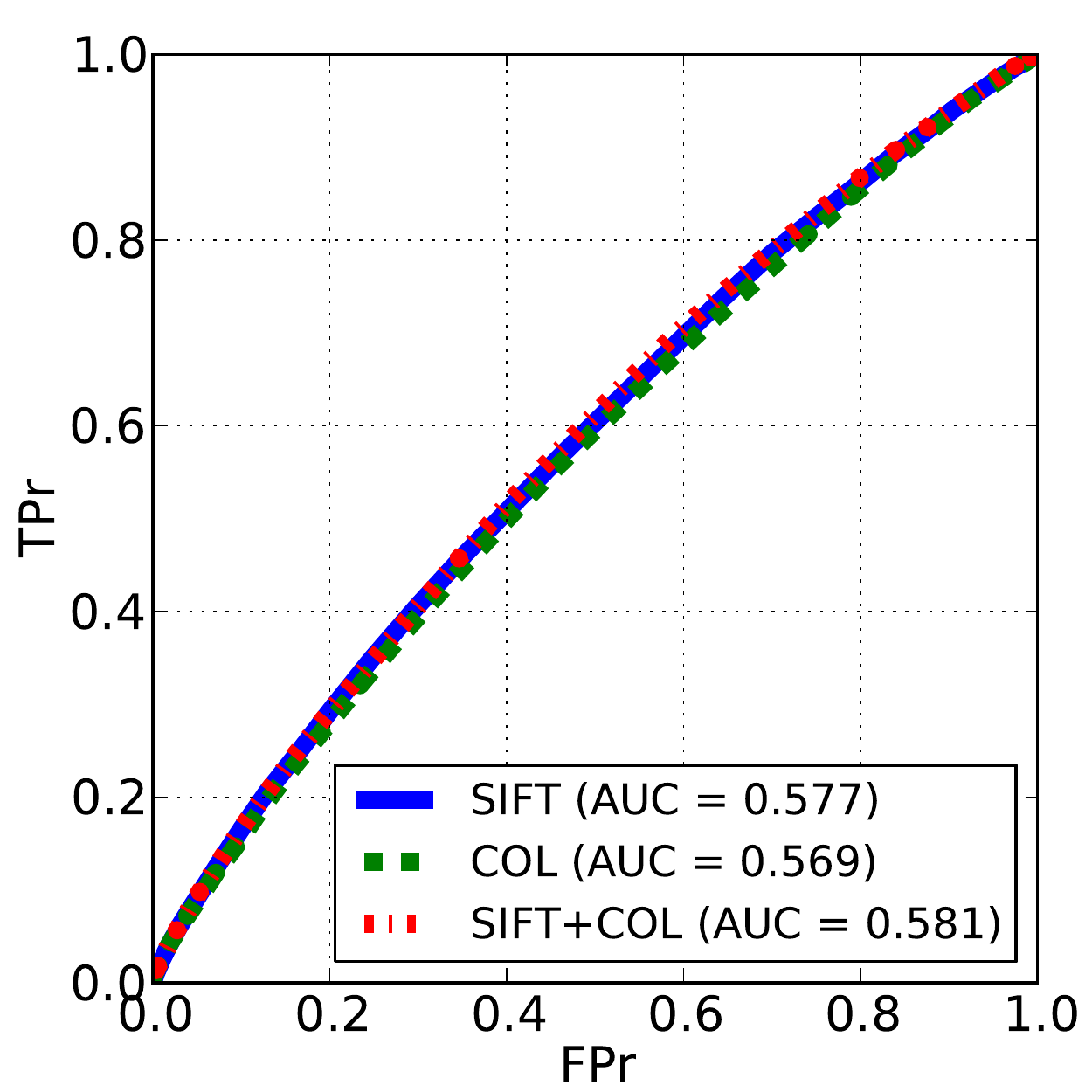}
}
\caption{ROC curves for SIFT features and color statistics features, averaged over (a) beautiful; and (b) ugly attributes.}
  \label{fig:roc_ptype}
\end{figure}

\begin{figure}
 \centering
\subfigure[Beautiful Attributes]{
    \includegraphics[width=0.49\linewidth]{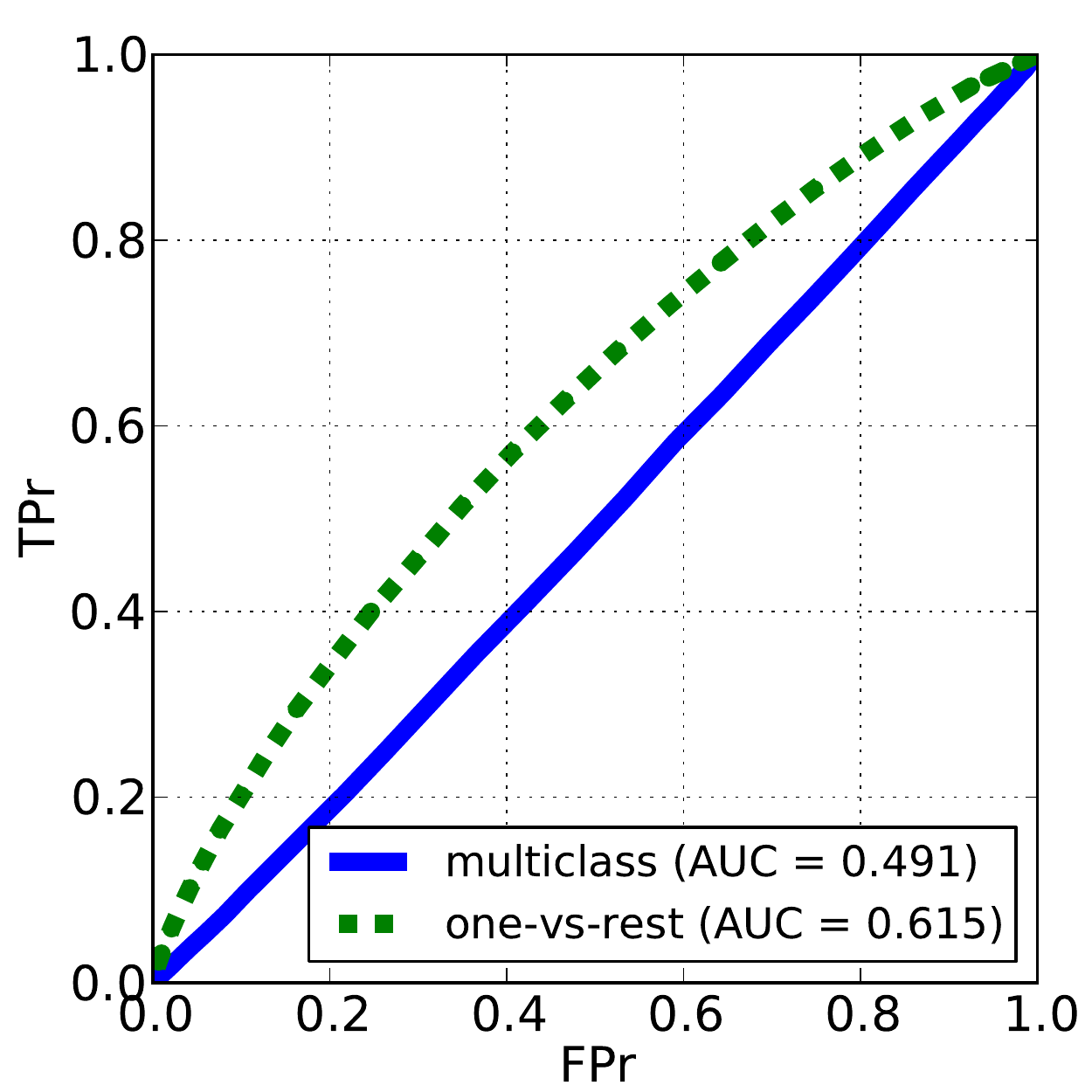}
}\subfigure[Ugly Attributes]{
    \includegraphics[width=0.49\linewidth]{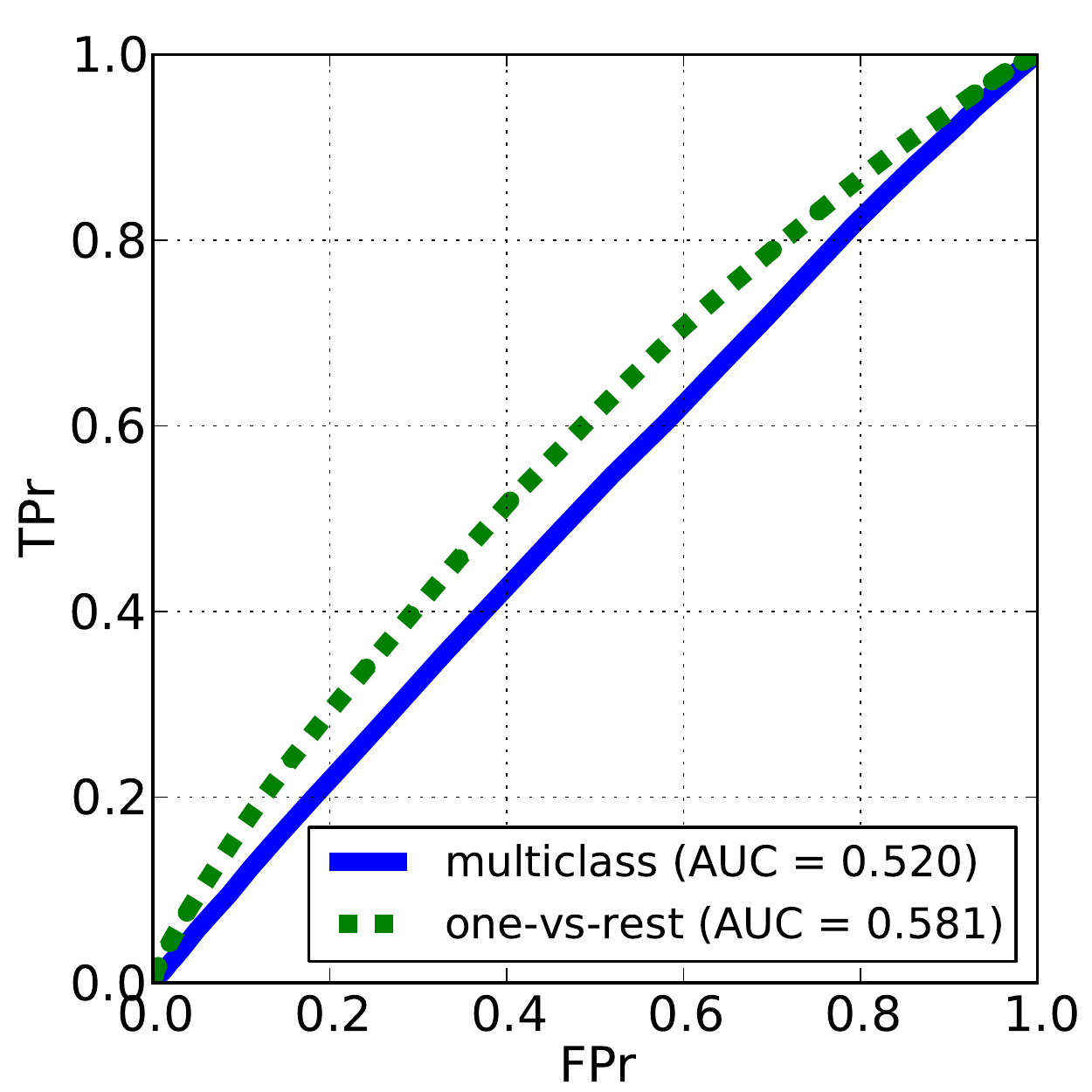}
}
\caption{ROC curves for multi-class classifiers and 1-vs-rest classifiers, averaged over (a) beautiful; and (b) ugly attributes.}
  \label{fig:roc_learning}
\end{figure}

It is also worth noting that both SIFT and color-based features are useful for the classification of attributes.
This is not surprising since some attributes are very color-related (``nice colors", ``black background"), while others are well-captured by gradient information (``leading lines", ``great sharpness"). As Fig.~\ref{fig:roc_ptype} shows, combining SIFT and color features results in increased performance.
We also compared the performances of two learning approaches: 1-vs-rest against multi-class classifiers \citep{crammer2002algorithmic}. As shown in Fig.~\ref{fig:roc_learning}, the former strategy provided significantly better results experimentally. This may result from a large overlap between attribute classes in feature space, a regime in which multi-class classification has been observed to perform poorly compared to one-vs-rest classification \citep{Akata2014}.

\subsection{The attribute representation}
\begin{figure*}[!ht]
\setlength{\tabcolsep}{3pt}
\centering
\begin{tabular}{r|l}
\input{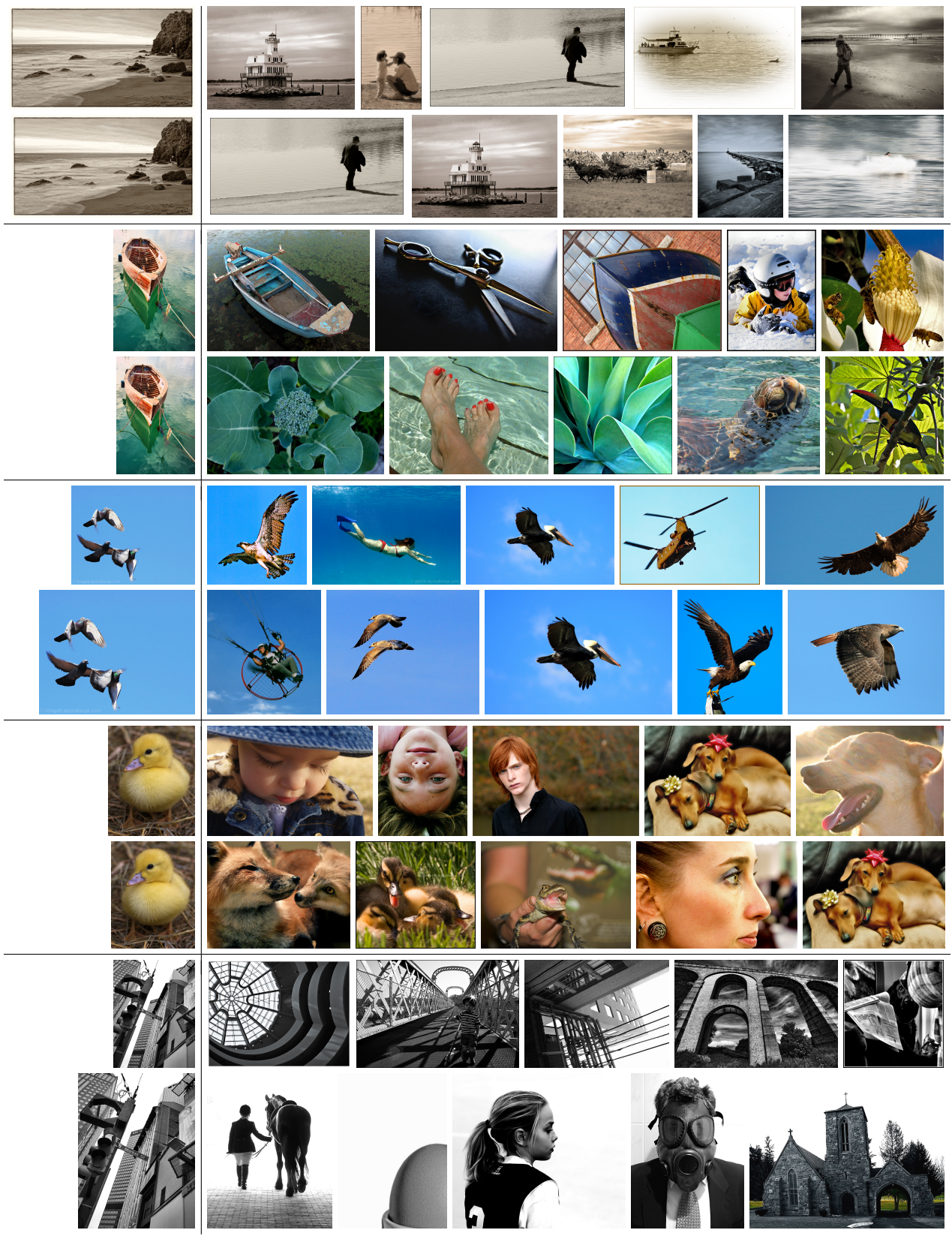}
\end{tabular}
\caption{Five randomly-chosen images and their 5 nearest neighbors in (i) the 200-dimensional attribute space (top rows); (ii) the original high-dimensional FV space (bottom rows).}
\label{fig:nn_attributes}
\end{figure*}

To form an image representation using our learned attribute classifiers,
we compute the classifier scores given to the image's FV by the 100 best-performing (in terms of AUC) beautiful and ugly attributes.
This results in a 200-dimensional real-valued attribute vector which we can use to train preference models (see section~\ref{sec:apps} for several applications).
Fig.~\ref{fig:nn_attributes} shows a random sample of images and their 5 nearest neighbors in the 200-dimensional attribute space,
as well as the original high-dimensional FV space.
The nearest neighbors often have similar color and composition attributes and, as with textual queries, similar semantic content.
When the query image has strong stylistic or compositional attributes, this is reflected in its nearest neighbors in attribute space.
This can be seen in the first query image in Fig.~\ref{fig:nn_attributes},
whose strong sepia tones and uncluttered composition are reflected in its nearest nearest neighbors in the attribute space.
The last query image contains strong line patterns and a black and white palette, attributes which are well represented in its nearest neighbors.
The nearest neighbors in the FV space reflect these attributes less uniformly.

\section{Applications}
\label{sec:apps}

In this section we consider three applications of the proposed attributes:
aesthetic prediction,
image tagging,
and query-by-text image retrieval.

\subsection{Aesthetic prediction}

\begin{figure}
 \centering
\subfigure[AVA]{
    \includegraphics[width=0.49\linewidth]{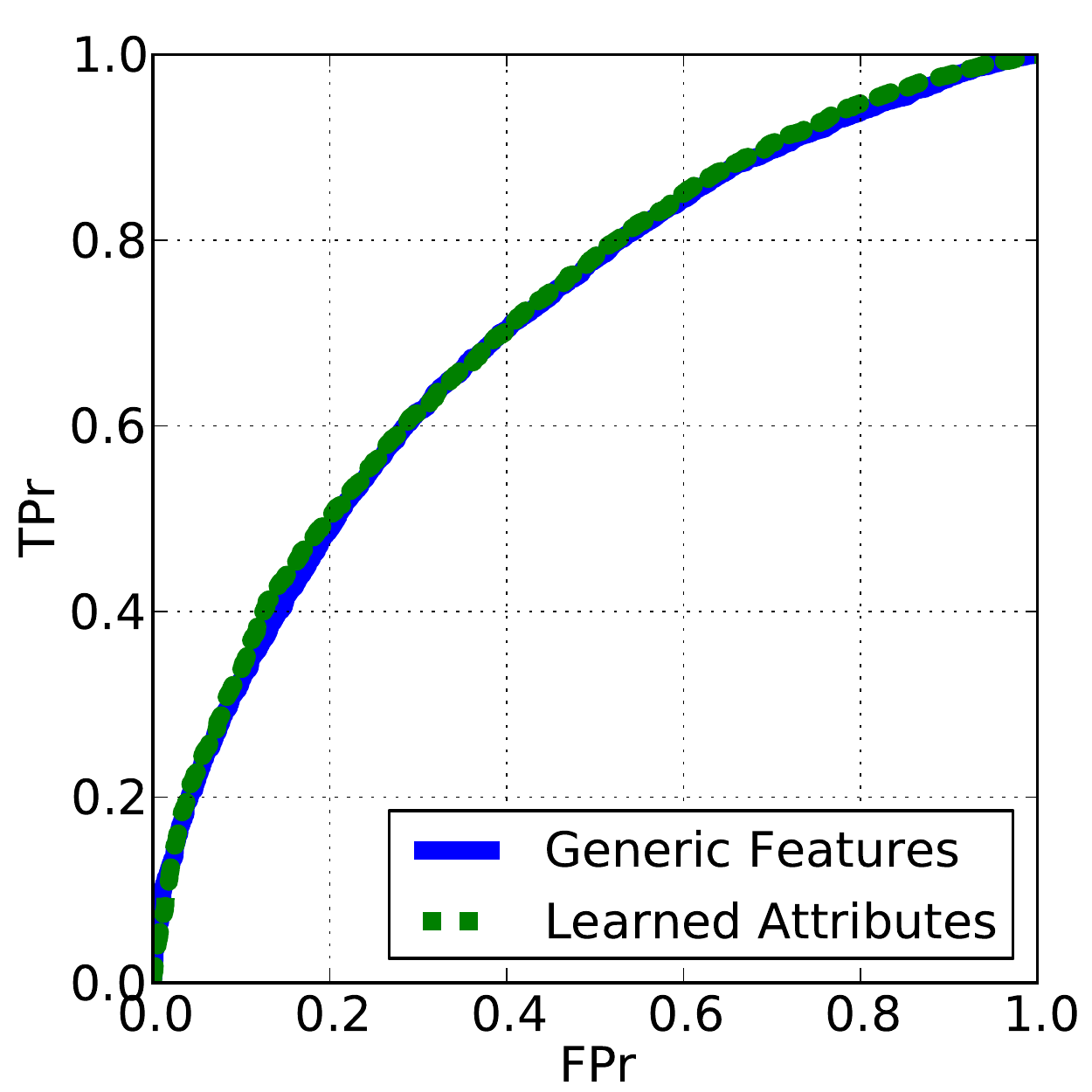}\label{fig:roc_ava}
}\subfigure[Photo.net]{
    \includegraphics[width=0.49\linewidth]{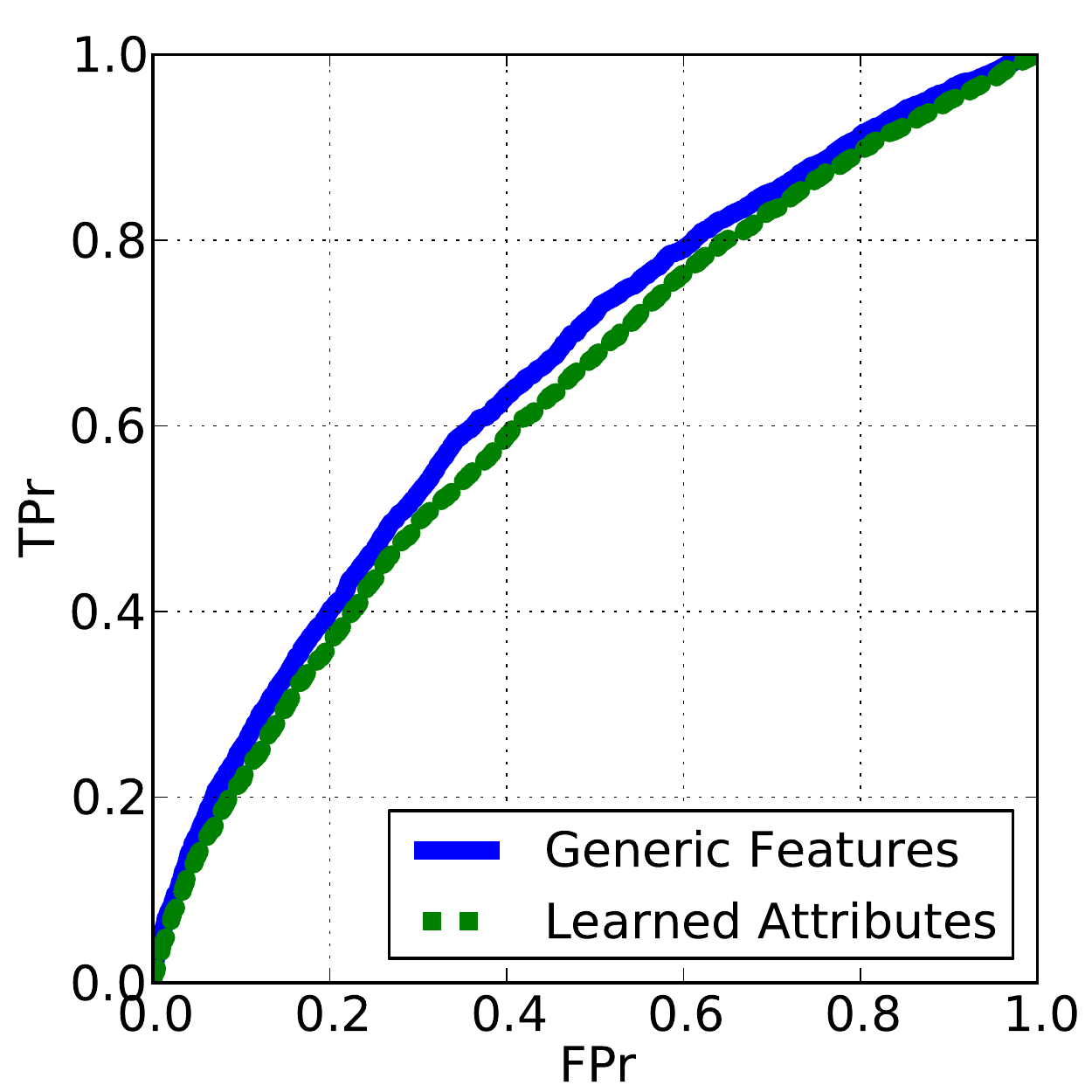}\label{fig:roc_pn}
}
\caption{Aesthetic preference prediction: comparison between learned attributes and generic visual features \citep{Marches2011} for the AVA dataset (a), and with generalization to Photo.net images (b).}
  \label{fig:roc}
\end{figure}

In some cases, we might be interested in giving a binary answer regarding the attractiveness of an image: beautiful versus ugly.
Such binary decisions are the organizing principle behind online photo-sharing venues such as {\tt www.imgur.com} (via ``like" and ``dislike" buttons)
and {\tt http://www.reddit.com/r/itookapicture/} (via ``upvote" and ``downvote" buttons). 
We therefore propose to use our learned attributes to make such a prediction
and compare to the approach of \citep{Marches2011} which is based on generic image features and is to date the best-performing baseline on AVA dataset.
To make the comparison with \citep{Marches2011}, we use the same FV features and linear classifiers in both cases. 
As can be seen in Fig.~\ref{fig:roc_ava}, attributes perform on par with
low-level generic features, despite the significant difference in dimensionality (131,072
dimensions for the low-level features and 200 dimensions for the
attributes).
Therefore attributes achieve equivalent performance (AUC=0.718 for attributes, versus 0.715 for generic generic features)
and introduce the possibility of replacing a single image attractiveness label
(beautiful or ugly) with the labels of the most responsive attributes.
Note that while one can also reduce the dimensionality of the FVs using random projections or PCA,
there is no guarantee that the new dimensions will be human-interpretable, and even if so, they would need to be manually labeled.
\\[2mm]
{\bf Generalization performance.}
To investigate the generalizability of the attributes, we evaluated their performance on images obtained from Photo.net.
We downloaded a random selection of 10K training and testing images, and 7K validation images, along with their mean aesthetic scores.
Our attribute vectors achieved AUC=0.631 on the test set, compared to AUC=0.659 for generic FV (see also Fig.~\ref{fig:roc_pn}), demonstrating that our attributes
can indeed be applied to predict aesthetic preference for images collected in an entirely different context.
In addition, for the price of a small performance decrease compared to FV, one gains interpretability of aesthetic preference,
without using any textual meta-data that may be associated with the Photo.net images.
We note that this scenario, in which images are assumed to have no aesthetics-related textual meta-data, is by far the most typical in existing image corpora.

\subsection{Image tagging}
We now go beyond tagging an image as {\em beautiful} or
{\em ugly}, as a binary decision can be too aggressive for a problem as subjective as aesthetic quality.
Indeed, it could form a positive or negative prior in the user's mind in contradiction
to his/her tastes and opinions. 

To gain users' consensus we design an application that not only predicts
aesthetic quality (\textit{Is this image beautiful or ugly?})  
but also produces a qualitative description of the aesthetic properties of an
image in terms of beautiful/ugly attributes. 
As can be seen from the examples of Table \ref{tab:qualitative}, this strategy
gives the user higher degree of interpretation of the aesthetic quality.  
For instance, while many users might agree that the leftmost image is a
beautiful picture,  
others might disagree that the yellow flower on the right is ugly: in general
people tend to refuse criticism.  
Instead, with attributes such as \textit{more light}, \textit{more depth field
  of view} and \textit{not sure} the application takes a more cautious approach
and enables the user to form his/her own opinion.  
Finally, we realize that these are just plausible hypotheses that should be tested with a full-fledged user study.
However such an evaluation is out of the scope of this work.
\begin{table*}[t!]
\footnotesize
\centering
  \begin{tabular}[c]{|p{3.5cm}|p{3.5cm}|p{3.5cm}|p{3.5cm}|}
\hline
 \includegraphics[width=.2\textwidth]{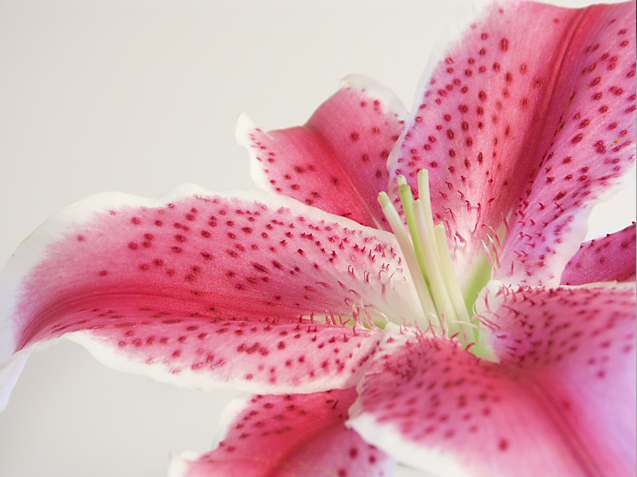}& \includegraphics[width=.12\textwidth]{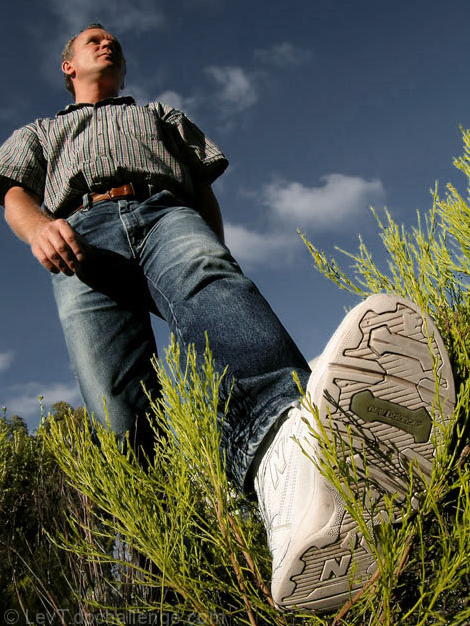}& \includegraphics[width=.2\textwidth]{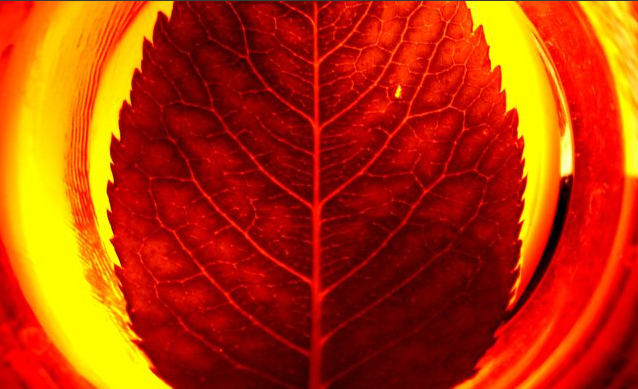} & \includegraphics[width=.2\textwidth]{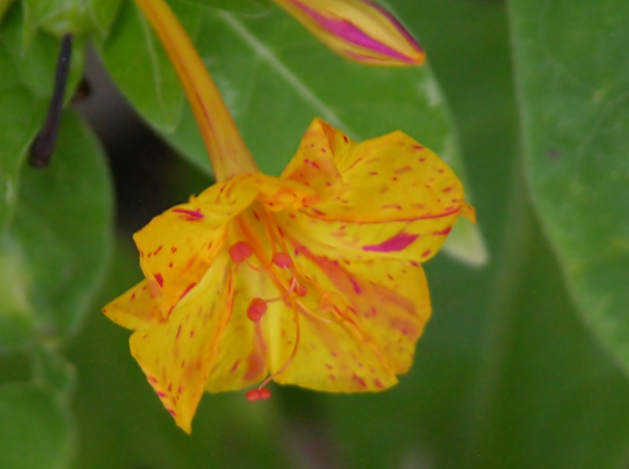} \\ \small great\_macro, very\_pretty, great\_focus, nice\_detail, so\_cute & \small great\_capture, great\_angle, nice\_perspective, lovely\_photo, nice\_detail & \small  more\_dof, not\_sure, too\_busy, motion\_blur, blown\_out &  \small soft\_focus, not\_sure, more\_light, sharper\_focus, more\_dof\\
\hline
  \end{tabular}
  \caption{Sample results for an image annotation application where the aesthetic quality of each image is described using the 5 most reactive attributes.}
  \label{tab:qualitative}
\end{table*}

\subsection{Query-by-text image retrieval}
\begin{figure*}[!ht]
  \centering
{\tiny beautiful\_colors}
\includegraphics[width=\linewidth]{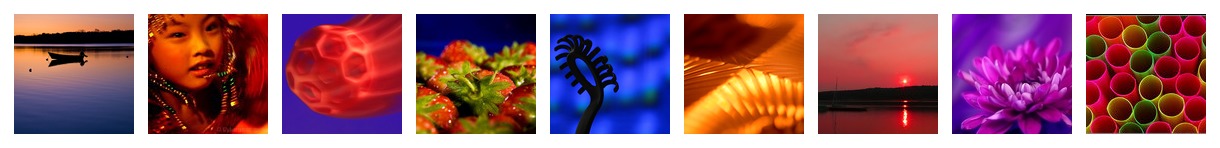}
{\tiny nice\_perspective}
\includegraphics[width=\linewidth]{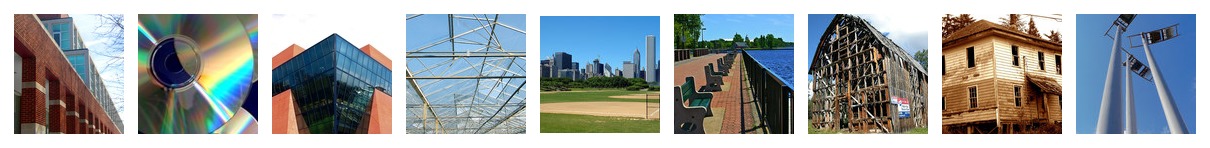}
{\tiny great\_sharpness}
\includegraphics[width=\linewidth]{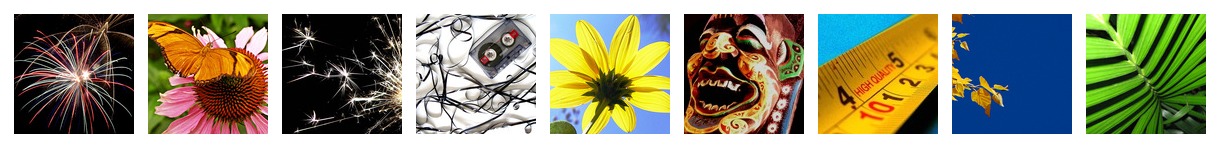}
{\tiny white\_balance}
\includegraphics[width=\linewidth]{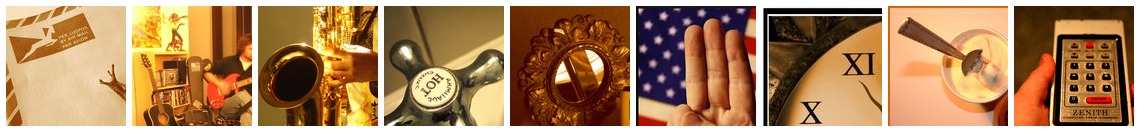}
{\tiny blown\_out}
\includegraphics[width=\linewidth]{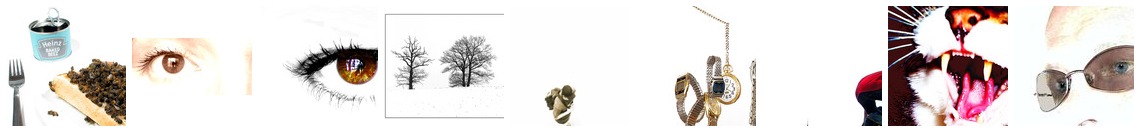}
{\tiny too\_busy}
\includegraphics[width=\linewidth]{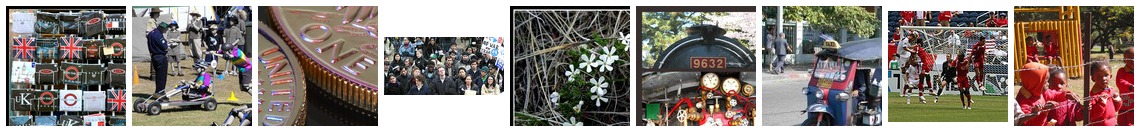}

\caption{Images with top scores for some representative beautiful and ugly attributes.}
\label{fig:mosaic1_attributes}
\end{figure*}

We now show how the learned attributes, evaluated quantitatively in section~\ref{sec:att_learning} (see Fig.~\ref{fig:apbu}),
can be used to perform attribute-based image retrieval.
We display the top-returned results of several queries for Beautiful and Ugly attributes in the mosaic of Fig.~\ref{fig:mosaic1_attributes}. 
We notice that the images clearly explain the labels discovered in AVA even for fairly complex attributes such as \textit{too busy}, \textit{blown out},
\textit{white balance} (note the various kind of color casts present in the images of row 6) or {\em Much noise} in the last row. 

The top-ranked images sometime contain very similar semantic content.
For example, the top-ranked images for the attribute \textit{nice perspective} are almost all images of architectural structures.
This indicates that our visual attributes may be highly correlated with semantic information,
which is unsurprising given that photographic style is very content-dependent.
An interesting topic for future work would involve leveraging semantic annotations (which are present in AVA)
in order to design learning strategies that overcome this potential limitation.

\begin{figure*}[!ht]
\setlength{\tabcolsep}{3pt}
\centering
\begin{tabular}{b{3cm}|l}
\input{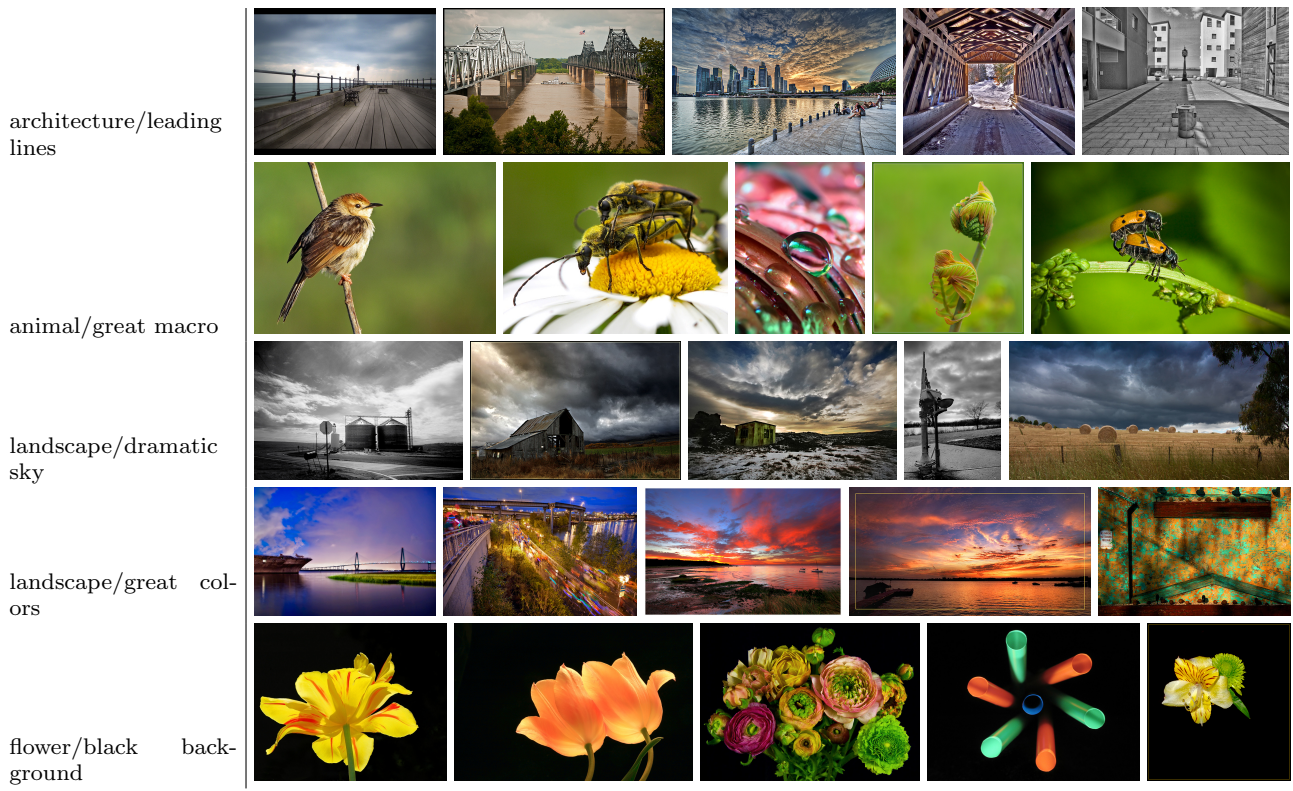}
\end{tabular}
\caption{Images with top scores for some sample joint attribute/semantic queries.}
\label{fig:att_sem_qry}
\end{figure*}

Our learned attributes may also be used in combination with semantic models to enable joint attribute/semantic queries.
To demonstrate this we train classifiers, using the same train/val/test splits, 
for the 8 semantic categories studied by \cite{murray12}: ``animal", ``architecture", ``cityscape", ``floral", ``fooddrink", ``landscape", ``portrait", and ``stilllife".
For a joint query such as \textit{landscape} with \textit{great colors}, 
we first apply the \textit{landscape} semantic classifier and 
the \textit{great colours} attribute classifier to the test images.
These scores are converted to probabilities,
multiplied and then sorted to produce a final ranking of the test images with respect to the joint query.
We use multiplication to approximate the ``AND'' operator,
as we want images relevant to {\em both} terms in the query to be the most highly ranked.
While more sophisticated fusions are possible \citep{murraylearning},
their evaluation for this task falls out of the score of this work.
Fig.~\ref{fig:att_sem_qry} shows the top 5 results for some sample joint queries.
Once again, the images clearly reflect the attributes, and also contain relevant semantic content.
Note for instance that the two landscape-related queries return very different top results
due to the different attributes requested: \textit{dramatic sky} vs \textit{great colours}.

\subsubsection*{User Study}

Images in AVA are only partially annotated with semantic and attribute information.
For instance, although many of the displayed results shown in Fig.~\ref{fig:att_sem_qry} are reasonable, they could be counted as errors because they lack the corresponding semantic or aesthetic tag. 
Consequently, a quantitative evaluation of the image retrieval results that relies solely on AVA annotations would provide a very pessimistic performance estimate.

Therefore, to assess the quality of these results we performed a user study using CrowdFlower \footnote{http://www.crowdflower.com/}, one of the leading crowdsourcing platforms.
The setup of the experiment was the following: we showed crowdsourcing workers an image and we asked two questions about its relevance to the query (e.g. ``1. Determine if the image subject is ARCHITECTURE", ``2. Determine if the image features the photographic technique LEADING LINES").
The semantic and attribute relevance were assessed independently for two reasons: 
firstly, we wanted to simplify as much as possible the task of the workers.
Secondly, we wanted to compare the performance of semantic to aesthetic attribute retrieval.
A three-value scale (Agree, Unsure, Disagree) was chosen.
Three judgments per image were sufficient to get a high degree of agreement among randomly-chosen workers ($>$ 84\%).

We launched the experiment on the 5 joint queries 
shown in Fig.~\ref{fig:att_sem_qry}.
For each query the top 200 images retrieved by the automatic classifier were used in the study.
Images were randomized before they were shown to workers. 

\begin{figure}
 \centering
\subfigure[Precision @ K]{\hspace{11mm}
    \includegraphics[width=0.88\linewidth]{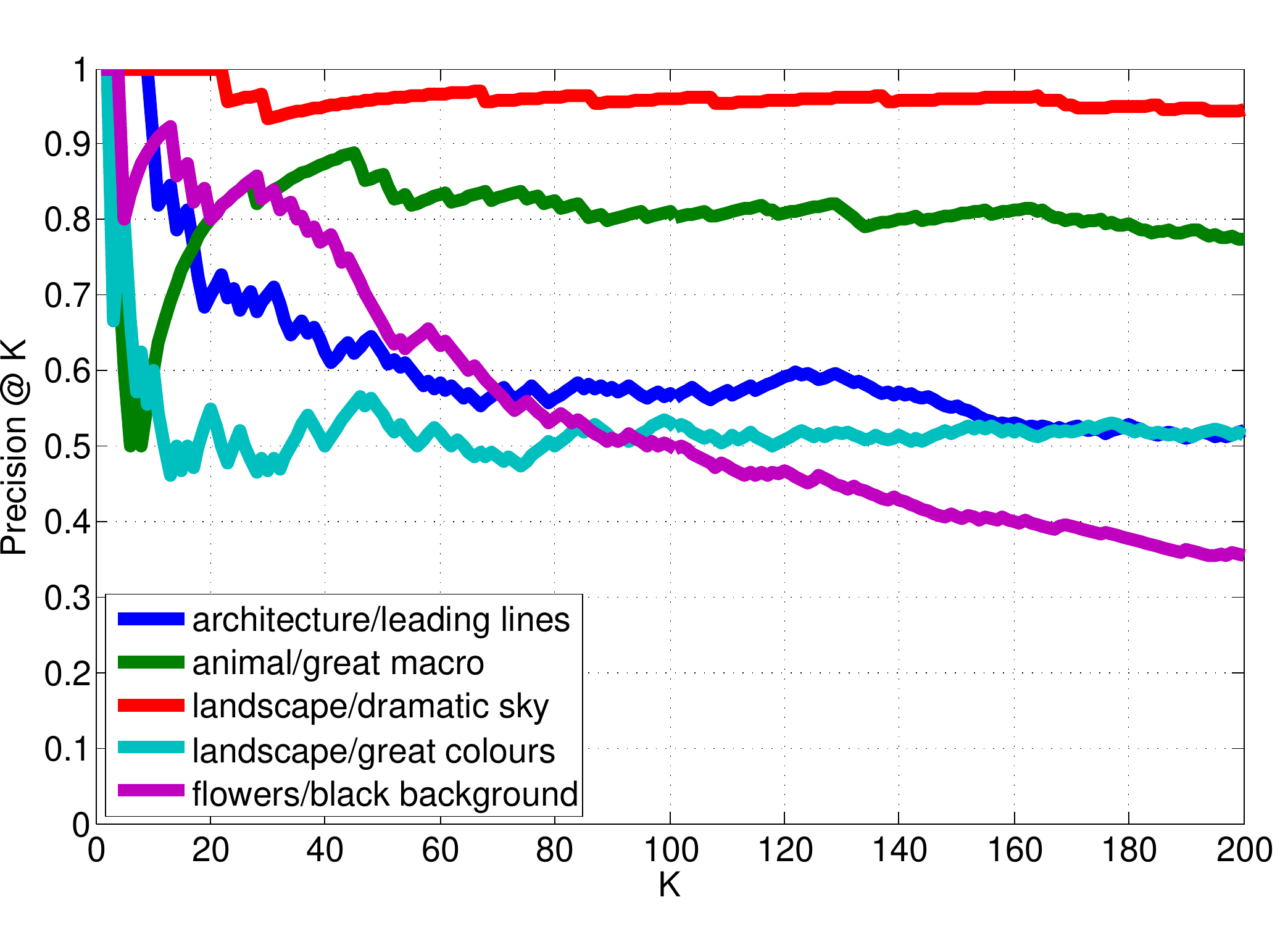}\label{fig:qbe_prec}
}
\subfigure[Errors per query]{
    \includegraphics[width=0.99\linewidth]{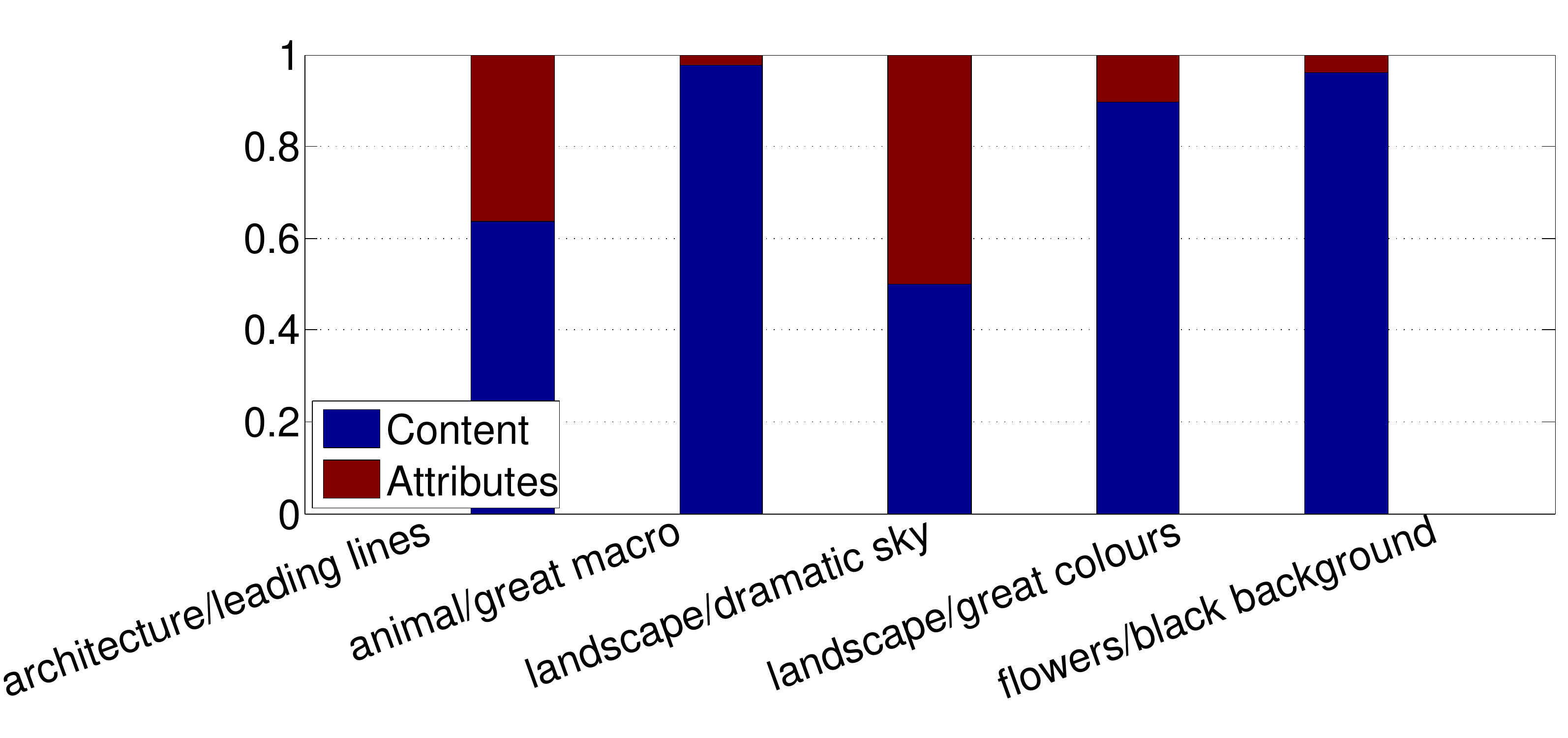}\label{fig:qbe_err}
}
\subfigure[Average annotation confidence]{
    \includegraphics[width=0.99\linewidth]{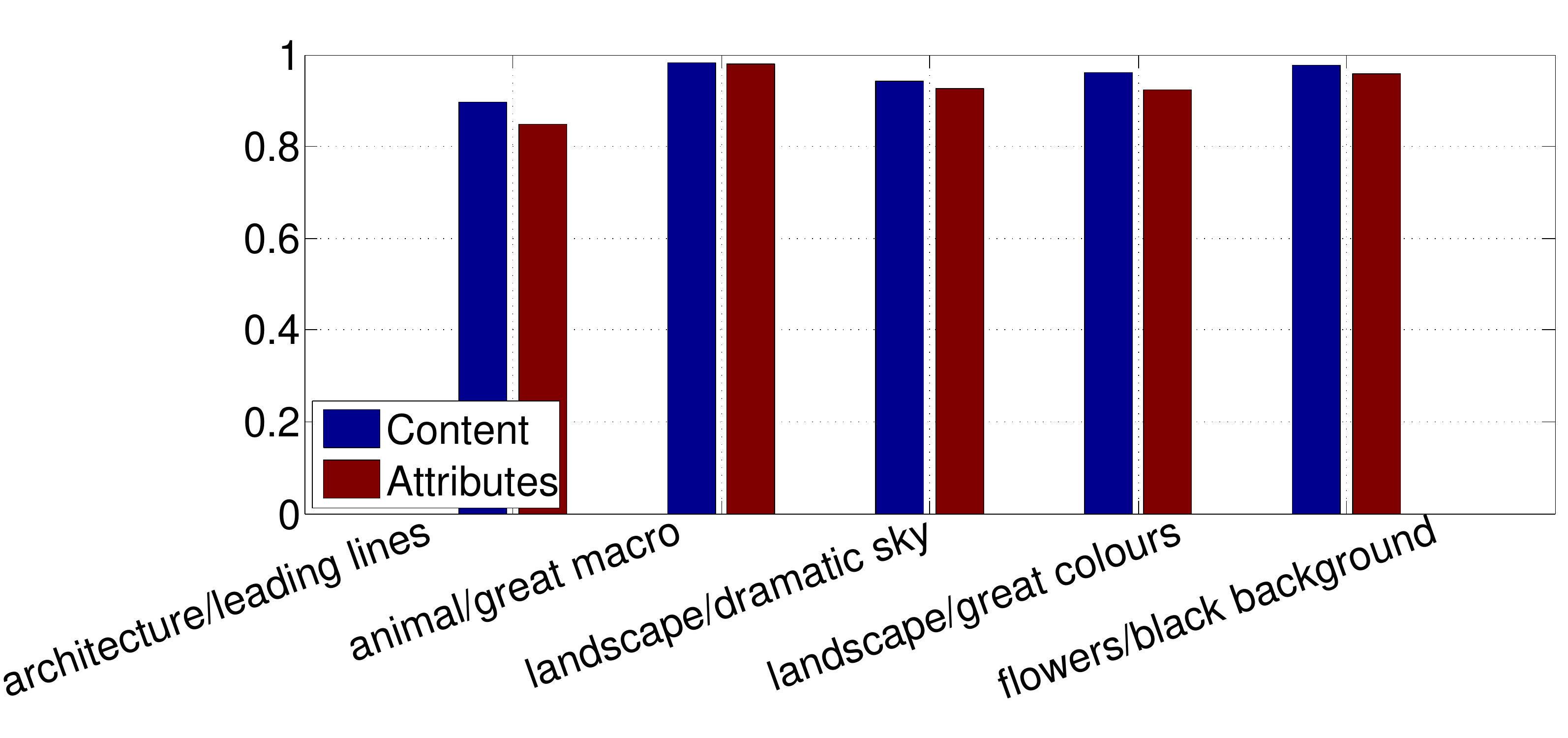}\label{fig:qbe_conf}
}
\caption{Results of user study on image retrieval, showing (a) precision@K; (b) proportion of errors per query related to either semantics or attributes; and (c) the average annotation confidence, i.e. agreement, per query.}
  \label{fig:qbe_study}
\end{figure}

 
To assess the quality of the ranks we used Precision@K (see Fig.~\ref{fig:qbe_prec}). To get maximum precision, both attributes had to be assessed by three workers in an image. 
The best performing queries are landscape/dramatic sky and animals/macro.

To gain a deeper understanding of these results, we also counted the errors among aesthetic and semantic attributes on a per-query basis. 
The results are shown in Fig.~\ref{fig:qbe_err}. 
Two conclusions can be drawn: firstly, most errors are semantic.
Secondly, flowers and animals are the queries where the content attributes have lowest performance.
While aesthetic attributes based on color ( e.g. ``black background", ``great colour") or simple composition properties  such as macro ( big object and out of focus background) perform well, other more complex composition attributes such as leading lines are more difficult to capture.

We also measured the confidence of responses for each query, shown in Fig.~\ref{fig:qbe_conf}.
The confidence here is measured as the agreement between the responses of the three workers on a per-image basis. 
As can be seen, confidence on content attributes is higher than confidence on aesthetic attributes: this coincides with the fact that, in general, semantic attributes are less subjective than aesthetic attributes. 

\section{Conclusions}
\label{sec:conclusions}

In this paper, we tackled the problem of visual attractiveness analysis using visual attributes as mid-level features. 
Despite the great deal of subjectivity of the problem, we showed that we can
automatically learn semantically-meaningful attributes
using the unique conjunction of image, scoring, and textual data in the AVA dataset, 
for which we provided an in-depth analysis.
We demonstrated the effectiveness of our attributes in various
applications such as score prediction, image auto-tagging or image retrieval. 
Future work will focus on testing with users the advantage of our beautiful and ugly attributes and
on mitigating biases introduced by semantic information.

\begin{acknowledgements}
The authors would like to thank Jean-Michel Renders for the discussions about
text analysis and Isaac Alonso for having supported the experimental work of
this paper. 
\end{acknowledgements}

\bibliographystyle{spbasic}           
\bibliography{ijcv}                   

\end{document}